\documentclass[11pt]{article}

\usepackage[final]{acl}

\usepackage{times}
\usepackage{latexsym}
\usepackage[T1]{fontenc}
\usepackage[utf8]{inputenc}
\usepackage{microtype}
\usepackage{inconsolata}
\usepackage{graphicx}

\usepackage{amsmath}
\usepackage{amssymb}
\usepackage{amsfonts}
\usepackage{mathtools}
\usepackage{amsthm}
\usepackage{nicefrac}
\usepackage{bbm}
\usepackage{commath}
\usepackage{contour}

\usepackage{algorithm}
\usepackage[noend]{algorithmic}

\usepackage{booktabs}
\usepackage{array}
\usepackage{multirow}
\usepackage{makecell}
\usepackage{colortbl}
\usepackage{float}
\usepackage{wrapfig}
\usepackage{subfigure}

\usepackage{color}
\usepackage{tcolorbox}
\usepackage{mdframed}

\usepackage{placeins}

\newtcolorbox[auto counter, number freestyle={\noexpand\arabic{\tcbcounter}}]{mycolorbox}[3][]{%
    fonttitle=\bfseries,
    title=#3,
    #1
}

\usepackage{tikz}
\usetikzlibrary{decorations.pathreplacing,calc}
\newcommand{\tikzmark}[1]{\tikz[overlay,remember picture] \node (#1) {};}
\newcommand*{\AddNote}[4]{%
    \begin{tikzpicture}[overlay, remember picture]
        \draw [decoration={brace,amplitude=0.5em},decorate,ultra thick,red]
            ($(#3)!(#1.north)!($(#3)-(0,1)$)$) --
            ($(#3)!(#2.south)!($(#3)-(0,1)$)$)
                node [align=center, text width=1.0cm, pos=0.5, anchor=west] {#4};
    \end{tikzpicture}
}

\usepackage[capitalize,noabbrev]{cleveref}
\crefformat{footnote}{#2\footnotemark[#1]#3}

\usepackage[toc,page]{appendix}

\usepackage[textsize=tiny]{todonotes}

\newcommand{\alg}{\texttt{EXPO}}
\newcommand{\alges}{\texttt{EXPO-ES}}
\newcommand{\expect}[1]{\mathbb{E}\left[ #1 \right]}

\newcommand{\squishlisttwo}{
 \begin{list}{$\bullet$}
  { \setlength{\itemsep}{1pt}
    \setlength{\parsep}{0pt}
    \setlength{\topsep}{0pt}
    \setlength{\partopsep}{0pt}
    \setlength{\leftmargin}{1em}
    \setlength{\labelwidth}{1.5em}
    \setlength{\labelsep}{0.5em} } }
\newcommand{\squishend}{
  \end{list}  }

\makeatletter
\def\blankfootnote{\xdef\@thefnmark{}\@footnotetext}
\makeatother

\title{Meta-Prompt Optimization for LLM-Based Sequential Decision Making}

\author{
  Mingze Kong\textsuperscript{1} \quad
  Zhiyong Wang\textsuperscript{2} \quad
  Yao Shu\textsuperscript{3} \quad
  Zhongxiang Dai\textsuperscript{1$\dagger$} \\
  \textsuperscript{1}The Chinese University of Hong Kong, Shenzhen \\
  \textsuperscript{2}Harbin Institute of Technology (Shenzhen) \\
  \textsuperscript{3}The Hong Kong University of Science and Technology (Guangzhou) \\
  \small{\textsuperscript{$\dagger$}Corresponding author: \href{mailto:daizhongxiang@cuhk.edu.cn}{daizhongxiang@cuhk.edu.cn}}
}

\begin{document}
% Override acl.sty's \flushbottom: send vertical slack to column bottoms
% instead of stretching it into visible gaps between paragraphs/after headings.
\raggedbottom
\maketitle

\begin{abstract}
Large language models (LLMs) have recently been employed as agents to solve sequential decision-making tasks such as Bayesian optimization and multi-armed bandits (MAB). These works usually adopt an LLM for sequential action selection by providing it with a fixed, manually designed \emph{meta-prompt}. However, numerous previous works have found that the prompt has a significant impact on the performance of the LLM, which calls for a method to automatically optimize the meta-prompt for LLM-based agents. Unfortunately, the non-stationarity in the reward observations during LLM-based sequential decision-making makes meta-prompt optimization highly challenging. To address this challenge, we draw inspirations from \emph{adversarial bandit} algorithms, which are inherently capable of handling non-stationary reward observations. Building on this foundation, we propose our \emph{\underline{EXP}onential-weight algorithm for prompt \underline{O}ptimization} (\alg) to automatically optimize the task description and meta-instruction in the meta-prompt for LLM-based agents. We also extend \alg{}~to additionally optimize the exemplars (i.e., history of interactions) in the meta-prompt to further enhance the performance, hence introducing our \alges{}~algorithm. We use extensive experiments to show that our algorithms significantly improve the performance of LLM-based sequential decision-making.
\end{abstract}

\section{Introduction}
\label{sec:intro}
The strong capabilities of LLMs have spurred significant recent interests in adopting them as agents to solve sequential decision-making problems, such as multi-armed bandits (MAB) \cite{krishnamurthy2024can}, Bayesian optimization (BO) \cite{yang2023large} and reinforcement learning (RL) \cite{dai2024context}. Specifically, these methods often use an LLM to sequentially select the actions by providing it with a specially designed prompt, which we refer to as the \emph{meta-prompt}.
The meta-prompt often contains several components, such as the \emph{task description}, the \emph{meta-instruction} (which is used to instruct the LLM to select an action in every step), the \emph{history of interactions} with the environment, among others. 
The previous methods have all adopted a fixed, manually designed meta-prompt for the LLM-based agent throughout the entire sequential decision-making process. However, numerous previous works have highlighted that the output text generated by LLMs is heavily dependent on its input prompt \cite{zhou2023large}. 
Therefore, using fixed, manually designed meta-prompt may significantly limit the performance of the LLM-based agents, because handcrafted prompts are often far from optimal \cite{zhou2025multi}. This naturally begs the question: \emph{can we automatically optimize the meta-prompt for LLM-based agents to enhance their performance?}

The sensitivity of LLM-generated text to its input prompt has given rise to many recent works on \emph{automated prompt optimization}, among which a representative line of works have adopted the method of multi-armed bandits (MAB) to automatically optimize the prompt \cite{lin2024prompt,lin2023instinct,wu2024prompt}.
Unfortunately, the problem of meta-prompt optimization for LLM-based agents presents significant challenges compared to traditional prompt optimization. 
This is mostly due to the \emph{non-stationarity in the observed rewards} during the LLM-based sequential decision-making process. 
Specifically, as the LLM-based agent engages in more interactions with the environment, its state in the environment changes, making its observed rewards non-stationary. 
For example, in MAB \cite{krishnamurthy2024can} and BO \cite{yang2023large}, the observed rewards in later iterations tend to be higher than those in initial iterations; similarly, in RL \cite{dai2024context}, the rewards depend on the agent's state, which itself evolves across iterations.
As a consequence, for the same meta-prompt, its observed reward is likely to change dynamically across iterations. This is in stark contrast to classical prompt optimization, in which the score for a prompt remains stationary. 
As a result, this renders the previous works on prompt optimization (such as those based on MAB \cite{lin2024prompt,lin2023instinct,wu2024prompt}) inapplicable, and hence calls for novel algorithmic designs to solve the problem of meta-prompt optimization for LLM-based agents. To this end, we draw inspirations from the field of \emph{adversarial bandits} \cite{lattimore2020bandit}.

In adversarial bandits, the reward observations of each arm are chosen by an adversary and may change arbitrarily across iterations, in contrast to classical stochastic MAB in which each arm's rewards are sampled from a fixed stationary distribution. This ability to handle non-stationary reward observations makes adversarial bandits an ideal candidate for meta-prompt optimization for LLM-based agents. 
Specifically, drawing inspirations from the EXP3 algorithm for adversarial bandits, we introduce our \emph{\underline{EXP}onential-weight algorithm for prompt \underline{O}ptimization} (\alg) to optimize the task description and meta-instruction in the meta-prompt of an LLM-based agent.\footnote{Other components of the meta-prompt can also be optimized in a similar fashion.}
\begin{figure*}[t]
\vspace{-4mm}
\centering
\includegraphics[width=0.95\linewidth]{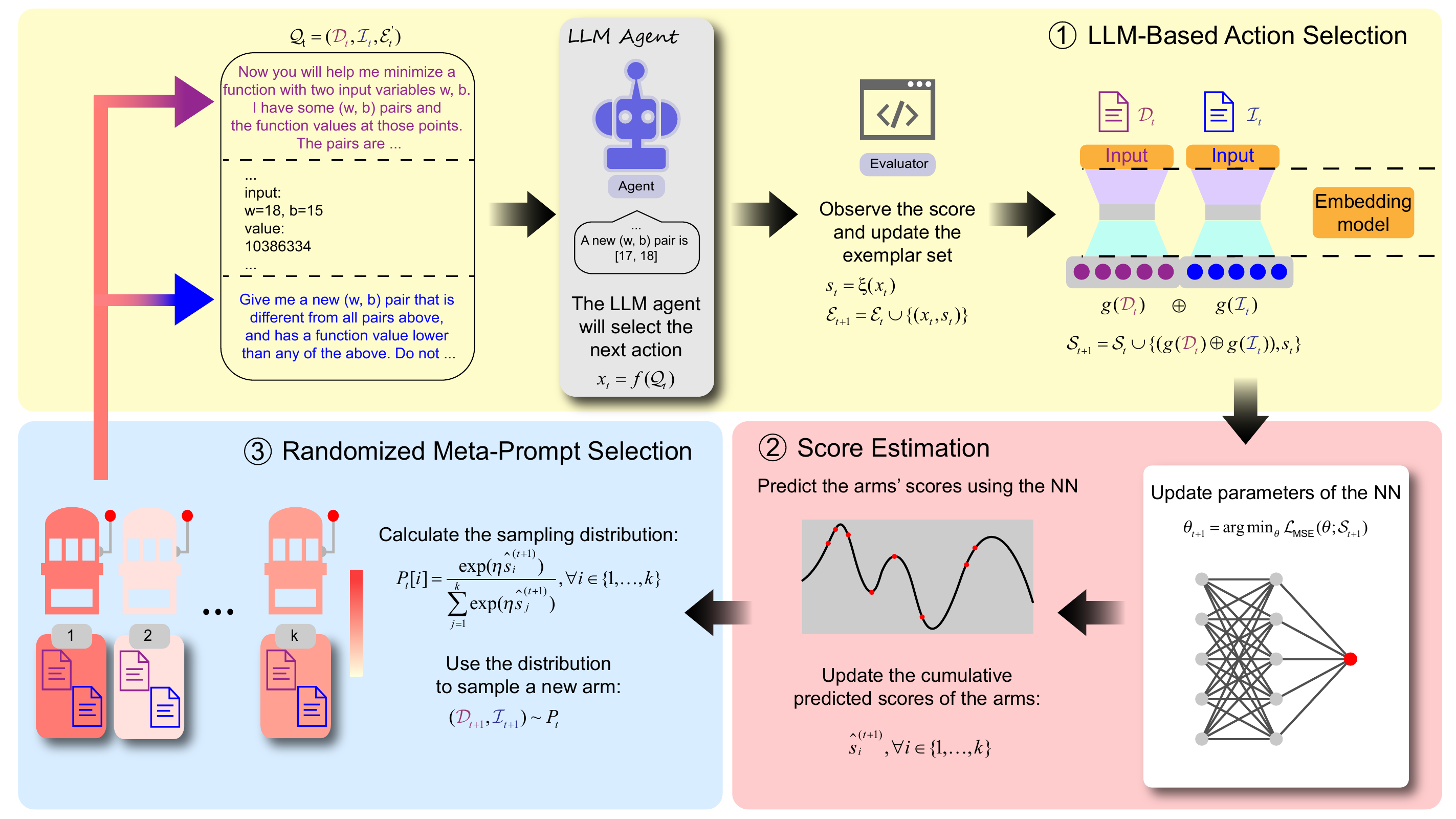} \\
\vspace{-3mm}
 \caption{
 % Illustration of \alg algorithm. 
 % {\color{violet}Purple} denotes task description and {\color{blue}blue} denotes meta-instruction.
 Illustration of our \alg~algorithm. 
 We use {\color{violet}purple} to denote the task description and {\color{blue}blue} to represent the meta-instruction.
 }
 \label{fig:expo:algo}
\vspace{-5mm}
\end{figure*}

In addition to the task description and meta-instruction, the \emph{history of interactions} with the environment (i.e., the \emph{exemplars}) is another crucial component of the meta-prompt, and previous works on in-context learning have shown that both the content and the ordering of exemplars significantly affect the performance of LLMs \cite{lu2022fantastically}.
Therefore, we further extend our \alg~to also optimize the subset of exemplars included in the meta-prompt and their ordering.
% , in order to further boost the performance of LLM-based agent.
However, since the task description and meta-instruction selected by \alg~vary across iterations, the reward observations for a fixed exemplar sequence become non-stationary as well.
To this end, we use a separate adversarial bandit method to optimize the exemplars (i.e., the interaction history), and hence introduce our \emph{\alg~with \underline{E}xemplar \underline{S}election} (\alges) algorithm.

We use extensive experiments to show that our \alg~significantly improves the performance of the LLM-based BO algorithm from \cite{yang2023large} (Sec.~\ref{subsec:exp:opro}) and the LLM-based MAB algorithm from \cite{krishnamurthy2024can} (Sec.~\ref{subsec:exp:bandits}).
Furthermore, in tasks where the exemplars 
% (i.e., the history of interactions) 
provide crucial information for the LLM-based agent, our \alges~further enhances the performance of \alg~via automated exemplar selection (Sec.~\ref{subsec:exp:opro}).
We also use ablation study to unveil 
% other
interesting insights about our algorithms (Sec.~\ref{sec:ablation}).
Due to space constraints, a detailed discussion of related work is deferred to App.~\ref{app:sec:related_work}.
% e.g., our adversarial bandit-based algorithm indeed considerably outperforms baseline methods adopting stochastic MAB algorithms which corroborates the motivation for our algorithms described above (Sec.~\ref{subsec:ablation:ucb}), the optimal meta-prompts discovered by our algorithm lead to dramatic performance improvement to LLM-based agents (Sec.~\ref{subsec:ablation:using:optimal:prompt}).

\section{Problem Setting}
Throughout our work, we use \emph{arms} to represent meta-prompts, and use \emph{actions} to denote the actions selected by an LLM-based agent.
Consider an algorithm which uses an LLM to perform a sequential decision-making task by sequentially instructing the LLM to select an action in every iteration.
A representative example of such algorithms is the \emph{Optimization by PROmpting} (OPRO) algorithm from \cite{yang2023large}. OPRO aims to solve an optimization problem, i.e., to find $x^* = {\arg\min}_{x}f(x)$. To achieve this, in every iteration $t$, OPRO uses an LLM to select a batch of $B$ input queries $\{x_{t,1},\ldots,x_{t,B}\}$, after which their corresponding scores $\{s_{t,1},\ldots,s_{t,B}\}$ are observed. When instructing the LLM to select the input queries, the meta-prompt $\mathcal{Q}$ given to the LLM contains a number of important components, including \emph{a fixed task description} $\mathcal{D}$, \emph{a fixed meta-instruction} $\mathcal{I}$, and \emph{a sequence of exemplars} $\mathcal{E}_t$ corresponding to a subset of the observations (i.e., pairs of input queries and observed scores) collected so far.
The same paradigm of LLM-based sequential decision making has also been adopted by other works, such as the LLM-based MAB algorithm from \cite{krishnamurthy2024can} (more details in Sec.~\ref{subsec:exp:bandits}).

In this work, our first algorithm, \alg~(Sec.~\ref{subsec:expo}), dynamically optimizes the task description $\mathcal{D}$ and meta-instruction $\mathcal{I}$ (i.e., selects a new $\mathcal{D}_t$ and $\mathcal{I}_t$ in every iteration $t$), and our extension \alges~(Sec.~\ref{subsec:expo:es}) additionally optimizes the exemplar sequence $\mathcal{E}_t$.
We use $g(\cdot)$ to denote a pre-trained embedding function, and separately obtain the embeddings of the task description $g(\mathcal{D}_t)$, the meta-instruction $g(\mathcal{I}_t)$ and the exemplar sequence $g(\mathcal{E}_t)$.
In every iteration, we use the history of selected meta-prompts and their scores to train a neural network (NN), denoted $\mathcal{M}(g(\cdot); \theta)$ with parameters $\theta$, which predicts the scores of meta-prompts in the domain.

\textbf{Adversarial Bandits.}
In adversarial bandits, the goal is to compete against the best arm \emph{in hindsight} \cite{lattimore2020bandit}. 
Consider an MAB problem with $k$ arms (i.e., meta-prompts).
For each arm $i=1,\ldots,k$, denote its corresponding sequence of rewards (i.e., scores) in $T$ iterations as $\{r_{t,i}\}_{t=1,\ldots,T}$.
% , and denote the arm selected by an adversarial bandit algorithm in iteration $t$ as $A_t$.
The best arm in hindsight is then defined as $i^*={\arg\max}_{i=1,\ldots,k}\sum^T_{t=1} r_{t,i}$.
Then, the goal of an adversarial bandit algorithm (which selects arm $A_t$ in iteration $t$) is to minimize the following definition of regret: $R_T=\sum^T_{t=1} r_{t,i^*} - \sum^T_{t=1} r_{t,A_t}$.

\textbf{Adversarial Bandits for LLM-Based Agents.}
LLM-based sequential decision-making methods often aim to maximize either \emph{(a)} the cumulative rewards 
(e.g., the LLM-based MAB algorithm from \cite{krishnamurthy2024can}) 
or \emph{(b)} the final reward (e.g., OPRO from \cite{yang2023large}).
% In our formulation of adversarial bandit, for both types of methods, 
In the former case, the cumulative rewards of the best arm $i^*$ are higher than those of the other arms; in the latter case, we implicitly assume that the arm with the largest final reward after $T$ iterations also tends to have large rewards across all iterations.
As a result, in both cases, \emph{the observed rewards of an arm in every iteration are indicative of the quality of the corresponding meta-prompt}.
So, when training the NN $\mathcal{M}(g(\cdot); \theta)$ for score prediction, we simply use the observed scores (i.e., rewards) as the training labels, which we find sufficient for strong empirical performance (Sec.~\ref{sec:experiments}).

\section{Algorithms}
\label{sec:expo}
\vspace{-2mm}
\subsection{The \alg~Algorithm (Algo.~\ref{algo:EXPO})}
\label{subsec:expo}
\vspace{-2mm}
% Our \alg~is used to dynamically optimize the task description $\mathcal{D}$ and the meta-instruction $\mathcal{I}$ in the meta-prompt. 
% Our \alg~ dynamically optimizes the task description $\mathcal{D}$ and the meta-instruction $\mathcal{I}$ in the meta-prompt. 

\textbf{Domain Generation.}
We first generate the domain of task descriptions and meta-instructions. Following previous works on prompt optimization \cite{lin2024prompt,lin2023instinct,zhou2023large}, we use an LLM to rephrase an initial task description $\mathcal{D}_0$ (resp.~initial meta-instruction $\mathcal{I}_0$) into $k_1$ task descriptions (resp.~$k_2$ meta-instructions), yielding a domain of size $k=k_1\times k_2$ (more details in App.~\ref{app:subsec:detail:domain:generation}).
We treat each combination of a task description $\mathcal{D}$ and a meta-instruction $\mathcal{I}$ as an \emph{arm}, so our adversarial bandit problem has $k$ arms.
We jointly optimize $\mathcal{D}$ and $\mathcal{I}$, as the ablation in Fig.~\ref{fig:task_description_meta_instructions_ablation} shows this outperforms optimizing them separately.

\begin{algorithm}[!ht]
\begin{algorithmic}[1]
    % \REQUIRE Number of rounds $T$; initial prompt $\mathcal{Q}_0 = (\mathcal{D}_0, \mathcal{I}_0, \mathcal{E}_0)$, where $\mathcal{D}_0$ denotes the task description, $\mathcal{I}_0$ denotes the meta-instructions, and $\mathcal{E}_0$ denotes the exemplars set $\mathcal{E}_0 = \{(g(y_i), s_i)\}$ containing embeddings of reasoning results $g(y_i)$ and their associated scores $s_i$; Neural Network model $\mathcal{M}$; EXP3 parameters $\eta$ and $k$ meta-prompt combinations (arms).
    \INPUT: Initial task description $\mathcal{D}_0$, initial meta-instruction $\mathcal{I}_0$.
    \STATE Initialize the exemplar set $\mathcal{E}_0 = \emptyset$, and the subset $\mathcal{E}_0' = \emptyset$, meta-prompt-score set $\mathcal{S}_0 = \emptyset$, and cumulative score estimates $\hat{s}^{(0)}_i=0$ for all $i \in \{1, \ldots, k\}$.
    \FOR{iteration $t = 0, 1, \ldots, T-1$}
        \STATE Construct 
        % the current 
        meta-prompt $\mathcal{Q}_t = (\mathcal{D}_t, \mathcal{I}_t, \mathcal{E}_t')$.
        % \STATE Obtain reasoning results $x_t$ by querying the LLM with the current prompt $\mathcal{Q}_t$:
        % \begin{equation}
        % x_t = f(\mathcal{Q}_t),
        % \end{equation}
        % where $f(\cdot)$ denotes the LLM.
        \STATE Query the LLM $f(\cdot)$ using the meta-prompt $\mathcal{Q}_t$ to select the next action $x_t$: $x_t = f(\mathcal{Q}_t)$.
        % \begin{equation}
        % x_t = f(\mathcal{Q}_t).
        % \end{equation}
        \STATE Observe the score $s_t$ for $x_t$ using the task-specific evaluator: $s_t = \xi(x_t)$.
        % \begin{equation}
        % s_t = \xi(x_t).
        % \end{equation}
        % where $\xi(\cdot)$ denotes task-specific evaluator.
        \STATE Update the exemplar set $\mathcal{E}_{t+1} = \mathcal{E}_t \cup \{(x_t, s_t)\}$.

        \STATE Update the meta-prompt-score set $\mathcal{S}_{t+1} = \mathcal{S}_{t} \cup \{(\left[g(\mathcal{D}_t) \oplus g(\mathcal{I}_t)\right], s_t)\}$, where \( g(\cdot) \) denotes the embedding function and \( \oplus \) denotes concatenation.

        \STATE Update the parameters $\theta$ of the neural network (NN) $\mathcal{M}(g(\cdot); \theta)$ by using the updated $\mathcal{S}_{t+1}$ as the training set to minimize the MSE loss, yielding $\theta_{t+1}$.
        % neural network parameters $\theta_t$ to $\theta_{t+1}$ by optimizing $\mathcal{M}_t(g(\cdot); \theta_t)$ on the meta-prompt score set $\mathcal{S}_{t+1}$ to minimize the MSE loss.

        % \STATE Calculate the cumulative score estimates $\hat{s}_i$ for all arms $i$ based on the predicted scores from $\mathcal{M}_t$:
        \STATE Update the cumulative score estimates $\hat{s}^{(t)}_i$ for all arms $i$ using the predicted scores from $\mathcal{M}(g(\cdot); \theta_{t+1})$:
        % \begin{align}
        $
        \hat{s}_i^{(t+1)} = \hat{s}_i^{(t)} + \mathcal{M}(\left[g(\mathcal{D}_i) \oplus g(\mathcal{I}_i)\right]; \theta_{t+1}), \quad \forall i \in \{1, \ldots, k\}.
        $
        % \end{align}
        % \begin{align}
        % \hat{s}_i^{(t+1)} &= \hat{s}_i^{(t)} + \mathcal{M}_{t+1}(\left[g(\mathcal{D}_i) \oplus g(\mathcal{I}_i)\right]; \theta_{t+1}) \nonumber \\
        % &\quad \forall i \in \{1, \ldots, k\}.
        % \end{align}

        % \STATE Compute the sampling distribution $P_t$ over arms using EXP3:
        \STATE Compute the sampling distribution $P_t$ over all arms:
        % \begin{equation}
        % P_{t}[i] = \frac{\exp(\eta \hat{s}_i^{(t+1)})}{\sum_{j=1}^k \exp(\eta \hat{s}_j^{(t+1)})},
        % \quad \forall i \in \{1, \ldots, k\}
        % \label{eq:exp3}
        % \end{equation}
        \begin{equation}
        \begin{split}
            P_{t}[i] = {}& \frac{\exp(\eta \hat{s}_i^{(t+1)})}
            {\sum_{j=1}^k \exp(\eta \hat{s}_j^{(t+1)})}, \\
            & \forall i \in \{1, \ldots, k\}
        \end{split}
        \label{eq:exp3}
        \end{equation}
        % \STATE Sample a meta-prompt combination $(\mathcal{D}_{t+1}, \mathcal{I}_{t+1})$ from $P_t$:
        \STATE Sample an arm (i.e., the combination of a task description and a meta-instruction) from $P_t$: $(\mathcal{D}_{t+1}, \mathcal{I}_{t+1}) \sim P_t$.
        % \begin{equation}
        % (\mathcal{D}_{t+1}, \mathcal{I}_{t+1}) \sim P_t.
        % \end{equation}
        \STATE Select a sequence of exemplars $\mathcal{E}_{t+1}'$ from $\mathcal{E}_{t+1}$ following a pre-defined heuristic method.
        % \begin{equation}
        % \mathcal{E}_{t+1}' \subseteq \mathcal{E}_{t+1}.
        % \end{equation}
        % \STATE (\emph{optional}) Optimize exemplars using Algorithm~\ref{algo:ES}.
    \ENDFOR
\end{algorithmic}
\caption{\alg}
\label{algo:EXPO}
\end{algorithm}

\textbf{{\textcircled{\scriptsize 1}} LLM-Based Action Selection (lines 3-7 of Algo.~\ref{algo:EXPO}).}
At the beginning of every iteration $t$, we firstly use the current
% newly selected 
task description $\mathcal{D}_t$, meta-instruction $\mathcal{I}_t$ and exemplar sequence $\mathcal{E}_t'$ selected at the end of the last iteration $t-1$ (more details below) to construct a meta-prompt $\mathcal{Q}_t = (\mathcal{D}_t, \mathcal{I}_t, \mathcal{E}_t')$ (line 3). 
Then, we use $\mathcal{Q}_t$ as the input prompt to the LLM $f(\cdot)$ to select the next action $x_t$ and collect its score $s_t$ (lines 4-5). After that, we update the set of exemplars $\mathcal{E}_t$ and the meta-prompt-score set $\mathcal{S}_{t}$ (lines 6-7).

\textbf{{\textcircled{\scriptsize 2}} Score Estimation (lines 8-9).}
In the classical EXP3 algorithm for adversarial bandits, the cumulative sum of \emph{the observed rewards} of every arm is used to construct the arm sampling distribution via an exponential-weight mechanism \cite{lattimore2020bandit}.
However, when the number of arms is excessively large (as in our problem of meta-prompt optimization), the reward observations for many arms are unavailable, so the cumulative sum of \emph{the estimated rewards} is used instead \cite{lattimore2020bandit}.
We therefore estimate the scores of all $k$ arms (i.e., meta-prompts) in the domain and use these estimates to derive the arm sampling distribution for our \alg.
Following recent works showing that a neural network (NN) taking the pre-trained embedding $g(\cdot)$ as input leads to powerful prompt optimization \cite{lin2024prompt,lin2023instinct,wu2024prompt}, we adopt an NN $\mathcal{M}(g(\cdot); \theta)$ for score estimation in our \alg.
Specifically, in every iteration $t$, we use the history of selected meta-prompts and their scores, denoted as $\mathcal{S}_{t+1}$ (line 7 of Algo.~\ref{algo:EXPO}), to train an NN
% train an NN using the past history of selected meta-prompts and their scores 
by minimizing the mean-squared error (MSE) loss (line 8 of Algo.~\ref{algo:EXPO}).
The trained NN with parameters $\theta_{t+1}$ can then be used to estimate the score of every arm (i.e., every combination of task description and meta-instruction) in the domain.
For every arm,
% (i.e., every combination of a task description and a meta-instruction), 
its estimated score is then added to its corresponding \emph{cumulative sum of score estimates} $\hat{s}_i^{(t+1)}$ (line 9 of Algo.~\ref{algo:EXPO}).
Note that each term in the cumulative sum $\hat{s}_i^{(t+1)}$ is the score estimate for arm $i$ from an NN trained using the observation history up to that iteration.

% to predict the score for every combination of the task description and meta-instruction in the domain.

\textbf{{\textcircled{\scriptsize 3}} Randomized Meta-Prompt Selection (lines 10-12).}
After the cumulative sum $\hat{s}_i^{(t+1)}$
% of the score estimates 
of every arm $i$ is updated, 
we 
% use them to 
follow the EXP3 algorithm \cite{lattimore2020bandit} and use the cumulative sums to
construct a distribution following Equation \eqref{eq:exp3}.
Then, we use this distribution to randomly sample the next arm,
% , from which the next arm 
i.e., the next task description $\mathcal{D}_{t+1}$ and meta-instruction $\mathcal{I}_{t+1}$ (line 11 of Algo.~\ref{algo:EXPO}).
\emph{Randomization} is a key principle in adversarial bandits \cite{lattimore2020bandit}, and the randomization involved in our arm selection strategy is crucial for the ability of our \alg~to deal with non-stationary reward observations.
The heuristic for selecting the exemplar sequence $\mathcal{E}'_{t+1}$ (line 12 of Algo.~\ref{algo:EXPO}) is usually specified by the underlying LLM-based algorithm \cite{yang2023large}; we discuss its automated optimization in Sec.~\ref{subsec:expo:es}.

\textbf{Exploitation vs.~Exploration.}
Our \alg~achieves a principled balance between exploitation and exploration.
The powerful pre-trained embedding and NN yield accurate score estimation, so the cumulative score estimate $\hat{s}_i^{(t+1)}$ (line 9 of Algo.~\ref{algo:EXPO}) provides a reliable assessment of the quality of every arm $i$; this ensures that high-scoring arms receive large weights in the sampling distribution $P_t$ (line 10), leading to reliable \emph{exploitation}.
Meanwhile, the inherent randomness in our arm selection strategy ensures that enough \emph{exploration} is performed in the domain of meta-prompts.

\textbf{Batch Action Selection.}
Although our \alg~(Algo.~\ref{algo:EXPO}) selects one action $x_t$ in every iteration $t$, it can be easily generalized to select a batch of actions.
For example, when applying \alg~to improve OPRO \cite{yang2023large} (Sec.~\ref{subsec:exp:opro}), we follow the practice of OPRO to select a batch of $8$ actions per iteration (step 4 of Algo.~\ref{algo:EXPO}) with the LLM temperature set to $1$ for diversity, while selecting the last action with temperature $0$ to obtain a noiseless and reliable score $s_t$ for the meta-prompt $\mathcal{Q}_t$ (line 5 of Algo.~\ref{algo:EXPO}).
\vspace{-1.5mm}
\subsection{\alg~with Exemplar Selection (\alges)}
\label{subsec:expo:es}
\vspace{-1.5mm}
Previous works on LLM-based for sequential decision making often select the sequence of exemplars $\mathcal{E}_{t+1}'$ included in the meta-prompt $\mathcal{Q}_t$ using a fixed pre-defined heuristic
% is selected following a simple and fixed rule 
(line 12 of Algo.~\ref{algo:EXPO}).
For example, OPRO includes the 20 exemplars with the best observed performance in the meta-prompt, arranged in ascending order of performance \cite{yang2023large}; 
the LLM-based MAB method from \cite{krishnamurthy2024can} either includes all exemplars (ordered by their iteration sequence) in the prompt or includes a summarized representation of all exemplars.
However, numerous previous works have reported that both the subset of exemplars and their ordering have significant impacts on the performance of LLM \cite{wu2024prompt}.
Therefore, here we further extend our \alg~(Algo.~\ref{algo:EXPO}) to additionally optimize the sequence of exemplars $\mathcal{E}'_{t+1}$ (i.e., to replace line 12 of Algo.~\ref{algo:EXPO} by an automated method to select $\mathcal{E}'_{t+1}$), 
hence introducing our \alges~algorithm (Algo.~\ref{algo:ES-subset}, App.~\ref{app:expo:es}).
% in order to further boost the performance of LLM-based sequential decision-making.

Since the task description and meta-instruction selected by \alg~change across iterations, the optimization of exemplar sequences is non-stationary as well, so we also optimize it based on the EXP3 algorithm. That is, in every iteration of our \alges~(Algo.~\ref{algo:ES-subset}), we first optimize the task description and meta-instruction (lines 3-11 of Algo.~\ref{algo:EXPO}), and then optimize the exemplar sequence $\mathcal{E}'_{t+1}$ in a similar way.
% The optimization of the exemplar sequence presents a number of unique challenges which require novel algorithmic modifications.

\textbf{Details of \alges~(Algo.~\ref{algo:ES-subset}).}
Specifically, after the task description and meta-instruction are optimized (i.e., after lines 3-11 of Algo.~\ref{algo:EXPO}), we firstly extract the embedding of the exemplar sequence 
% $\mathcal{E}'_{t+1}$ 
$\mathcal{E}'_{t}$ used in this iteration: $g(\mathcal{E}'_{t})$, and add $\left(g(\mathcal{E}'_{t}), s_t\right)$ to the 
% history of exemplar sequence observations 
\emph{exemplar training set}
$\mathcal{T}_{t+1}$ (line 4 of Algo.~\ref{algo:ES-subset}).
Next, the updated dataset $\mathcal{T}_{t+1}$ is used to train an NN with parameters $\theta^{\text{ES}}_{t+1}$ (line 5), which is able to \emph{estimate the score of any exemplar sequence}.
% is used to obtain the updated NN parameters $\theta^{\text{ES}}_{t+1}$ (line 5).
% $\mathcal{M}^{\text{ES}}(g(\cdot),\theta^{\text{ES}})$.
% , which can then be used to predict the score for any exemplar sequence and hence obtain the cumulative score estimates (to be used in the sampling distribution).
Subsequently, we randomly sample $k^{\text{ES}}$ exemplars sequences, each containing $\mathcal{L}$ exemplars, to be used as our \emph{domain of exemplar sequences} (line 8). 
Next, for every candidate exemplar sequence in the domain, we obtain its cumulative score estimate (similar to line 9 of Algo.~\ref{algo:EXPO}).
However, due to \emph{the time-varying nature of the domain of exemplar sequences} (caused by the addition of new exemplars and the random sampling of sequences), we can no longer maintain and incrementally update a cumulative score estimate for every arm.
To this end, we save the trained NN parameters of every iteration; then for each sampled exemplar sequence, we obtain its score estimates from all NNs in the history and use their sum as its \emph{cumulative score estimate} (lines 9-14 of Algo.~\ref{algo:ES-subset}).
Next, the cumulative score estimates for all exemplar sequences are used to compute the sampling distribution, from which the next exemplar sequence $\mathcal{E}'_{t+1}$ is sampled and used to the meta-prompt in the next iteration (lines 15-16).

% \vspace{-5mm}

\section{Experiments}
\label{sec:experiments}
\begin{figure*}[t]
\vspace{-3mm}
     \centering
     \begin{tabular}{ccccc}
         \hspace{-5mm}
        \includegraphics[width=0.2\linewidth]{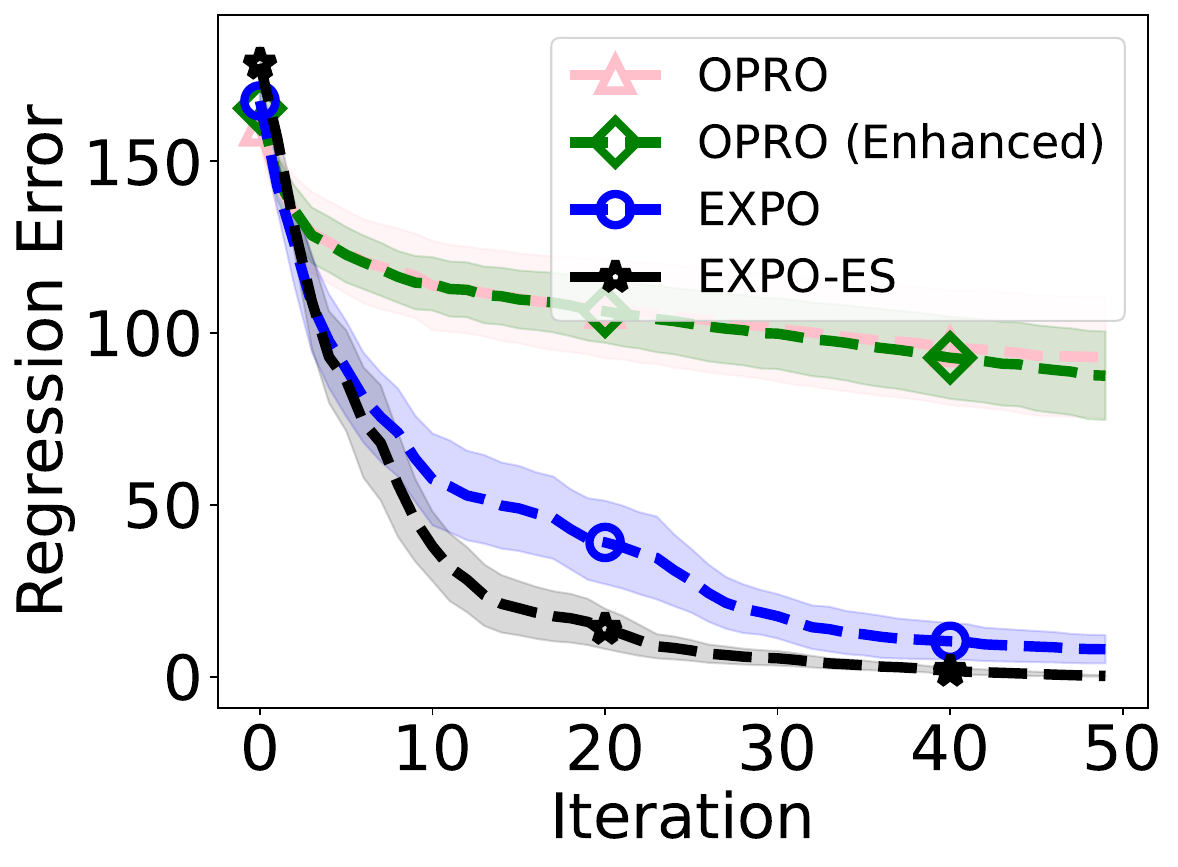} & \hspace{-5mm}
        \includegraphics[width=0.2\linewidth]{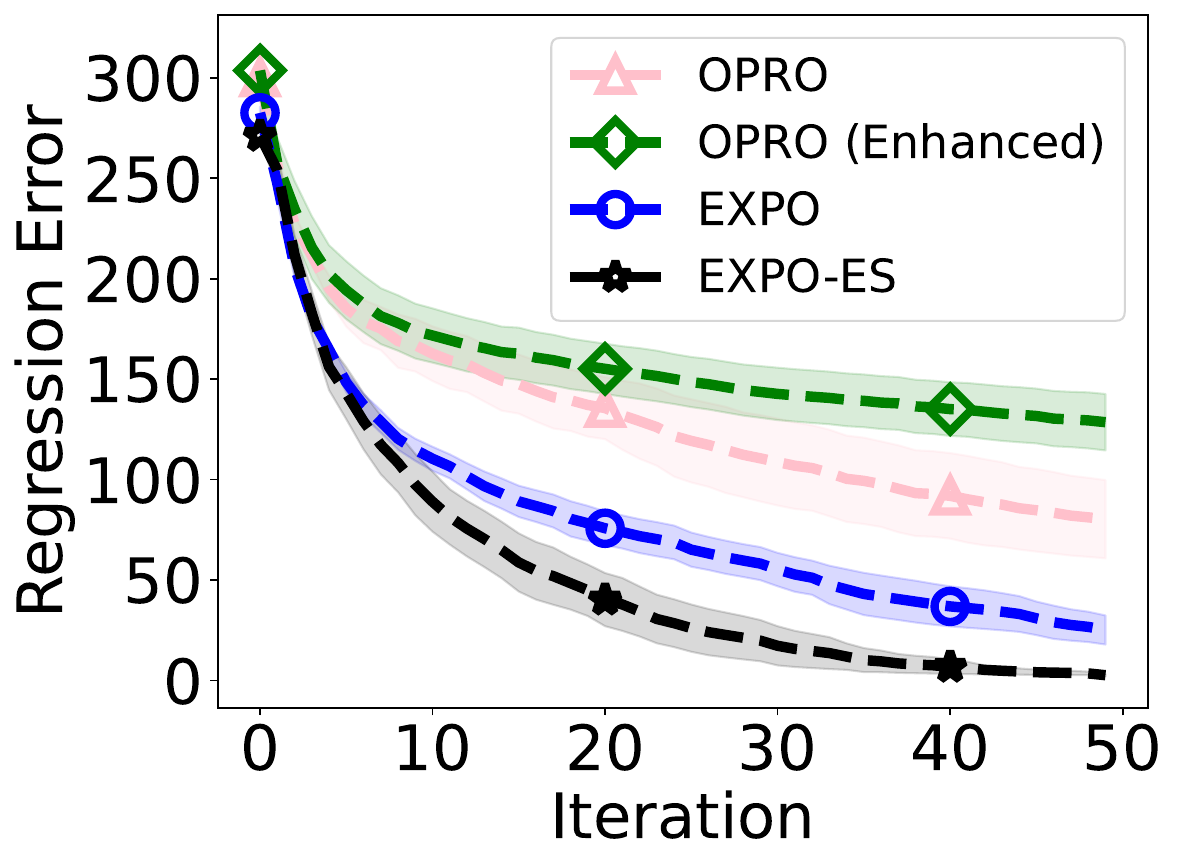} & \hspace{-5mm}
        \includegraphics[width=0.2\linewidth]{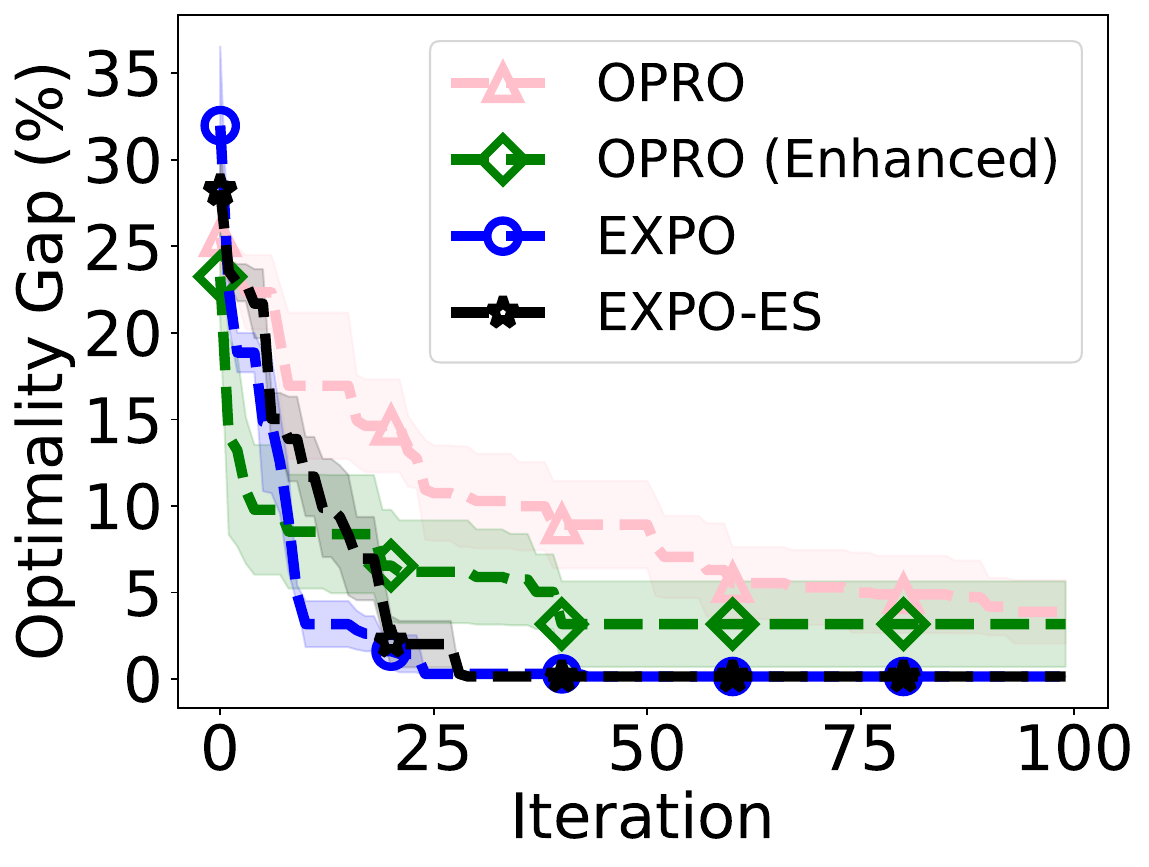} &\hspace{-5mm}
        \includegraphics[width=0.195\linewidth]{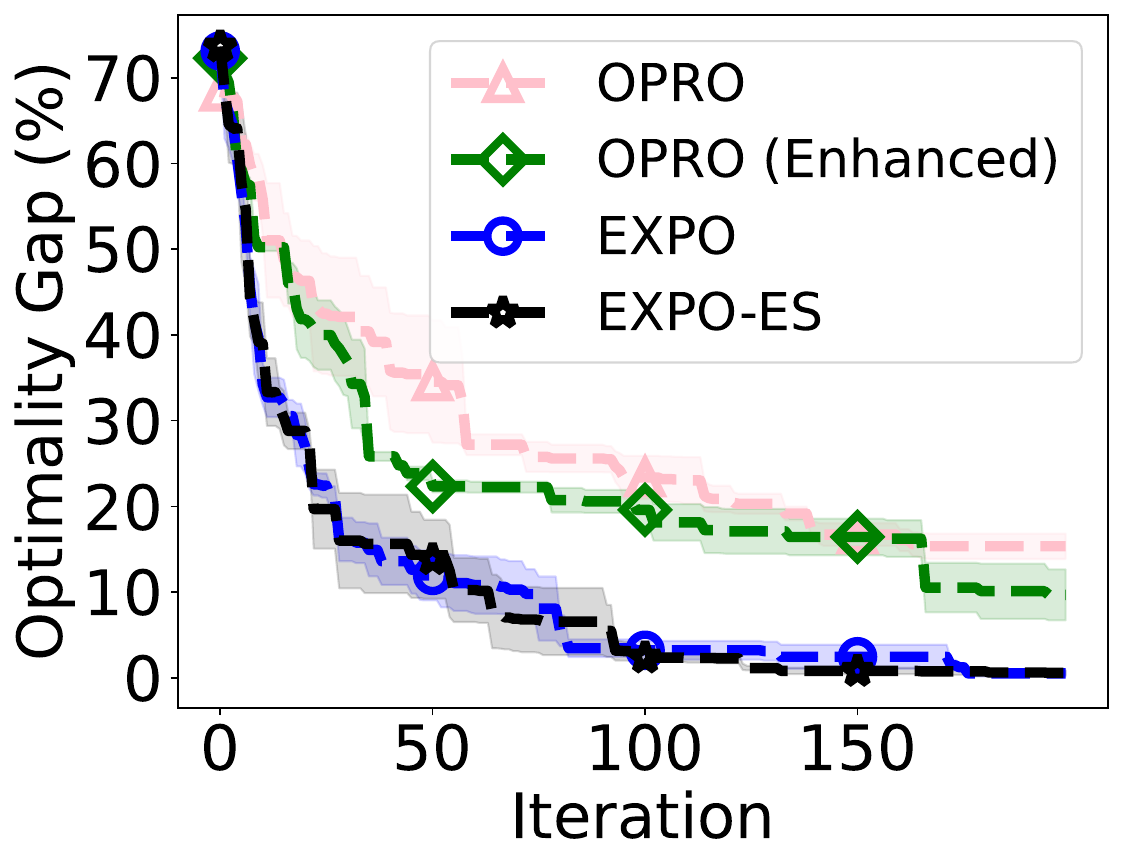} &\hspace{-5mm}
        \includegraphics[width=0.2\linewidth]{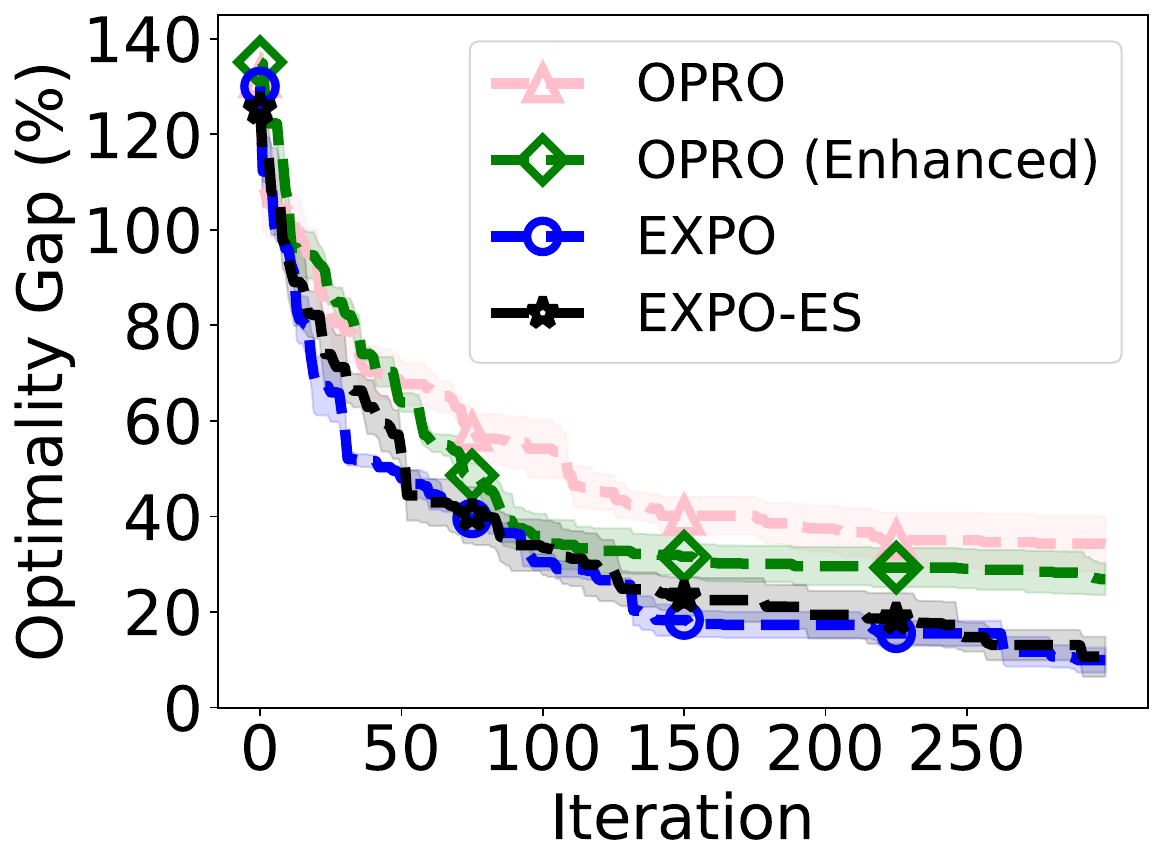}\\
         Linear Regression & Linear Regression & TSP & TSP & TSP\\
         ($w=2, b=30$) & ($w=36, b=-1$) & (10 Nodes) & (15 Nodes) & (20 Nodes)\\
     \end{tabular}
 \vspace{-3mm}
    \caption{
% Visualizations for different scenarios: Linear Regression under two settings in the first row and TSP problems for 10, 15, and 20 nodes in the second row.
Results of different algorithms (mean $\pm$ standard error) in the Linear Regression and TSP task (Sec.~\ref{subsec:exp:opro}). Lower is better. 
% The results are presented as the mean $ \pm \ \text{standard deviation} / \sqrt{n}$, where $n$ is the number of repeated experiments.
    }
\label{fig:linear_tsp_2_main_results}
\vspace{-4mm}
\end{figure*}
We firstly apply our algorithms to improve the performance of OPRO in Linear Regression (LR), traveling salesman problem (TSP) and instruction optimization tasks, adopting the same experimental setting as \cite{yang2023large} (Sec.~\ref{subsec:exp:opro}).
Then, we use our algorithms to enhance the performance of the LLM-based MAB algorithm from \cite{krishnamurthy2024can}.
% adopt the experiments from \cite{krishnamurthy2024can} in which an LLM is used to solve multi-armed bandit problems (Sec.~\ref{subsec:exp:bandits}).

\vspace{-1mm}
% \subsection{Linear Regression and Traveling Salesman Problem}
\subsection{Linear Regression (LR), Traveling Salesman Problem (TSP) and Instruction Optimization}
\label{subsec:exp:opro}
\vspace{-1mm}
% For both tasks here, we adopt GPT-3.5-Turbo as the LLM.

\textbf{Linear Regression (LR).}
In the LR task, our goal is to find the optimal LR coefficients, $w$ and $b$, that best fit a set of given noisy observations. We firstly choose the groundtruth LR coefficients $w_{\text{true}}$ and $b_{\text{true}}$, and use them to generate noisy observations for 50 inputs $x$ which are randomly and uniformly selected within $[-1, 1]$.
% inputs (ranging from 1 to 50): 
Specifically, for each input $x$, we generate its noisy observation as $y = w_{\text{true}} x + b_{\text{true}} + \epsilon$ where $\epsilon$ is a Gaussian noise.
We adopt the two most challenging choices of coefficients from \cite{yang2023large}: (1) $w_{\text{true}}=2,b_{\text{true}}=30$ and (2) $w_{\text{true}}=36,b_{\text{true}}=-1$.
In this task, OPRO aims to find the optimal $w$ and $b$ which minimizes the regression error (i.e., mean squared error).

% We generate 50 random data points uniformly distributed within the range $[-1, 1]$. These data points are generated to perfectly align with the ground truth $w_{\text{true}}, b_{\text{true}}$. OPRO requires warm-starting the LLM with initial exemplars. To initialize the optimization process, we randomly sample 5 $(w, b)$ pairs and their corresponding scores, where $w$ and $b$ are independently sampled from the range $[10, 20]$. These sampled pairs are used as the initial exemplars for the LLM. During each iteration, the GPT-3.5-turbo is prompted 8 times using the same prompt, consisting of 1 inference with a temperature setting of $T=0$ and 7 inferences with a temperature setting of $T=1$. The experiment runs for a total of 50 iterations per configuration, and each configuration is repeated 5 times using the same data point set and initial exemplars.

\textbf{Traveling Salesman Problem (TSP).}
In the classical TSP problem \cite{junger1995traveling}, given a set of $n$ nodes with their coordinates, the objective is to find the shortest route that starts from a given node, traverses all nodes exactly once, and finally returns to the starting node.
Therefore, our goal is to solve a discrete optimization problem in which the input variable is a trajectory and the goal is to minimize the total distance of the trajectory.
We adopt TSP instances with 10, 15, and 20 randomly generated nodes, respectively, which represent increasing levels of difficulty.

% For each TSP instance, the problem is defined by randomly generating $n$ nodes, where the $x$ and $y$ coordinates of each node are sampled uniformly from the range $[-100, 100]$. To initialize the optimization process, we randomly sample 5 different TSP routes along with their corresponding total distances. These routes and their lengths are used as the initial exemplars for the LLM. During each iteration, the GPT-3.5-turbo is prompted 8 times using the same prompt, consisting of 1 inference with a temperature setting of $T=0$ to ensure stability and 7 inferences with a temperature setting of $T=1$ to encourage exploration. The number of iterations is set to 100, 200, and 300 for TSP instances with 10 nodes, 15 nodes, and 20 nodes, respectively. For each TSP configuration, the experiment is repeated 3 times using the same problem point set and initial exemplars.

We adopt GPT-3.5-Turbo as the LLM in both the LR and TSP tasks.
The results for both tasks are shown in Fig.~\ref{fig:linear_tsp_2_main_results}, which plot the regression error (i.e., mean squared error) for the LR tasks and optimality gap (i.e., the difference between the total distance of the discovered route and that of the optimal route) for the TSP tasks (lower is better for both tasks).
Of note, in addition to the standard OPRO (pink curves) \cite{yang2023large}, we have also proposed an enhanced variant of OPRO (green curves) in which we added some further clarifications to the task description 
% of the standard OPRO 
(see App.~\ref{app:sec:enhanced:opro} for more details). 
The enhanced variant consistently improves the performance of the standard OPRO (Fig.~\ref{fig:linear_tsp_2_main_results}).\footnote{As discussed in the last paragraph of Sec.~\ref{subsec:expo}, we have slightly modified OPRO to select the last action in the batch using a temperature of $0$. We empirically show that this leads to comparable performance with the original OPRO which uses a temperature of $1$ to choose all $8$ actions (see Fig.~\ref{fig:full_results} in App.~\ref{exp:app:opro:full:results}).}
% {\color{red}in which we added a sentence to the task description to explain......}
More importantly, the results in Fig.~\ref{fig:linear_tsp_2_main_results} show that \emph{\textbf{in all tasks, our \alg~algorithm (blue curves) significantly and consistently outperforms OPRO}}, including both standard OPRO and its enhanced variant.
This demonstrates that our meta-prompt optimization approach, grounded in adversarial bandits, leads to more efficient (i.e., faster convergence) and more effective (i.e., improved final performance) LLM-based sequential decision making.
\begin{figure*}[t]
% \onecolumn
\begin{minipage}[t]{0.32\textwidth}
\begin{mdframed}[linewidth=0.9pt]  % adjust linewidth as you desire
\scriptsize  % adjust text size as required
% \footnotesize
\centerline{{\normalsize OPRO}}
    {\color{purple}Now you will help me minimize a function with two input variables \(w, b\). I have some \((w, b)\) pairs and the function values at those points. The pairs are arranged in descending order based on their function values, where lower values are better.} \\ \\
    \{EXEMPLARS\}
    \\\\
    {\color{blue}Give me a new \((w, b)\) pair that is different from all pairs above, and has a function value lower than any of the above. Do not write code. The output must end with a pair \([w, b]\), where \(w\) and \(b\) are numerical values.}
    % Output:
\end{mdframed}
\end{minipage}
\hfill
\begin{minipage}[t]{0.67\textwidth}
\begin{mdframed}[linewidth=0.9pt]  % adjust linewidth as you desire
    \scriptsize  % adjust text size as required
    % \footnotesize
\centerline{{\normalsize \alg}}
    {\color{purple}We will collaborate to optimize a function involving two parameters, \textbackslash(w\textbackslash) and \textbackslash(b\textbackslash). I possess a set of data points, each consisting of \textbackslash((w, b)\textbackslash) pairs and their corresponding function values. These pairs are systematically organized in reverse order, starting from the greatest to the smallest function values. Essentially, the lower the function value, the more optimal or preferable the pair. Consequently, our goal is to identify and analyze the \textbackslash((w, b)\textbackslash) pair that manifests the lowest function value, as this represents the perspective of optimum efficacy.} \\ \\
    \{EXEMPLARS\}
    \\\\
    {\color{blue}To enhance the quality and expand on the existing instructions, follow these improved guidelines vis-à-vis designing a new and distinctive numerical pair: ensure the selected \((w, b)\) combination diverges from prior examples and secures a function output lower than preceding values. Key details on methodology or calculations are not required—just ensure clarity in presenting a returned value that closes with the specific format \([w, b]\), where both \(w\) and \(b\) are distinct numerical figures.} 
\end{mdframed}
\end{minipage}
\vspace{-2.5mm}
\caption{
The {\color{purple}task description} and {\color{blue}meta-instruction} used by OPRO (left) and optimized by our \alg~(right) in a Linear Regression task.
}
\label{fig:example:descriptions:lr}
\end{figure*}

\begin{figure*}[t]
% \vspace{3mm}
\centering
% \includegraphics[width=0.25\linewidth]{figures/BSSND_easy_Cumulative_Regret.pdf}
% \includegraphics[width=0.25\linewidth]{figures/BSSCD_easy_Cumulative_Regret.pdf}
% % \vspace{5mm}
% \includegraphics[width=0.25\linewidth]{figures/BSSND_hard_Cumulative_Regret.pdf}
% \includegraphics[width=0.25\linewidth]{figures/BSSCD_hard_Cumulative_Regret.pdf}
% % \vspace{-3mm}
% \makebox[0.25\linewidth]{BSSND (easy)}
% \makebox[0.25\linewidth]{BSSCD (easy)}
% \makebox[0.25\linewidth]{BSSND (hard)}
% \makebox[0.25\linewidth]{BSSCD (hard)}
     \begin{tabular}{cccc}
     \hspace{-7mm}
         \includegraphics[width=0.24\linewidth]{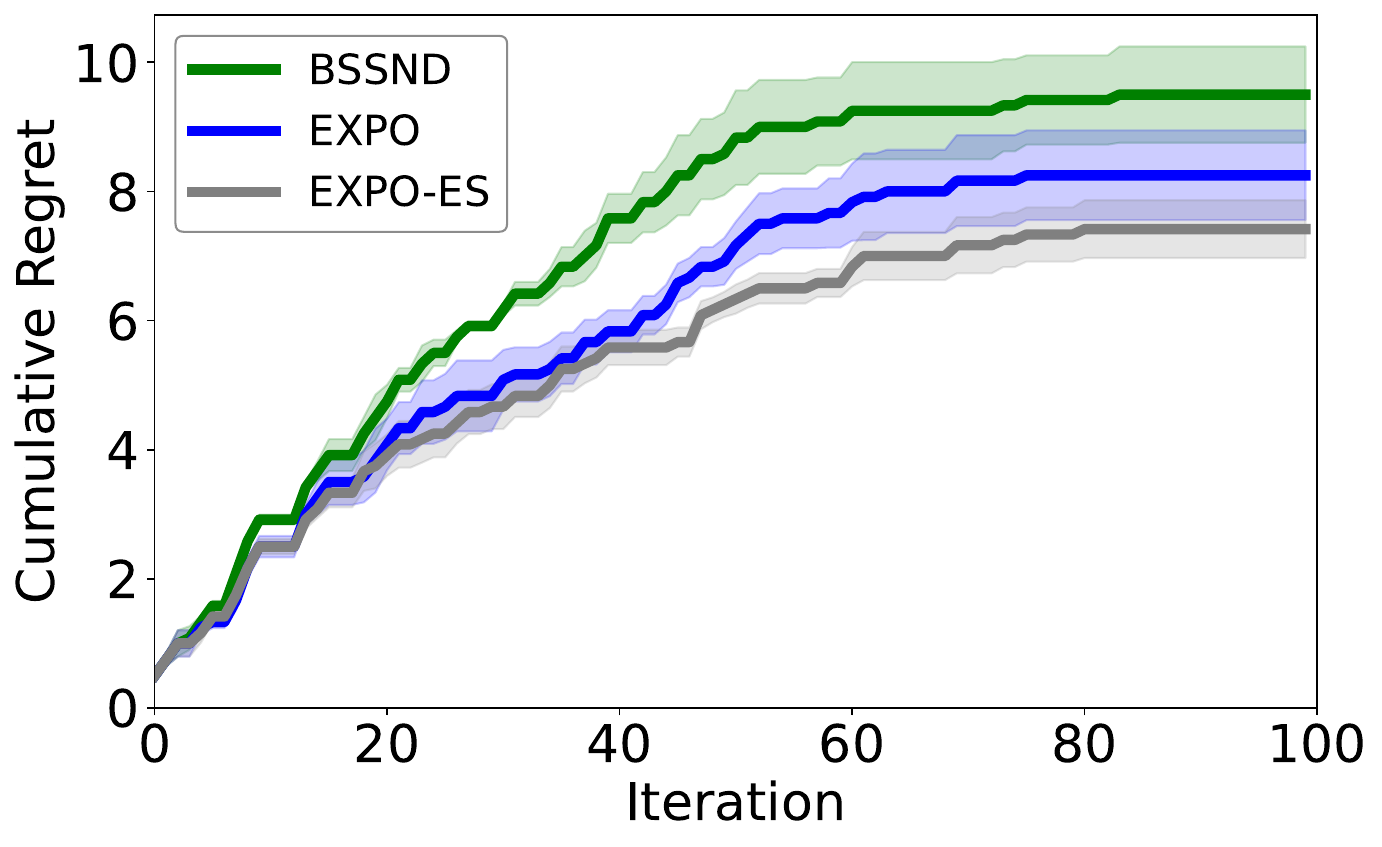} & \hspace{-5mm} 
         \includegraphics[width=0.24\linewidth]{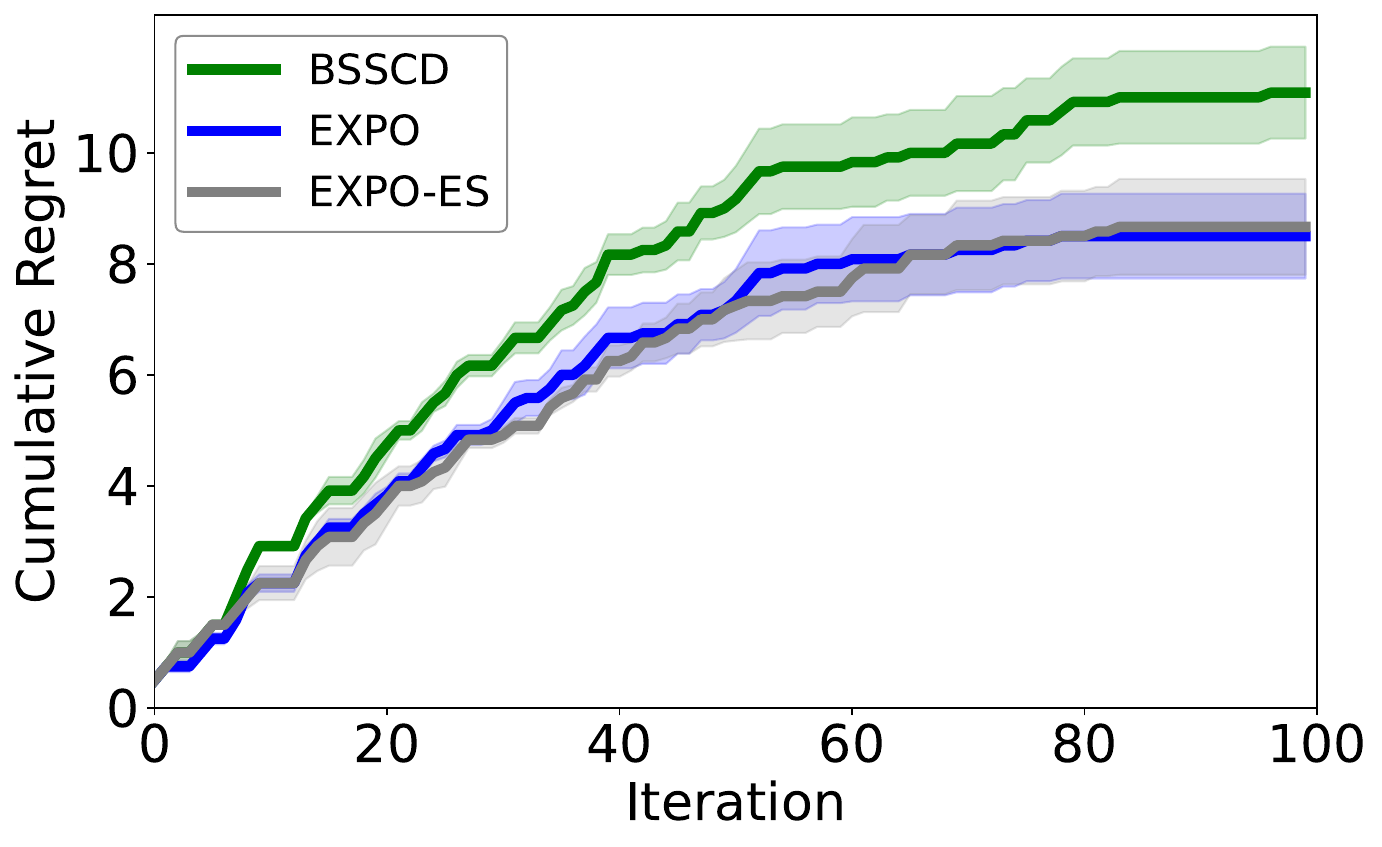}& \hspace{-5mm}
         \includegraphics[width=0.24\linewidth]{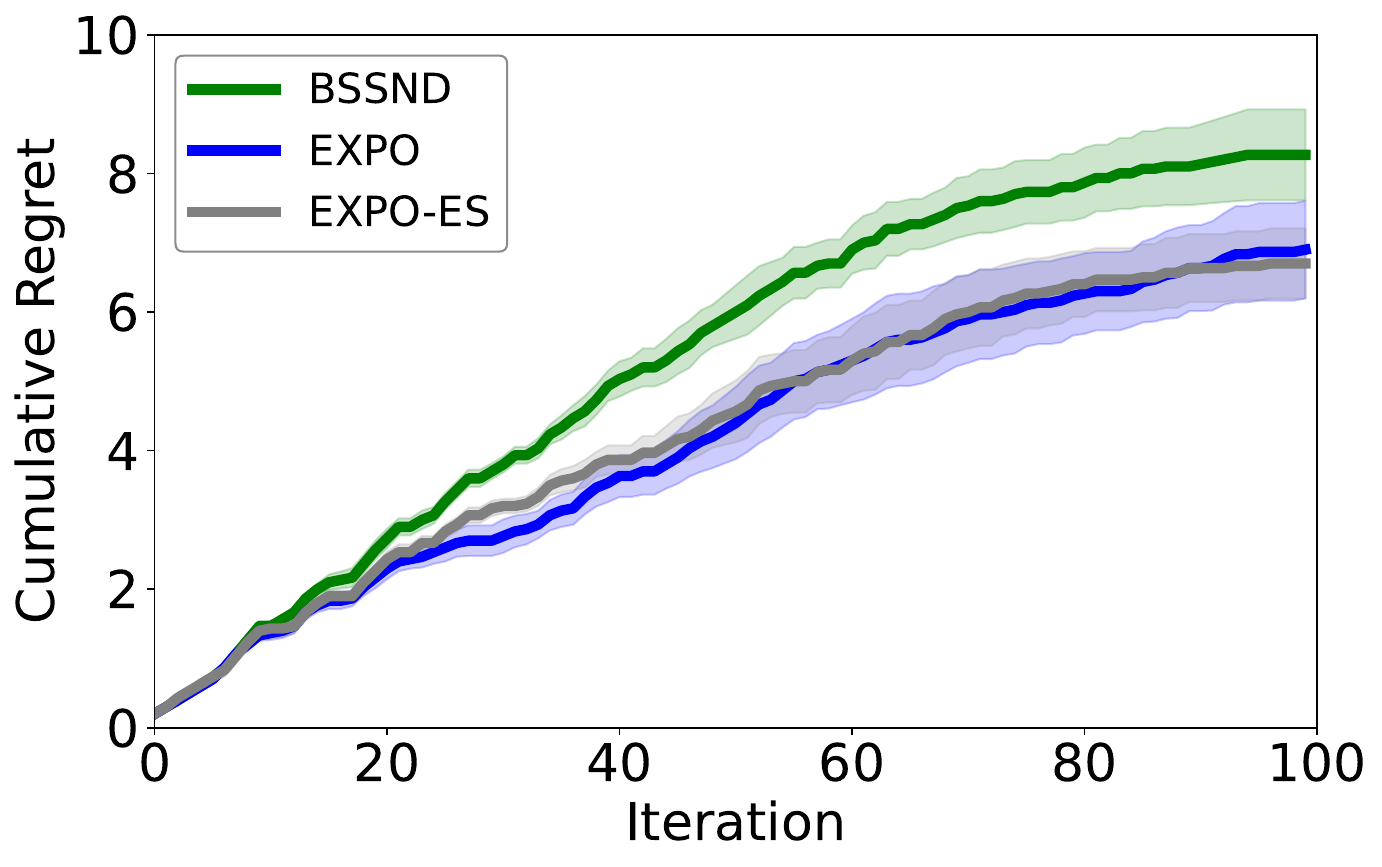}& \hspace{-5mm}
         \includegraphics[width=0.24\linewidth]{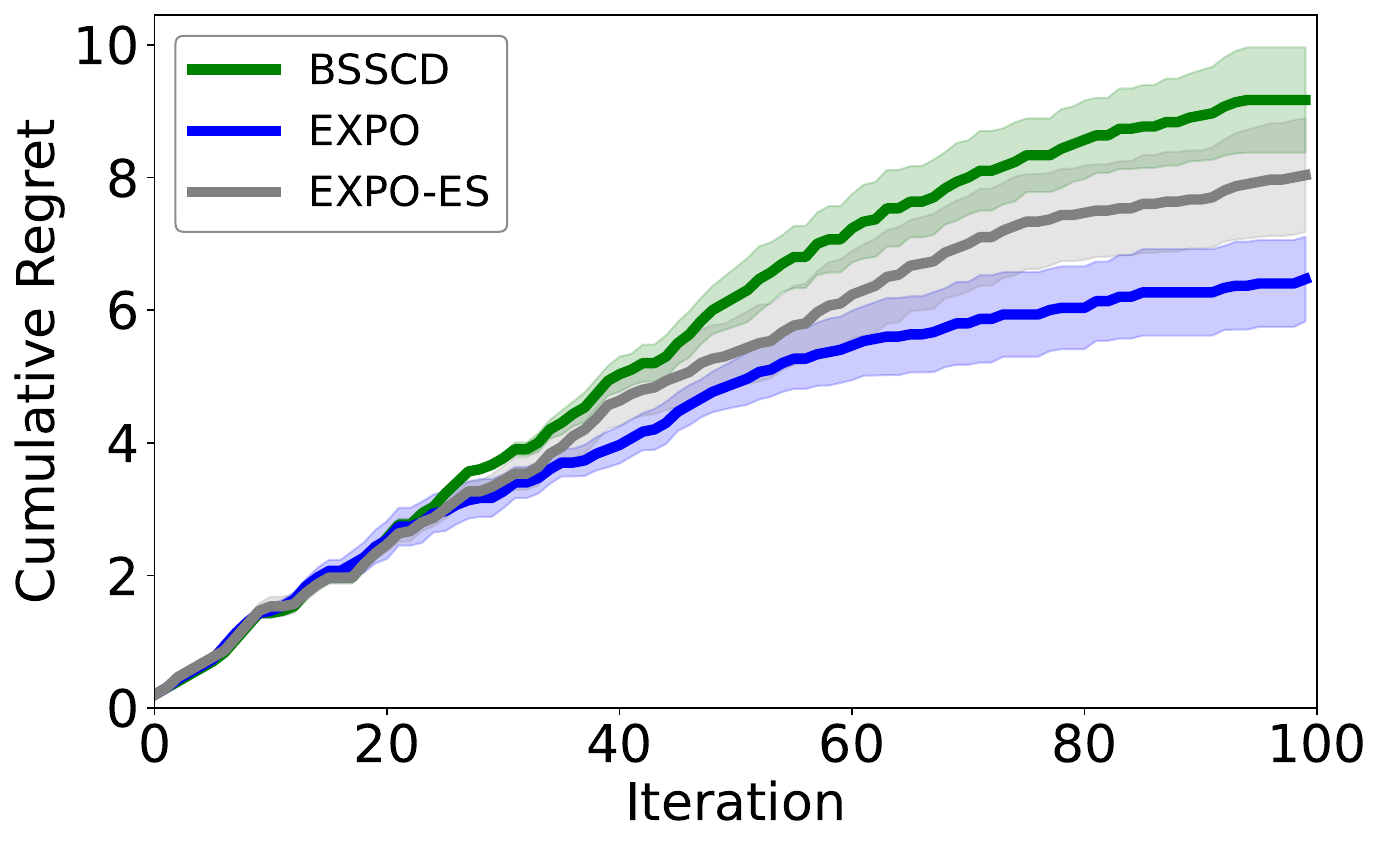}\\
         {\hspace{-2mm} BSSND (easy)} & {\hspace{-5mm} BSSCD (easy)} & {\hspace{-6mm} BSSND (hard)} & {\hspace{-2mm} BSSCD (hard)}
     \end{tabular}
\vspace{-2mm}
\caption{
Cumulative regret 
% of different algorithms 
in the LLM-based MAB experiments (Sec.~\ref{subsec:exp:bandits}). Lower is better.
}
\label{fig:cumulative_regret_BSSND_BSSCD}
\vspace{-3mm}
\end{figure*}

Meanwhile, our \alges~algorithm, which is additionally equipped with automated exemplar selection, considerably improves the performance of \alg~in the LR tasks yet performs on par with \alg~in the TSP tasks.
This is likely because the exemplars play a more important role in the LR tasks than the TSP tasks. 
Specifically, in LR, \emph{the input-output exemplars provide important information for identifying the optimal LR coefficients} \cite{wu2024prompt}. Therefore, selecting better exemplars (via our \alges) brings significant performance boost. 
On the other hand, in the TSP tasks, due to the challenging nature of the tasks, it is difficult for the LLM to infer crucial and useful information from the exemplars. 
Therefore, the other components in the meta-prompt (i.e., the task description and meta-instruction) provide more useful information 
% for the decision-making of the LLM 
in the TSP tasks.
As a result, \emph{selecting better exemplars does not lead to noticeable performance gains in the TSP tasks}.
Fig.~\ref{fig:example:descriptions:lr} provides an illustration of the original task description and meta-instruction used by OPRO and those discovered by our \alg~algorithm for the LR tasks, whereas the corresponding meta-prompts for the TSP tasks are displayed in Fig.~\ref{fig:example:descriptions:tsp} in App.~\ref{app:subsec:more:illustration:meta-prompt}.

\begin{wrapfigure}{r}{0.33\textwidth}  % 'r' for right, 'l' for left
\vspace{-4mm}
    \centering
    \includegraphics[width=\linewidth]{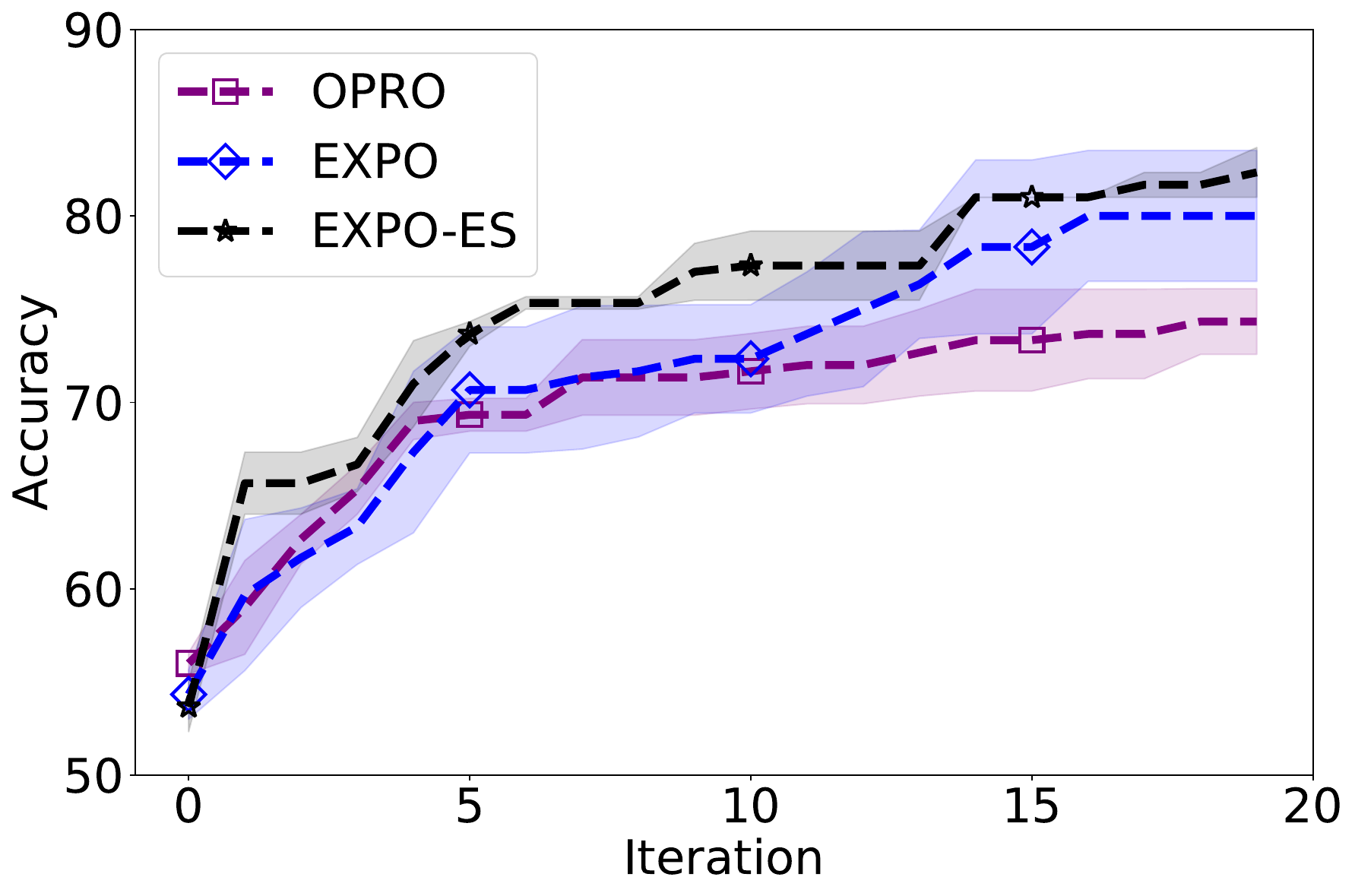}
    \vspace{-6mm}
    \caption{
    Instruction optimization task (higher is better).
    % The performance of our \algts~algorithm in stochastic MAB tasks with different temperatures.
    }
    \vspace{-4mm}
    \label{fig:instruction:opt}
\end{wrapfigure}
\textbf{Instruction Optimization.}
We have also adopted our \alg~to improve the performance of OPRO in instruction optimization tasks using the GSM8K dataset \cite{cobbe2021gsm8k}.
We adopt GPT-4-Turbo as the optimizer LLM to sequentially propose different instructions, and use Qwen-2.5-7B as the scorer LLM to evaluate the effectiveness of these instructions (more details in App.~\ref{app:subsubsec:task:setting:opro}).
% , which aim to optimize the instruction given to an LLM.
The results in Fig.~\ref{fig:instruction:opt} show that our \alg~also considerably outperforms OPRO in this challenging task.
Moreover, \alges~further improves the performance of \alg,  which is likely because the exemplars (i.e., instructions and their scores) provide valuable information for instruction optimization in this task.

\subsection{LLM-Based Multi-Armed Bandits (MAB)}
\label{subsec:exp:bandits}
The work of \cite{krishnamurthy2024can} has used an LLM to sequentially select the arms/actions in MAB and proposed methods to manually design the meta-prompt.
Their prompt design consists of 5 components with each having 2 possible choices, which gives rise to a total of $2^5=32$ possible prompts.
Here we show that our algorithms can be used to automatically optimize their manually designed prompts to further enhance their performance.
Specifically, we adopt 2 of their prompt designs: BSSND and BSSCD,
% (refer to App.~\ref{subsec:exp:app:more:details:bandit} for more details on these prompt designs), 
and apply our \alg~and \alges~algorithms to optimize the important components in these prompt designs.
% some components in these prompt.
Following \cite{krishnamurthy2024can}, we use two MAB instances: \emph{easy} and \emph{hard}.
We adopt GPT-4-Turbo as the LLM here.
More details on the experimental design are deferred to App.~\ref{subsec:exp:app:more:details:bandit}.
The results for the 4 experimental settings (i.e., 2 prompt designs $\times$ 2 MAB instances) are shown in Fig.~\ref{fig:cumulative_regret_BSSND_BSSCD}, which demonstrate that our \alg~and \alges~algorithms are able to significantly reduce the cumulative regret of MAB in this task across different prompt desings and MAB instances.
We illustrate the comparison between the original meta-prompt and the one optimized by our \alg~in Figs.~\ref{fig:example:descriptions:MABbssnd} and~\ref{fig:example:descriptions:MAB} in App.~\ref{app:subsec:more:illustration:meta-prompt}.

% The experiments are conducted for both the BSSND and BSSCD prompts under two predefined difficulty settings: \textit{hard} and \textit{easy}. For the \textit{hard} setting, the MAB instance consists of $K=5$ arms, where the best arm has a mean reward of $\mu^\star = 0.5 + \Delta / 2$ with $\Delta = 0.2$, and all other arms have a mean reward of $\mu = 0.5 - \Delta / 2$. For the \textit{easy} setting, the MAB instance consists of $K=4$ arms with a larger gap $\Delta = 0.5$ between the best arm and the suboptimal arm. The work of \cite{krishnamurthy2024can} has reported that GPT-3.5 models encounter exploration failures in MAB tasks, making them unsuitable as agents for solving such problems. In contrast, GPT-4 demonstrates the capability to effectively handle the exploration-exploitation trade-off inherent in MAB settings. Therefore, we adopt GPT-4-turbo as the LLM agent for this experiment. In each iteration, GPT-4-turbo is prompted with a temperature setting of $T=0$ to generate one inference result. Each experimental configuration runs for 100 iterations. For each MAB configuration, the experiment is repeated 6 times.

\vspace{-2mm}
\section{Ablation Study}
\label{sec:ablation}
\begin{figure*}[h]
\centering
\begin{minipage}{0.48\textwidth}
% \vspace{-0.8mm}
\begin{tabular}{cc}
\hspace{-2mm}
    \includegraphics[width=0.49\linewidth]{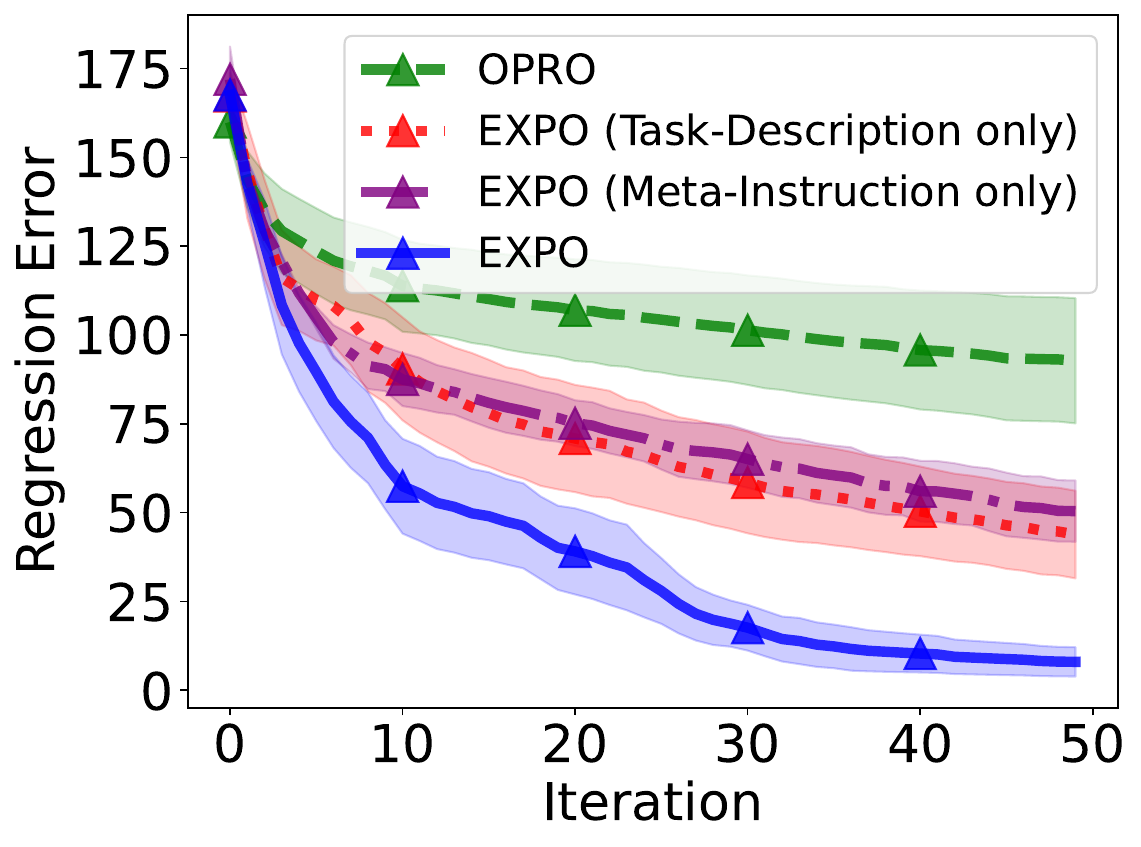} & \hspace{-3mm}
    \includegraphics[width=0.49\linewidth]{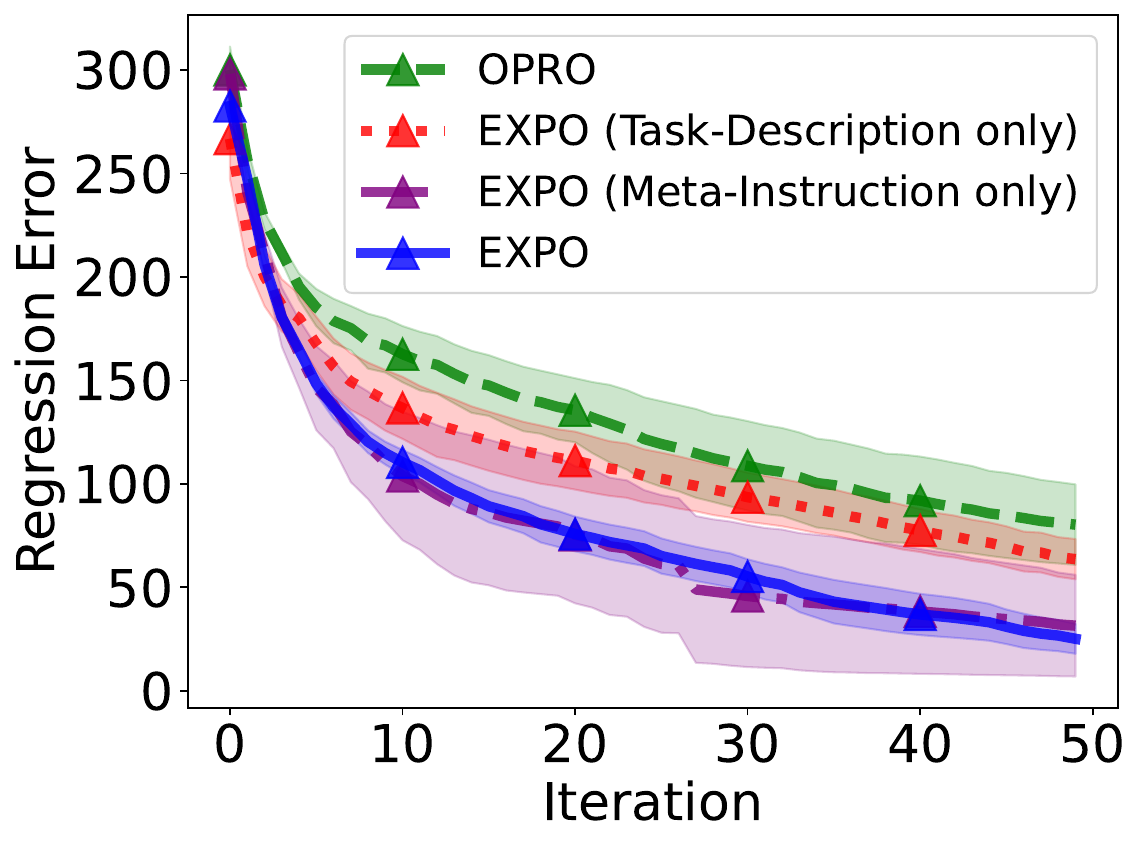} \\
    {\small \makecell{Linear Regression \\ ($w=2$, \ $b=30$)}} &
    {\small \makecell{Linear Regression \\ ($w=36$, \ $b=-1$)}} \\
\end{tabular}
\vspace{-2mm}
\caption{
Results of \alg~when \emph{only} optimizing the task description or the meta-instruction.
% Convergence curves for Linear Regression tasks using Task Descriptions or Meta-Instructions optimization Only. The results are shown for $w=2$, $b=30$ (left) and $w=36$, $b=-1$ (right).
}
\label{fig:task_description_meta_instructions_ablation}
\end{minipage}
% \vspace{-3mm}
% \end{figure}
% \begin{figure}[h]
% \vspace{-3mm}
\hfill
\begin{minipage}{0.48\textwidth}
\vspace{-1.5mm}
\centering
\begin{tabular}{cc}
\hspace{-2mm}
    \includegraphics[width=0.49\linewidth]{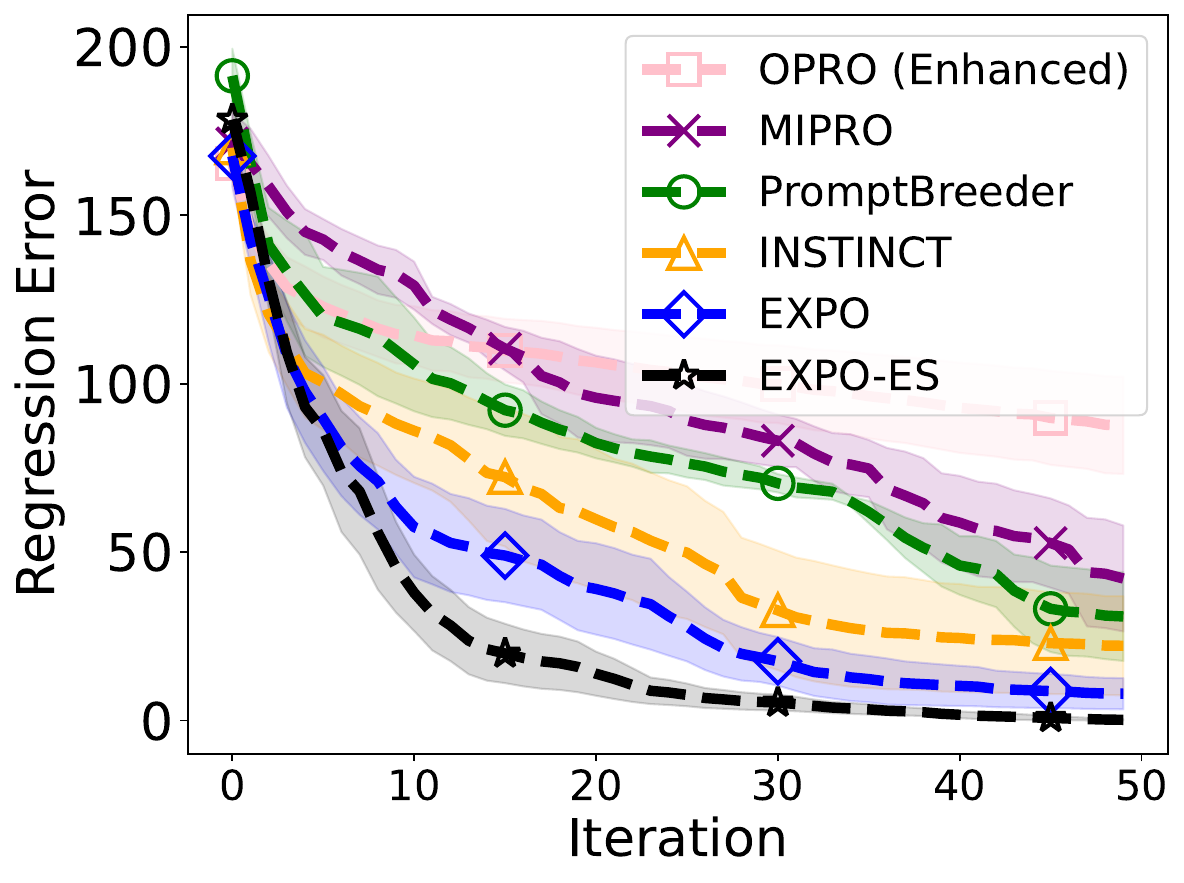} & \hspace{-3mm}
    \includegraphics[width=0.49\linewidth]{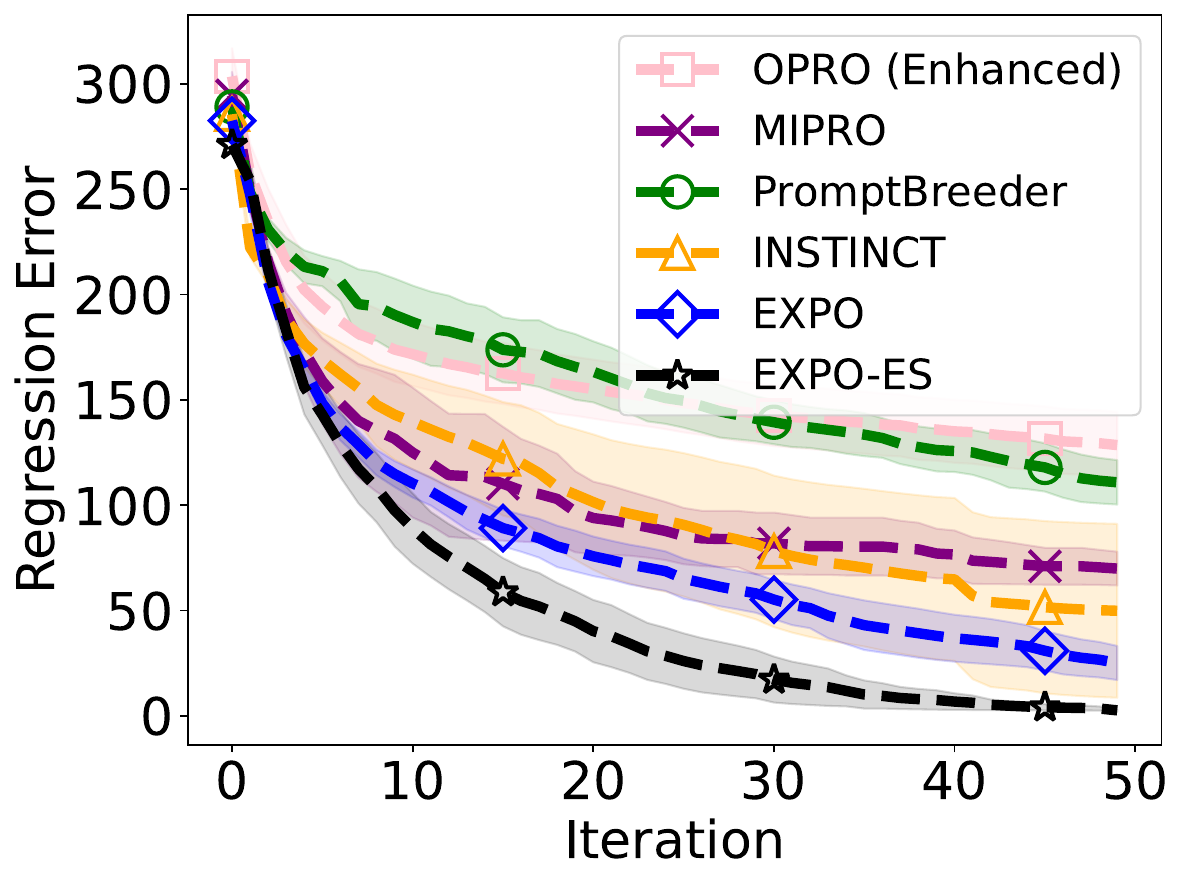}
    \\
    {\small \makecell{Linear Regression \\ ($w=2$, \ $b=30$)}} &
    {\small \makecell{Linear Regression \\ ($w=36$, \ $b=-1$)}} \\
\end{tabular}
\vspace{-2.5mm}
\caption{
Comparison of our \alg~with general prompt optimization methods
% NeuralUCB (i.e., a representative stochastic MAB algorithm) 
in the LR tasks.
% Comparison of our \alg~with NeuralUCB (i.e., a representative stochastic MAB algorithm) in the LR tasks.
% Ablation study results using NeuralUCB. The first row shows results for Linear Regression under two settings. The second row presents the TSP tasks for 10 and 15 nodes, respectively, and the third row provides results for the TSP with 20 nodes.
}
\label{fig:ablation_neuralucb:lr}
\vspace{-2mm}
\end{minipage}
\end{figure*}

\vspace{-2mm}
% \subsection{Only Optimizing Task Description or Meta-Instruction}
% \label{subsec:ablation:separate:description:instruction}
\textbf{Only Optimizing Task Description or Meta-Instruction.}
% As described in Sec.~\ref{subsec:expo}, in 
Our \alg~jointly optimize the task description $\mathcal{D}$ and the meta-instruction $\mathcal{I}$.
Here we evaluate the performance of optimizing either $\mathcal{D}$ or $\mathcal{I}$ alone. 
The results in Fig.~\ref{fig:task_description_meta_instructions_ablation} show that jointly optimizing them indeed leads to significantly better performance.
However, optimizing these components alone still consistently outperforms OPRO.
% \begin{figure}[h]
% \centering
% \begin{tabular}{cc}
%     \includegraphics[width=0.45\linewidth]{figures/Task_Description_or_Meta_Instructions_Only_2_30.pdf} &
%     \includegraphics[width=0.45\linewidth]{figures/Task_Description_or_Meta_Instructions_Only_36_neg1.pdf} \\
%     {\small $w=2$, \ $b=30$} &
%     {\small $w=36$, \ $b=-1$} \\
% \end{tabular}
% \vspace{-2.5mm}
% \caption{
% Results of our \alg~when only optimizing the task description or the meta-instruction.
% % Convergence curves for Linear Regression tasks using Task Descriptions or Meta-Instructions optimization Only. The results are shown for $w=2$, $b=30$ (left) and $w=36$, $b=-1$ (right).
% }
% \label{fig:task_description_meta_instructions_ablation}
% \vspace{-3mm}
% \end{figure}

\begin{figure*}[h]
% \vspace{3mm}
\centering
     \begin{tabular}{cccc}
     \hspace{-7mm}
         \includegraphics[width=0.24\linewidth]{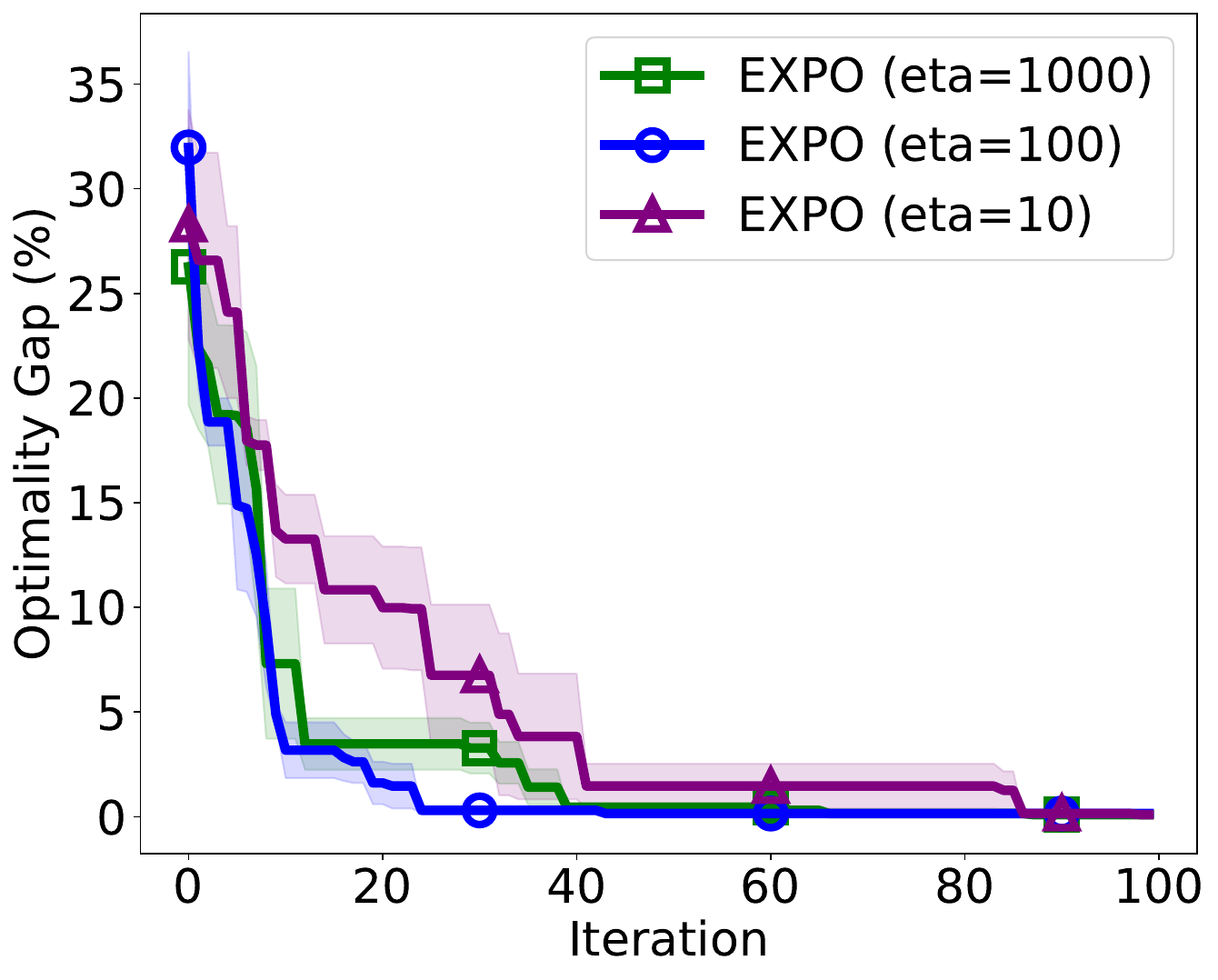} & \hspace{-5mm} 
         \includegraphics[width=0.24\linewidth]{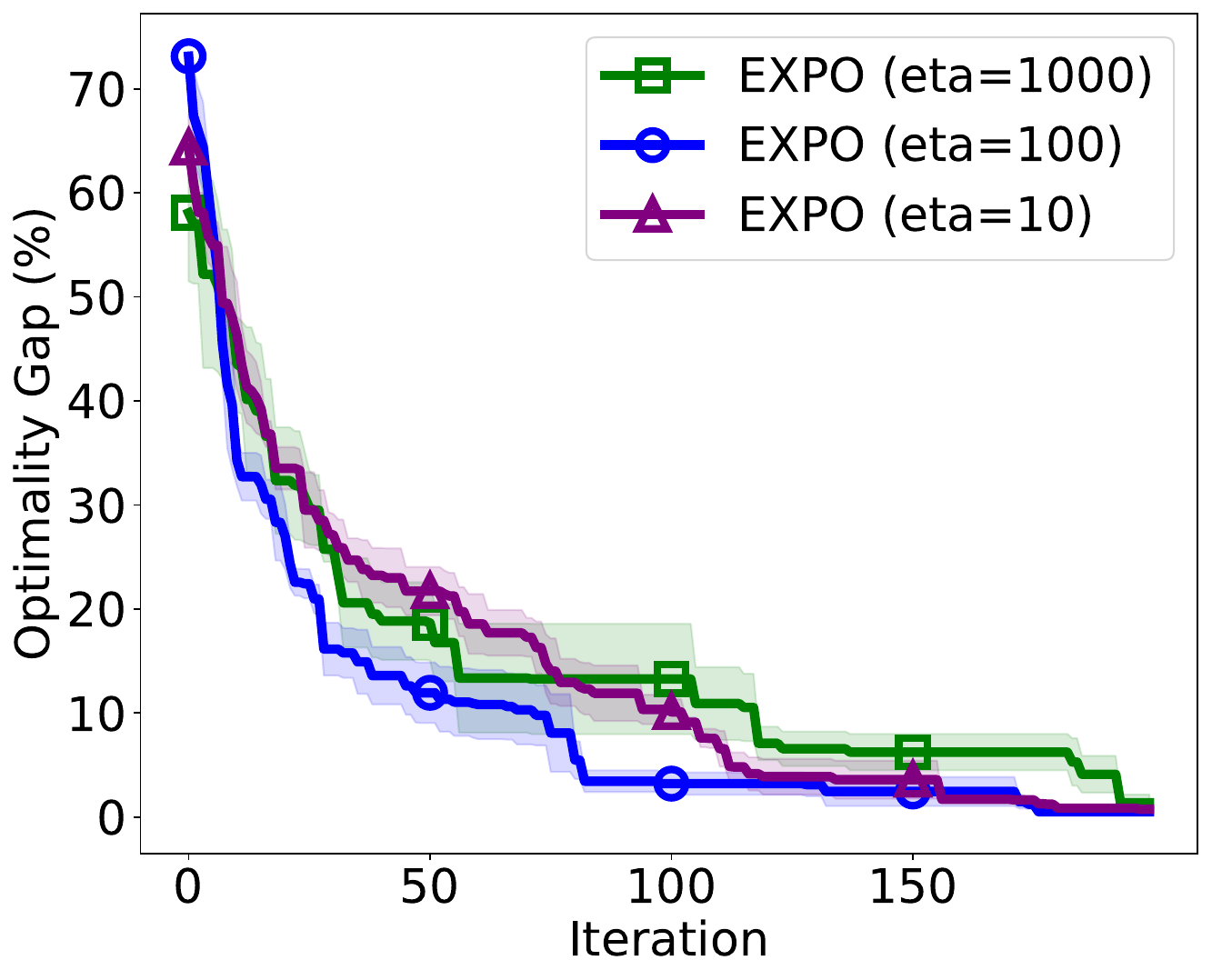}& \hspace{-5mm}
         \includegraphics[width=0.24\linewidth]{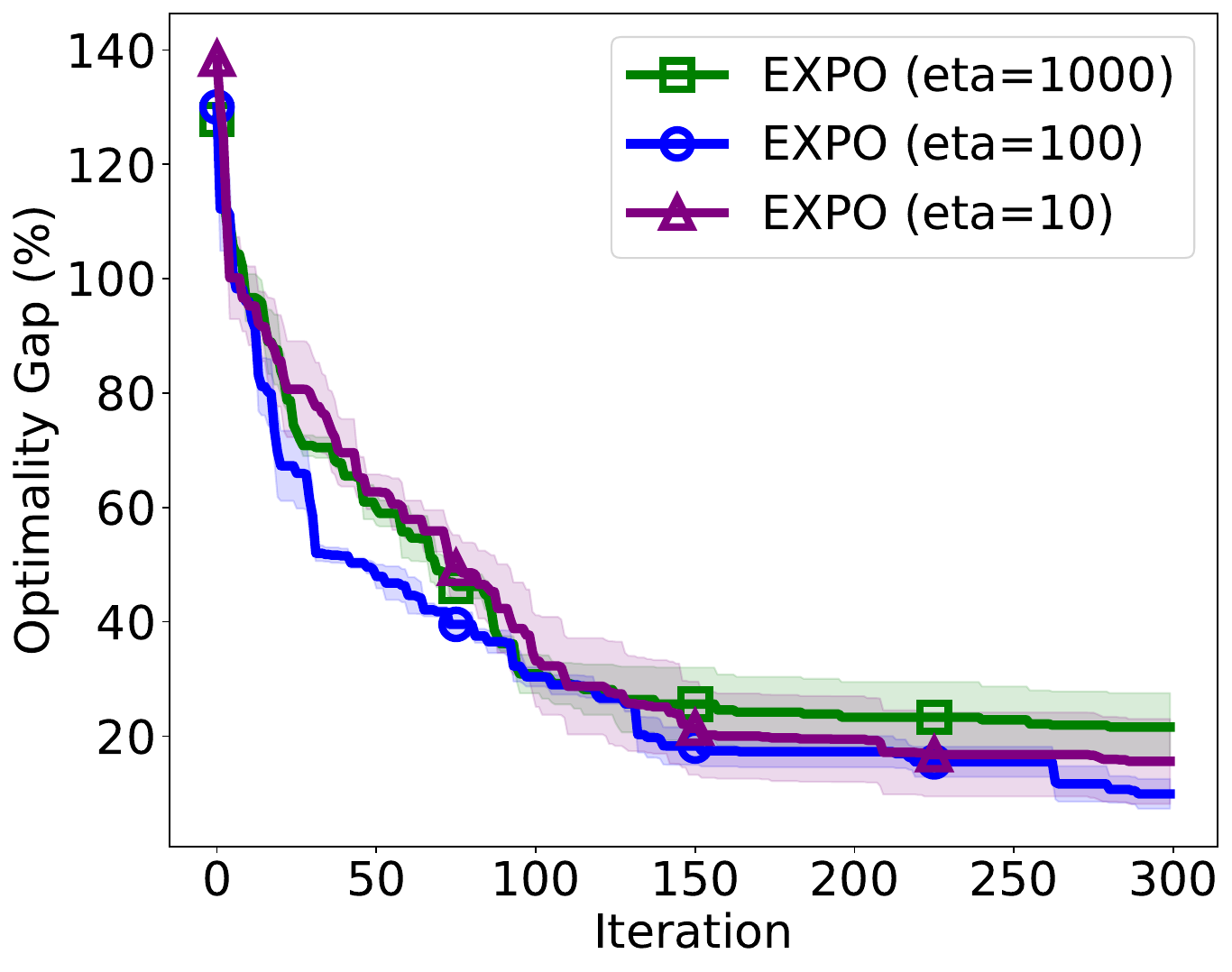}& \hspace{-8mm}
         \includegraphics[width=0.24\linewidth]{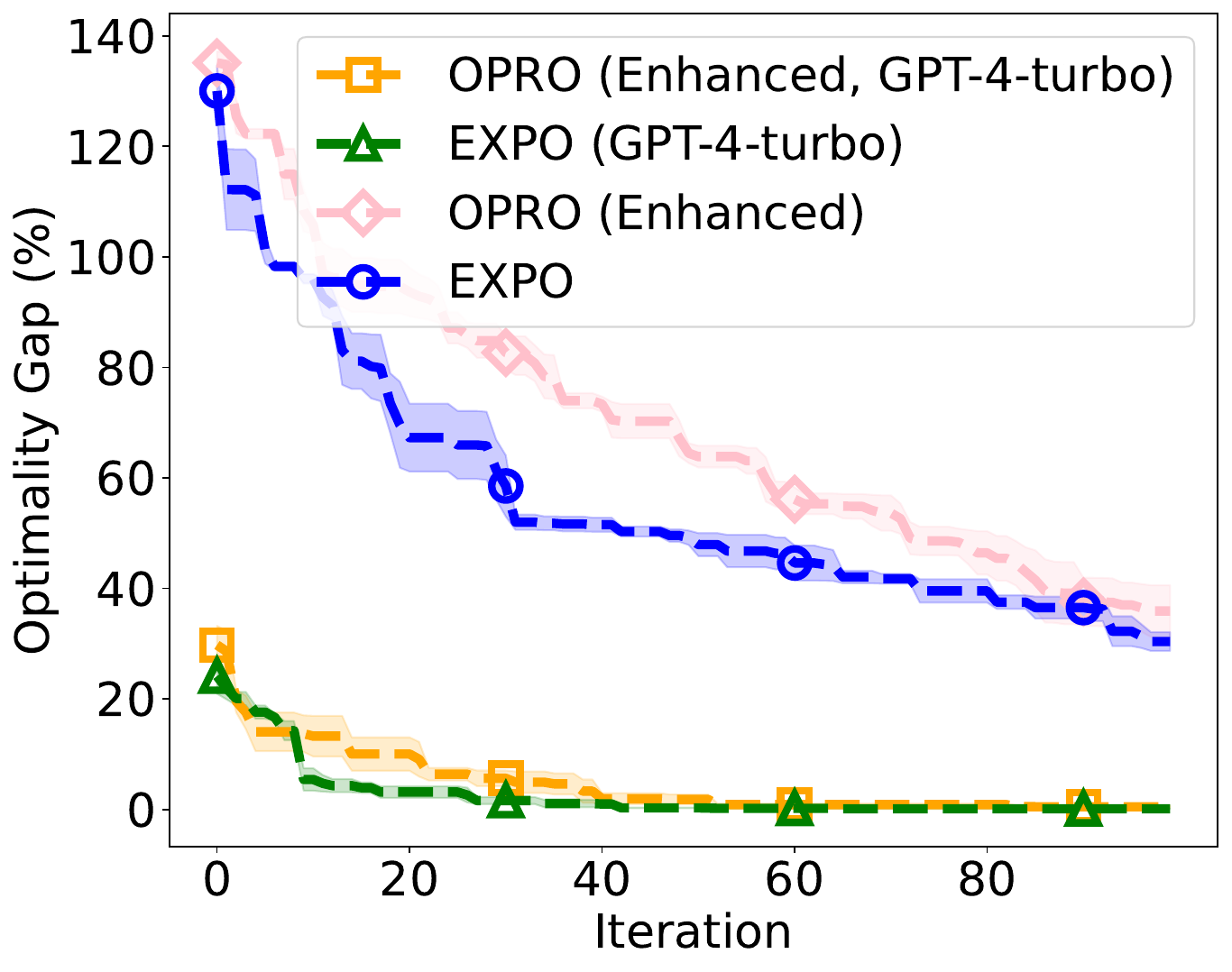}\\
         {\small\hspace{-2mm} TSP (10 nodes)} & {\small\hspace{-5mm} TSP (15 nodes)} & {\small\hspace{-5mm} TSP (20 nodes)} & {\small\hspace{-4mm} TSP (20 Nodes, GPT-4-Turbo)}
     \end{tabular}
\vspace{-2mm}
\caption{
    First three figures: ablation study on impact of exploration parameter $\eta$.
    Rightmost figure: results using GPT-4-Turbo.
}
% \label{fig:cumulative_regret_BSSND_BSSCD}
\label{fig:TSP_ablation:eta}
\vspace{-1mm}
\end{figure*}

% \textbf{Comparison with Stochastic MAB Algorithms: Upper Confidence Bound.}
\textbf{Comparison with General Prompt Optimization Methods.}
% Classical stochastic MAB algorithms, such as those based on upper confidence bound (UCB), have been applied to prompt optimization in a number of recent works \cite{lin2024prompt,lin2023instinct,wu2024prompt} and yielded strong performance.
% However, 
As discussed in Sec.~\ref{sec:intro}, in the problem of meta-prompt optimization for LLM-based agents, \emph{\textbf{the non-stationary reward observations render the previous general prompt optimization methods unsuitable}}.
Here we verify this by comparing our \alg~with a variety of general prompt optimization methods, including the INSTINCT algorithm based on stochastic MAB \cite{lin2023instinct}, PromptBreeder \cite{fernando2023promptbreeder} and MIPRO \cite{opsahl2024optimizing}.
% \emph{NeuralUCB} algorithm adopted by \cite{lin2023instinct,wu2024prompt}.
The results for the Linear Regression tasks are displayed in Fig.~\ref{fig:ablation_neuralucb:lr},
% (the results for the TSP tasks are deferred to Fig.~\ref{fig:ablation_neuralucb:tsp} in App.~\ref{app:subsec:more:ablation:ucb}), 
which demonstrate that general prompt optimization methods 
% NeuralUCB 
indeed significantly underperform in the problem of meta-prompt optimization for LLM-based agents.
We have also compared with INSTINCT \cite{lin2023instinct} (i.e., the best-performing baseline method from Fig.~\ref{fig:ablation_neuralucb:lr}) in the TSP tasks, and the results 
% The results for the TSP tasks are consistent with the results here 
(Fig.~\ref{fig:ablation_neuralucb:tsp} in App.~\ref{app:subsec:more:ablation:ucb}) also show that INSTINCT consistently performs worse than our \alg.
% are deferred to 
These results provide further justifications for our proposed adversarial bandit-based algorithms.

\textbf{Impact of the Degree of Exploration.}
Here we examine the impact of the degree of exploration, i.e., the value of $\eta$ (line 10 of Algo.~\ref{algo:EXPO}).
The results (Fig.~\ref{fig:TSP_ablation:eta}, first three figures) show that an excessively large degree of exploration (i.e., a small $\eta=10$) or an overly small degree of exploration (i.e., a large $\eta=1000$) both deteriorate the performance.
Moreover, the results also demonstrate that in easier tasks (i.e., TSP with 10 nodes), imposing a smaller degree of exploration (i.e., $\eta=1000$) leads to better performance compared to $\eta=10$, because it allows our \alg~to quickly converge to the optimal solution. 
% On the other hand, 
Meanwhile,
in more challenging tasks (i.e., TSP with 20 nodes), more exploration (i.e., $\eta=10$) results in better performance (than $\eta=1000$), because it makes it easier for our \alg~to escape local optimum.

\textbf{Experiments With Other LLMs.}
To evaluate the effectiveness of our approach when combined with different LLMs, here we adopt the challenging TSP task with 20 nodes and replace the GPT-3.5-Turbo model used in our original experiments (Sec.~\ref{subsec:exp:opro}) by the more advanced GPT-4-Turbo model.
The results in Fig.~\ref{fig:TSP_ablation:eta} (rightmost figure) show that 
% the use of the more advanced 
GPT-4-Turbo significantly improves the performance of both OPRO and our \alg.
More importantly, as visualized more clearly in Fig.~\ref{fig:TSP_ablation} in App.~\ref{app:ablation:subsec:gpt4-turbo}, when both adopting GPT-4-Turbo, our \alg~still significantly outperforms OPRO.
The results show that our \alg~can effectively improve the performance of LLM-based agents across different LLMs.
% Our main experiments presented above have all adopted GPT-3.5-turbo as the LLM for sequential decision-making. However, for the MAB experiments, we use GPT-4-turbo instead, while all other experiments are conducted with GPT-3.5-turbo. 
% To further validate the effectiveness of our \alg~algorithm across different LLMs, including more advanced models, we replaced the GPT-3.5-turbo agent with GPT-4-turbo in the TSP experiments. Specifically, we conducted additional evaluations under the high-difficulty setting (20 Nodes) to assess whether \alg~can maintain or even enhance performance when paired with a more capable LLM.
% Here we replace it by GPT-4-turbo and the results are shown in Fig.~\ref{fig:ablation_gpt4_turbo}.
% Detailed experimental results can be found in Appendix~\ref{fig:TSP_ablation}.
% The results demonstrate that \alg~remains effective with a more advanced model and consistently improves the agent's performance. In particular, \alg~with GPT-4-turbo achieves the best performance across all experimental comparisons under the TSP instance with 20 Nodes. 
We provide two further ablation studies in the appendix: fixing the meta-prompt to the optimal one discovered by \alg~yields dramatic performance gains (App.~\ref{appx:results_best_prompts}), and \alg~is substantially more computationally efficient than OPRO (App.~\ref{app:subsec:computational:efficiency}).

\section{Related Work}
\label{app:sec:related_work}

\textbf{Prompt Optimization.} Early prompt optimization focused on \emph{white-box} LLMs via gradient-based \cite{Shi2022TowardHR,Shin2020ElicitingKF}, RL \cite{deng2022rlprompt}, and soft-tuning methods \cite{lester2021power,Li2021PrefixTuningOC,zhong2021optiprompt}. Recent \emph{black-box} approaches leverage evolutionary algorithms \cite{fernando2023promptbreeder,guo2023connecting,zhou2023large}, zeroth-order optimization \cite{hu2024localized,zhan2024unlocking}, MCTS \cite{wang2023promptagent}, and Hyperband \cite{schneider2024hyperband}, with multi-armed bandits \cite{chen2023instructzero,lin2023instinct,shi2024best} and RL \cite{kong2024prewrite} further addressing sample efficiency. 
Prompt optimization has also been extended to vision language models \cite{codolzhang2025}.
Regarding exemplar selection, existing methods employ either static sets \cite{li2023support-examples,nguyen2023incontext,wang2023latent,zhang2022active} or dynamic retrievers \cite{albalak2024survey-data-selection,gao2024ambiguityaware,gupta2023coverage,levy2023diverse,liu2022makes,rubin2022retrieval,ye2023compositional}, with limited works exploring joint optimization \cite{opsahl2024optimizing,wan2024teach,wu2024prompt}. To our knowledge, our algorithm is the first to efficiently optimize \emph{meta-prompts} for agents in sequential decision-making tasks.

\textbf{LLM-Based Sequential Decision Making.} Recent studies leverage LLMs for sequential decision-making problems, including Bayesian optimization \cite{yang2023large}, multi-armed bandits \cite{chen2024efficient,krishnamurthy2024can,mukherjee2024pretraining,xia2024beyond}, and reinforcement learning \cite{dai2024context,monea2024llms,wang2024transformers}. However, most approaches rely on fixed, manual meta-prompts, limiting their full potential. Concurrently, the rapidly growing field of LLM-based agents has introduced a large number of benchmarks \cite{liu2023agentbench,wu2023smartplay,xi2024agentgym,castlejia2026}. 
A number of recent works have focused on improving the reasoning ability \cite{yang2026your,huang2026does,zhang2026heterogeneous,wang2026policy} and tool-used capability \cite{chen2026step,wei2026metaforge} of LLM-based agents via reinforcement learning.
For comprehensive overviews, we refer readers to recent surveys \cite{cheng2024exploring,wang2024survey,xi2023rise}.

\section{Conclusion}
In this work, we have proposed the \alg~algorithm to automatically optimize the meta-prompt for LLM-based sequential decision-making tasks, together with its extension \alges~which additionally optimizes the exemplars in the meta-prompt.
Both algorithms estimate meta-prompt scores with neural networks and select meta-prompts in a randomized fashion based on adversarial bandits.
Extensive experiments show that our algorithms considerably and consistently improve the performance of LLM-based sequential decision-making.
Extending our approach to LLM-based multi-agent systems is a promising direction for future work.

% ==========================================================================
% Limitations (REQUIRED by ACL/EMNLP; placed after Conclusion, before
% references; does NOT count toward the page limit).
% ==========================================================================
\section*{Limitations}
% TODO: replace this placeholder with a frank discussion of the
% limitations of your work --- scope of empirical validation, choice of
% LLMs, computational cost, hyperparameter sensitivity, theoretical
% assumptions that may not hold in practice, etc. ACL/EMNLP will desk
% reject papers without a Limitations section.
Our empirical study focuses on a specific set of LLM-based sequential decision-making
benchmarks (linear regression, TSP, multi-armed bandits, and GSM8K-style instruction
optimization) and a small number of frontier LLMs; performance and behavior may
differ on other tasks or models. The proposed algorithms also inherit the cost
profile of repeated LLM queries, which can be substantial for long horizons.
Finally, while our analysis draws on adversarial-bandit theory, the regret bounds
rely on assumptions (e.g., bounded reward feedback, fixed meta-prompt candidate set)
that may be violated in fully open-ended deployment settings.

\section*{Acknowledgements}
This work was supported in part by the National Natural Science Foundation of China (Grant No.~62506319), the Guangdong Basic and Applied Basic Research Foundation (Grant No.~2026A1515030032, 2025A1515110115), the Shenzhen Science and Technology Program (Grant No.~JCYJ20250604141031003) and the Pearl River Talent Program of Guangdong Province (Grant No.~2024QN11X069).

% ==========================================================================
% Optional but recommended: Ethical considerations
% ==========================================================================
% \section*{Ethical Considerations}
% (Optional. Use if your work has nontrivial ethical implications.)

% ==========================================================================
% Bibliography
% ==========================================================================
\bibliography{references}

% ==========================================================================
% Appendix (stays in two-column format per ACL formatting guidelines;
% papers with appendices not following the double column format may be
% desk rejected by ARR).
% ==========================================================================
\appendix
% \onecolumn
\newpage
% \section{Related Work}
% \label{app:sec:related_work}
% \input{tex_files/related_work}

\section{Our \alges~Algorithm to Additionally Optimize the Exemplar Sequences}
\label{app:expo:es}
% \begin{figure*}[t]
% \centering
% \includegraphics[width=0.95\linewidth]{figures/EXPO.pdf} \\
%  \caption{
%  Illustration of our \alg~algorithm. 
%  }
%  \label{fig:expo}
% \vspace{-2.5mm}
% \end{figure*}

\begin{algorithm*}[t]
\begin{algorithmic}[1]
    \INPUT: Initial task description $\mathcal{D}_0$, initial meta-instruction $\mathcal{I}_0$.\\
    Maximum number \(\mathcal{L}\) of exemplars in the meta-prompt, the number \(k^{\text{ES}}\) of exemplar sequences in the domain.
    % number of shuffled combinations 
 % , neural network model \(\mathcal{M}_{\text{ES}}\).
    \STATE Initialize the exemplar set $\mathcal{E}_0 = \emptyset$, and the subset $\mathcal{E}_0' = \emptyset$, meta-prompt-score set $\mathcal{S}_0 = \emptyset$, and cumulative score estimates $\hat{s}_j^{(0)}$ for all $j \in \{1, \ldots, k^{\text{ES}}\}$.
    \\
    Initialize the history of NN parameters \(\Theta_{\text{history}} = \emptyset\), 
    % initial neural network parameters \(\theta^{\text{ES}}_0\) for the model \(\mathcal{M}_{\text{ES}}\), 
    and the exemplar training set \(\mathcal{T}_0 = \emptyset\).
    % Initialize the exemplar set $\mathcal{E}_0 = \emptyset$, and the subset $\mathcal{E}_0' = \emptyset$.
    \FOR{iteration \(t = 0, 1, \ldots, T-1\)}
        \STATE \textbf{Lines 3-11 of Algo.~\ref{algo:EXPO}}.
        \STATE Compute the embedding \(g(\mathcal{E}_t')\) of the selected exemplar sequence \(\mathcal{E}_t'\), and add \(g(\mathcal{E}_t')\) and its score \(s_t\) to the exemplar training set:  $\mathcal{T}_{t+1} \gets \mathcal{T}_t \cup \{(g(\mathcal{E}_t'), s_t)\}$.
        % \STATE Update the NN parameters \(\theta^{\text{ES}}_t \to \theta^{\text{ES}}_{t+1}\) using the updated training set \(\mathcal{T}_{t+1}\) to minimize the MSE loss.
        \STATE Update the parameters $\theta^{\text{ES}}$ of the NN $\mathcal{M}_{\text{ES}}(g(\cdot); \theta^{\text{ES}})$ by using the updated $\mathcal{T}_{t+1}$ as the training set to minimize the MSE loss, yielding $\theta^{\text{ES}}_{t+1}$.
        \STATE Add the updated parameters to the history: $\Theta_{\text{history}} \gets \Theta_{\text{history}} \cup \{\theta^{\text{ES}}_{t+1}\}$.
        \IF{\(|\mathcal{E}_{t+1}| > \mathcal{L}\)}
            \STATE Randomly generate \(k^{\text{ES}}\) sequences of $\mathcal{L}$ exemplars from the exemplar set \(\mathcal{E}_{t+1}\): \(\{\mathcal{E}^{1}_{t+1}, \mathcal{E}^{2}_{t+1}, \ldots, \mathcal{E}^{k^{\text{ES}}}_{t+1}\}\), in which every \(\mathcal{E}^{j}_{t+1}\) represents an ordered set of \(\mathcal{L}\) exemplars from \(\mathcal{E}_{t+1}\).
            % where each \(\mathcal{E}_{t+1,j}\) contains exactly \(\mathcal{L}\) exemplars selected from a subset of \(\mathcal{E}_{t+1}\).

% {\color{magenta}
            \STATE Initialize cumulative score estimates \(\hat{s}_{j}^{(0)} = 0\) for all \(j \in \{1, \ldots, k^{\text{ES}}\}\). \qquad\qquad\qquad\qquad \tikzmark{right}\tikzmark{top}
            \FOR{each \(\mathcal{E}^{j}_{t+1}\) in \(\{\mathcal{E}^1_{t+1}, \ldots, \mathcal{E}^{k^{\text{ES}}}_{t+1}\}\)}
                \STATE Initialize cumulative score \(\hat{s}_{j}^{(0)} = 0\).
                \FOR{each historical model parameter \(\theta^{\text{ES}}_{i} \in \Theta_{\text{history}}\)}
                    \STATE Update the cumulative score for \(\mathcal{E}^j_{t+1}\): $\hat{s}_{j}^{(i)} = \hat{s}_{j}^{(i-1)} + \mathcal{M}_{\text{ES}}(g(\mathcal{E}^j_{t+1}); \theta^{\text{ES}}_{i})$.
                    % \[
                    % \]
                \ENDFOR
            \ENDFOR
            \STATE Compute the final cumulative score estimates: $\hat{s}_{j}^{(\text{final})} = \hat{s}_{j}^{(|\Theta_{\text{history}}|)}, \quad \forall j \in \{1, \ldots, k^{\text{ES}}\}$. \tikzmark{bottom}
            % }

            \STATE Compute the sampling distribution \(P_t^{\text{ES}}\) over the $k$ exemplar sequences:
            \[
            P_t^{\text{ES}}[j] = \frac{\exp(\eta \hat{s}_{j}^{(\text{final})})}{\sum_{l=1}^{k^{\text{ES}}} \exp(\eta \hat{s}_{l}^{(\text{final})})}, \quad \forall j \in \{1, \ldots, k^{\text{ES}}\}.
            \]
            \STATE Sample an exemplar sequence \(\mathcal{E}'_{t+1} \sim P_t^{\text{ES}}\).
        \ENDIF
    \ENDFOR
\end{algorithmic}
\AddNote{top}{bottom}{right}{\small cumulative score \\
estimates}
\caption{\alges}
\label{algo:ES-subset}
\end{algorithm*}

Our complete \alges~algorithm is described in Algo.~\ref{algo:ES-subset}.
As we have discussed in Sec.~\ref{subsec:expo:es}, there are two major differences compared to the way in which our \alg~algorithm optimizes the task description and meta-instruction (Algo.~\ref{algo:EXPO}).
Firstly, our domain of $k^{\text{ES}}$ arms (i.e., every arm corresponds to a randomly sampled exemplar sequence) changes in every iteration (line 8).
Secondly, as a result of the time-varying domains, we need to save a copy of the parameters of the NN trained in every iteration in order to compute the cumulative score estimates (lines 9-14).

\paragraph{Simplified Variant of Our \alges~Algorithm.}
When applying our \alges~algorithm to the LLM-based MAB algorithm in \cite{krishnamurthy2024can} (Sec.~\ref{subsec:exp:bandits}), we have adopted a simplified variant of our \alges.
This is because in the problem setting from \cite{krishnamurthy2024can}, the number of arms is small. Therefore, instead of including a subset of the history of exemplars in the prompt, their algorithm has instead included a summarized observation history.
An example of such summarized observation history with 5 arms (represented by 5 buttons with different colors) is given in Fig.~\ref{fig:example:summarized:history} below.
Therefore, here we aim to optimize the format of the summarized observation history.
Specifically, we \emph{optimize the order of the arms} in the summarized history, and our domain of arms consist of \emph{all cyclically shifted variants of the following sequence of buttons}: \{blue button, green button, red button, yellow button, purple button\}.
For example, some other arms (button sequences) in our domain include: \{green button, red button, yellow button, purple button, blue button\} and \{red button, yellow button, purple button, blue button, green button\}.
As a result, unlike our original \alges~algorithm described in Sec.~\ref{subsec:expo:es}, here we do not suffer from the issue of \emph{time-varying domain of arms}.
\begin{figure*}[t]
% \onecolumn
% \begin{minipage}[t]{0.4\textwidth}
\begin{mdframed}[linewidth=0.9pt]  % adjust linewidth as you desire
\footnotesize  % adjust text size as required
...\\\\
blue button: pressed 2 times with average reward 0.5 \\
green button: pressed 1 times with average reward 0.0 \\
red button: pressed 1 times with average reward 1.0 \\
yellow button: pressed 0 times \\
purple button: pressed 1 times with average reward 0.0 \\\\
...
\end{mdframed}
% \end{minipage}
% \hfill
% \begin{minipage}[t]{0.59\textwidth}
% \begin{mdframed}[linewidth=0.9pt]  % adjust linewidth as you desire
%     \footnotesize  % adjust text size as required
% \centerline{\textbf{\alg}}
%     {\color{purple}You are provided with a dataset containing a list of coordinates labeled as \texttt{<<POINTS>>} \par
% The dataset also includes a series of previously calculated routes, with associated lengths that are ordered from longest to shortest. 
% However, it’s key to note that shorter routes are more desirable. 
% Despite the presentation order, understand that the optimal route is identified by the smallest total length.} \\ \\
%     {\color{blue}Provide a unique trace that is distinct from any previous traces and shorter in length. 
% Ensure that this trace visits each point exactly once and adhere to the specified format 
% by starting with \texttt{<trace>} and concluding with \texttt{</trace>}.}
% \end{mdframed}
% \end{minipage}
\caption{
An example of the summarized observation history used by the LLM-based MAB algorithm from \cite{krishnamurthy2024can}.
}
\label{fig:example:summarized:history}
\end{figure*}

Therefore, when applying our \alges~algorithm to improve the LLM-based MAB method from \cite{krishnamurthy2024can} (Sec.~\ref{subsec:exp:bandits}), we make two modifications to our standard \alges~algorithm described in Algo.~\ref{algo:ES-subset}.
Firstly, instead of randomly sampling $k^{\text{ES}}$ exemplar sequences to form our domain of exemplar sequences, here our domain remains fixed across different iterations, i.e., all cyclically shifted variants of the arms.
Secondly, since here we do not suffer from the issue of time-varying domain of arms (i.e., exemplar sequences), we can resort to the incremental update of the cumulative reward estimates adopted by our \alg~algorithm (line 9 of Algo.~\ref{algo:EXPO}). As a result, we do not need to save a copy of the parameters of the NN trained in every iteration.

\section{More Details on Our Experimental Settings}
\label{app:sec:more:exp:details}
\subsection{More Details on the Generation of the Domain of Task Description and Meta-Instruction}
\label{app:subsec:detail:domain:generation}
Here we describe the details about how we generate the domain of task descriptions and meta-instructions.
Below we provide the prompt we have used to instruct the LLM to generate every prompt in the domain.

% \begin{table}[h]
% \renewcommand{\arraystretch}{1.5}
% \setlength{\tabcolsep}{10pt}
% \begin{tabular}{|p{0.95\textwidth}|}
% \hline
% \rowcolor[HTML]{E8E8E8}
% \textbf{Example Query: Meta-Prompt Instruction Rephrasing Template} \\
% \hline
% \textbf{To achieve a more effective TASK description and INSTRUCTION and convey its core essence more clearly, please enhance the content in the quote by rephrasing and changing some information:} "\{INITIAL\_META-PROMPT\}" \\ 
% \textbf{Please return directly the modified description without additional description.} \\
% \textbf{The modified description:} \\
% \hline
% \end{tabular}
% \end{table}

% \begin{figure}[h]
\begin{mycolorbox}{Query}{Example Query: Meta-Prompt Instruction Rephrasing Template}
\small
To achieve a more effective TASK description and INSTRUCTION and convey its core essence more clearly, please enhance the content in the quote by rephrasing and changing some information: "\{INITIAL\_META-PROMPT\}" \\ 
Please return directly the modified description without additional description. \\
The modified description: \\
\end{mycolorbox}
% \caption{The task description in the original OPRO prompt.}
% \label{fig:original:opro:prompt}
% \end{figure}

\textbf{Generation of the Domain.}
To effectively generate task-specific prompts, we utilized an initial prompt to guide the LLM in creating diverse task descriptions and meta-instructions. For each task, the LLM was prompted 100 times to rephrase the task description and meta-instruction separately, resulting in 100 unique rephrased prompts for each. Combined with the initial prompt, this process produced a total of \(101 \times 101\) combinations of task descriptions and meta-instructions for each task. 

To optimize computational efficiency, we pre-compute the embeddings of all task descriptions and meta-instructions in the domain using the embedding model $g(\cdot)$ and store the results to prevent redundant calculations during subsequent experiments. 

For the rephrasing process, we employed 
% the \texttt{gpt-4} model 
the GPT-4 model
with a temperature setting of \(1.3\), ensuring diverse and high-quality rephrased prompts for both task descriptions and meta-instructions.

% \subsection{More Details on the Tasks in Sec.~\ref{sec:experiments}}
\subsection{More Details on OPRO for the Linear Regression, Traveling Salesman Problem and Instruction Optimization (Sec.~\ref{subsec:exp:opro})}
\label{app:subsec:more:details:opro}

% \paragraph{Licenses.}
% For the baseline method of OPRO, we have adopted the official implementation from \url{https://github.com/google-deepmind/opro}, which is under the Apache-2.0 license.

\subsubsection{Task Setting.}
\label{app:subsubsec:task:setting:opro}
% \textbf{Linear Regression.} We conducted experiments on Linear Regression by selecting two challenging ground truth weight-bias (\(w, b\)) pairs. The experiments followed the OPRO framework, which requires warm-starting the LLM with initial exemplars. Using a fixed random seed, we first generated 50 random data points uniformly distributed within the range \([-1, 1]\), which perfectly satisfied the ground truth \(w_{\text{true}}, b_{\text{true}}\) pairs, ensuring that these data points could serve as the foundation for evaluating the LLM’s ability to model the relationships. Additionally, 5 \(w, b\) pairs with corresponding scores, sampled within the range \([10, 20]\), were generated using another fixed random seed to serve as the initial exemplars. At each iteration, the LLM was prompted 8 times (consisting of 1 inference with a temperature setting of $T=0$ and 7 inferences with a temperature setting of $T=1$) using the current exemplars, and the prompt was updated based on the generated outputs. The exemplars were dynamically updated to include the top 20 \(w, b\) pairs and their associated scores from all historical records across iterations, ensuring the LLM was always guided by the best-performing examples. The total number of iterations was set to 50, and each ground truth configuration was repeated 5 times for consistency.
\textbf{Linear Regression.} We conduct experiments on Linear Regression by selecting two challenging ground truth weight-bias (\(w, b\)) pairs. The experiments follow the OPRO framework, which requires warm-starting the LLM with initial exemplars. Using a fixed random seed, we first generate 50 random data points uniformly distributed within the range \([-1, 1]\), which perfectly satisfy the ground truth \(w_{\text{true}}, b_{\text{true}}\) pairs, ensuring that these data points can serve as the foundation for evaluating the LLM’s ability to model the relationships. Additionally, 5 \(w, b\) pairs with corresponding scores, sampled within the range \([10, 20]\), are generated using another fixed random seed to serve as the initial exemplars. At each iteration, the LLM is prompted 8 times (consisting of 1 inference with a temperature setting of $T=0$ and 7 inferences with a temperature setting of $T=1$) using the current exemplars, and the prompt is updated based on the generated outputs. The exemplars are dynamically updated to include the top 20 \(w, b\) pairs and their associated scores from all historical records across iterations, ensuring the LLM is always guided by the best-performing examples. The total number of iterations is set to 50, and each ground truth configuration is repeated 5 times for consistency.

\textbf{Traveling Salesman Problem (TSP).} For the TSP task, experiments are conducted on three problem sizes defined by the number of nodes: 10, 15, and 20. 
For each TSP instance, the problem is defined by randomly generating $n=10,15,20$ nodes, where the $x$ and $y$ coordinates of each node are sampled uniformly from the range $[-100, 100]$.
For each configuration, a specific TSP instance is generated using a fixed random seed, and a single random seed is used to generate warm-start exemplars to initialize the LLM prompts. 
To initialize the optimization process, we randomly sample 5 different TSP routes along with their corresponding total distances. These routes and their lengths are used as the initial exemplars for the LLM.
Each iteration consists of 8 prompt calls to the LLM, followed by an update of the exemplars based on the generated results. 
More specifically, during each iteration, the GPT-3.5-turbo is prompted 8 times using the same prompt, consisting of 1 inference with a temperature setting of $T=0$ to ensure stability and 7 inferences with a temperature setting of $T=1$ to encourage exploration. 
Similar to the Linear Regression task, the exemplars for TSP are updated to include the top 20 historical solutions with the best scores, ensuring the prompt leverages the most effective examples. The number of iterations is set to 100, 200, and 300 for 10-node, 15-node, and 20-node TSP problems, respectively, to account for the increasing complexity of the tasks. Each node configuration is repeated 3 times to ensure consistency and reliability.

% \textcolor{red}{
\textbf{Instruction Optimization.}
Following the experimental settings in OPRO, we have conducted instruction optimization experiments using the GSM8K benchmark dataset \cite{cobbe2021gsm8k}.
In line with OPRO, which emphasizes sample-efficient optimization, we randomly select 75 problem–answer pairs from the GSM8K training set as the instruction tuning subset. GPT-4-Turbo is used as the instruction optimizer to iteratively generate improved instructions, while Qwen-2.5-7B serves as the scorer LLM to evaluate the effectiveness of the instructions. During each iteration, GPT-4-Turbo is prompted 8 times with the current set of exemplars: 1 inference is performed with temperature $T = 0$ and the other 7 inferences are conducted with $T = 1$, yielding 8 candidate instructions per round. Each candidate is evaluated using the \texttt{Q\_end} strategy by prompting Qwen-2.5-7B on the 75 training questions. The exemplar set for the next iteration consists of the 20 best instructions based on evaluation scores, along with 3 randomly sampled training exemplars. 
Here our \alg~algorithm only considers optimizing 
% meta-prompt optimization focuses solely on refining 
the meta-instruction component in the meta-prompt, while keeping the task description fixed. This is because in the setting of the instruction optimization experiments in OPRO, 
% , under the OPRO framework, 
the task description 
% in the IO setting 
is instantiated as a set of three randomly selected exemplars from the training set, which makes it not directly amenable to the application of our meta-prompt optimization algorithm.
% does not conform to the structured prompt format required for our meta-prompt optimization algorithm. 
As such, only the meta-instruction is optimized in this experiment. 
The optimization is initialized with the Chain-of-Thought (CoT) prompt prefix "Let's think step by step", which serves as the initial instruction 
% starting point 
for OPRO, \alg~ and \alges~. Each experiment is repeated 3 times.
% }

\subsubsection{Evaluation Metrics.}
\textbf{Linear Regression.}
In the Linear Regression task, the performance of the algorithms is evaluated using the Mean Squared Error (MSE) metric. Given a set of $N$ one-dimensional input data points \(\mathbf{x} \in \mathbb{R}\) and their corresponding ground truth labels \(\mathbf{y} \in \mathbb{R}\), the MSE is computed as:
\[
\text{MSE} = \frac{1}{N} \left\lVert \mathbf{y} - (w \cdot \mathbf{x} + b) \right\rVert^2,
\]
where \(w \in \mathbb{R}\) and \(b \in \mathbb{R}\) are the weight and bias parameters inferred by the LLM, and \(N\) is the total number of data points.

\textbf{Traveling Salesman Problem (TSP).}
For the TSP task, the performance of the LLM-generated solutions is evaluated based on the total Euclidean distance of the TSP tour. Given a set of two-dimensional points \(\{(x_i, y_i)\}_{i=1}^N\), where \(N\) is the total number of nodes, the length of a proposed TSP tour \(\mathcal{P} = [\pi(1), \pi(2), \dots, \pi(N), \pi(1)]\) is computed as:
% \[
% \text{Length} = \sum_{i=1}^{N} \sqrt{\left( x_{\pi(i+1)} - x_{\pi(i)} \right)^2 + \left( y_{\pi(i+1)} - y_{\pi(i)} \right)^2},
% \]
\[
\sum_{i=1}^{N} \sqrt{\left( x_{\pi(i+1)} - x_{\pi(i)} \right)^2 + \left( y_{\pi(i+1)} - y_{\pi(i)} \right)^2},
\]
where \(\pi\) represents the permutation of nodes in the proposed tour, and \(\pi(N+1) = \pi(1)\) ensures the tour returns to the starting node.

To evaluate the convergence and effectiveness of the agents, we use the \textit{Optimality Gap} metric, which quantifies the deviation of the solver's best-found solution from the true optimal solution. It is defined as:
\[
\text{Optimality Gap} = \frac{\text{SolverOptimal} - \text{Optima}}{\text{Optima}} \times 100\%,
\]
where:
\begin{itemize}
    \item \(\text{SolverOptimal}\) denotes the shortest tour length found by the solver up to the current iteration.
    \item \(\text{Optima}\) is the length of the known optimal TSP tour.
\end{itemize}

% \textcolor{red}{
\textbf{Instruction Optimization.}
To assess the effectiveness of each generated instruction, we adopt the \texttt{Q\_end} evaluation strategy introduced in OPRO. For a given instruction, prompts are constructed by appending the instruction to each question in the set, and the Qwen-2.5-7B-Instruct model with the temperature set to 0 is queried to generate corresponding answers. The final predicted outputs are then compared to ground-truth answers using exact numerical match accuracy. Specifically, for the GSM8K dataset, an answer is considered correct if and only if it exactly matches the reference solution in its numerical form—partial matches, approximations, or differently formatted correct values are all treated as incorrect.
% }

% \textcolor{red}{
Formally, the score is defined as the average accuracy over the set:
\begin{equation}
\text{Accuracy} = \frac{1}{N} \sum_{i=1}^{N} \mathbbm{1}[\hat{y}_i = y_i],
\end{equation}
where $\hat{y}_i$ denotes the model-predicted answer for the $i$-th example, $y_i$ is the corresponding ground-truth answer, and $\mathbbm{1}[\cdot]$ is the indicator function that equals 1 when the predicted and true answers exactly match. Higher average accuracy reflects that the algorithm has identified more effective instructions, indicating greater optimization efficiency and overall method quality.
% }

\subsubsection{Design of Prompt Score.}
\textbf{Linear Regression and TSP.}
In both the Linear Regression and TSP tasks, optimal solutions are characterized by lower evaluation scores. To align with the requirements of the algorithm and ensure more stable learning, we define the \textit{Meta-Prompt Score} using the formula:
\[
\text{Meta-Prompt Score} = \frac{-\text{Evaluation Score} + b}{b},
\]
where \(b > 0\) is a stabilizing constant. This formulation ensures that lower evaluation scores correspond to higher prompt scores, which better facilitates the optimization process and contributes to steady algorithmic learning.

For the Linear Regression task, the \textit{Evaluation Score} is defined as the Mean Squared Error (MSE) of the weight-bias (\(w, b\)) pairs proposed by the algorithm at each iteration under a Temperature=0 stable inference. The MSE is computed based on the provided one-dimensional data points.

For the TSP task, the \textit{Evaluation Score} corresponds to the total Euclidean distance (\textit{Length}) of the TSP tour proposed by the algorithm at each iteration, also under a Temperature=0 stable inference.

% \textcolor{red}{
\textbf{Instruction Optimization.}  
Instruction Optimization task uses accuracy as the evaluation metric, where higher scores signify superior instruction quality and meta-prompt quality. Accordingly, we directly define the \textit{Meta-prompt Score} as the average accuracy (scaled to $[0,1]$) achieved by the instruction generated under the current meta-prompt.
% }

% \textcolor{red}{
Specifically, at each iteration, we perform a Temperature=0 inference with GPT-4-Turbo to generate a candidate instruction from the current meta-prompt. This instruction is then evaluated on the training subset using the Qwen-2.5-7B-Instruct model under the \texttt{Q\_end} setting, and the resulting average accuracy serves as the Prompt Score:
\begin{equation}
\text{Meta-Prompt Score} = \text{Accuracy},
\end{equation}
This formulation directly reflects the effectiveness of the meta-prompt: higher Accuracy values lead to higher Meta-Prompt Scores, indicating more successful instruction optimization.
% }

% \subsubsection{Related Models and \alg~  Parameters.}
\subsubsection{Details about the Models and Parameters in Our Algorithms}

\textbf{LLM Agents and Embedding Model.}
In our experiments, the primary LLM agent used is GPT-3.5-Turbo.
% \texttt{gpt-3.5-turbo}. 
For embedding generation, we utilized OpenAI's \texttt{text-embedding-3-large} model, which outputs embeddings of dimensionality 3072. These embeddings were used to represent both the task description and meta-instruction in the \alg~ framework. The embeddings were also employed to represent the exemplars in the \alges~ framework. During each iteration of inference, the LLM agent performed 1 prediction with a temperature setting of \(T=0\) to provide a stable solution and 7 additional predictions with a temperature setting of \(T=1\) to encourage exploration.

\textbf{Neural Network Parameters.}
For the Linear Regression and TSP experiments, the input to the neural network consists of the concatenated embeddings of the task description and meta-instruction, resulting in an input dimensionality of \(3072 + 3072 = 6144\). The neural network employs a single hidden layer with a width of 1536 and produces a single scalar output. The training objective is to minimize the Mean Squared Error (MSE) loss function.

For the Instruction Optimization experiments, we only optimize the meta-instruction without optimizing the task description. Consequently, the neural network input dimensionality for IO is 3072, corresponding solely to the meta-instruction embedding. The neural network in the IO experiment employs a single hidden layer with a width of 512 and outputs a single scalar value. The training objective is also to minimize the MSE loss.

For the \alges, the exemplar selection process differs depending on the iteration count. During the initial iterations, when fewer than 20 optimal historical records are available, we use all available exemplars. As the iteration count increases, exemplars are selected from the top \(\min(\text{total exemplar records}, 30)\) historical optimal records. From this pool, 257 exemplars are constructed, consisting of 256 randomly selected exemplars and 1 heuristic exemplar
% , which was 
generated from a combination of 20 best historical records. The neural network for \alges~ operates on an input dimensionality of 3072, corresponding to the embedding of a single exemplar. It employs a single hidden layer with a width of 512 and produces a single scalar output. The training objective is to minimize the Mean Squared Error (MSE) loss.

\textbf{EXP3 Learning Rate.}
In the \alg~, the learning rate parameter \(\eta_{\text{desc}}\) is set to \(100\) for selecting task description and meta-instruction combinations in both the Linear Regression and TSP experiments. For the Instruction Optimization experiments, where only the meta-instruction is optimized, the learning rate parameter \(\eta_{\text{meta-inst}}\) is set to \(50\). 

In the \alges, the learning rate parameter for selecting combinations of task description and meta-instruction is set identically to that in \alg. And the learning rate parameter for exemplar selection, \(\eta_{\text{exemplar}}\), is set to \(25\).

\subsubsection{Enhanced OPRO}
\label{app:sec:enhanced:opro}
Here, we describe how we have enhanced the original algorithm \cite{yang2023large} by modifying its prompts.

During initial experiments with the meta-prompts provided by the original OPRO algorithm \cite{yang2023large} for task description rephrasing, we observed that the LLM often misinterprets the \emph{descending order} semantics described in the original design. 
In tasks like TSP and Linear Regression, where better solutions correspond to lower evaluation scores, \emph{descending order} is intended to arrange solutions from high evaluation scores to low. However, the LLM frequently misunderstands this as a \emph{descending order of solution quality}, interpreting higher-ranked solutions as better and lower-ranked ones as worse, which is contrary to the intended meaning.

To address this issue, we enhance the orginal meta-prompts by explicitly clarifying the semantics of descending order in the context of evaluation scores. This modification ensures that the LLM accurately understand the intended instructions. When tested with the enhanced prompts, the problem was resolved, and the LLM is able to consistently generate correct rephrased task descriptions.
For a clearer illustration, we provide below the original OPRO meta-prompt (Fig.~\ref{fig:original:opro:prompt}) and our enhanced OPRO meta-prompt (Fig.~\ref{fig:enhanced:opro:prompt}).

% \begin{figure}[h]
% % \onecolumn
% % \begin{minipage}[t]{0.4\textwidth}
% \begin{mdframed}[linewidth=0.9pt]  % adjust linewidth as you desire
% \footnotesize  % adjust text size as required
% You are given a list of points with coordinates below: \{POINTS\}.\\
% Below are some previous traces and their lengths. 
% The traces are arranged in descending order based on their lengths, 
% where lower values are better. \\
% ......
% \end{mdframed}
% \caption{
% The task description in the original OPRO prompt.
% }
% \label{fig:original:opro:prompt}
% \end{figure}

\begin{figure}[H]
\begin{mycolorbox}{Query}{The task description in the original OPRO prompt}
\small
You are given a list of points with coordinates below: \{POINTS\}.\\
Below are some previous traces and their lengths. 
The traces are arranged in descending order based on their lengths, 
where lower values are better. \\
......
\end{mycolorbox}
\caption{The task description in the original OPRO prompt.}
\label{fig:original:opro:prompt}
\end{figure}

% \begin{figure}[h]
% % \onecolumn
% % \begin{minipage}[t]{0.4\textwidth}
% \begin{mdframed}[linewidth=0.9pt]  % adjust linewidth as you desire
% \footnotesize  % adjust text size as required
% You are given a list of points with coordinates below: \{POINTS\}.\\
% Below are some previous traces and their lengths. The traces are arranged in descending order based on their lengths, where {\color{red}smaller lengths indicate better solutions}. {\color{red}Therefore, the traces are listed from the largest length to the smallest, the trace with the smallest length is considered the most optimal}. \\
% ......
% \end{mdframed}
% \caption{
% The task description in the enhanced OPRO prompt. The texts we have mofified are highlighted in red.
% }
% \label{fig:enhanced:opro:prompt}
% \end{figure}

\begin{figure}[H]
\begin{mycolorbox}{Query}{The task description in our enhanced OPRO prompt}
\small
You are given a list of points with coordinates below: \{POINTS\}.\\
Below are some previous traces and their lengths. The traces are arranged in descending order based on their lengths, where {\color{red}smaller lengths indicate better solutions}. {\color{red}Therefore, the traces are listed from the largest length to the smallest, the trace with the smallest length is considered the most optimal}. \\
......
    % Rephrase the following description: [Initial instruction] \\
    % The rephrased description is: 
\end{mycolorbox}
\caption{The task description in the enhanced OPRO prompt. The texts we have modified are highlighted in {\color{red}red}.}
\label{fig:enhanced:opro:prompt}
\end{figure}

\subsection{More Details on the LLM-Based Multi-Armed Bandits Task (Sec.~\ref{subsec:exp:bandits})}

\subsubsection{Explanation of BSSCD and BSSND}

\label{subsec:exp:app:more:details:bandit}

We provide a detailed explanation and demonstration of prompt designs for both BSSCD and BSSND \cite{krishnamurthy2024can}, highlighting their key components and structures. Figure~\ref{fig:wholeprompt4BSSCD} illustrates a complete example of a BSSCD prompt designed for the MAB problem under the hard difficulty setting. It showcases the structure and color-coded components of the prompt in detail.

\begin{figure*}[t]
\begin{mycolorbox}{Query}{The setting of the prompt in MAB}
{[SYSTEM]
\vspace{-2mm}
\begin{tcolorbox}[colback=violet!15, colframe=violet!15, width=15.9cm, enlarge left by=-5mm, top=0.5mm, bottom=0.5mm, boxsep=0mm, arc=0mm, outer arc=0mm, left=4.5mm, right=4.5mm]
{\color{orange}You are a bandit algorithm in a room with 5 buttons labeled blue, green, red, yellow, purple.} 
{\color{teal}Each button is associated with a Bernoulli distribution with a fixed but unknown mean; the means for the buttons could be different. For each button, when you press it, you will get a reward that is sampled from the button's associated distribution. You have 100 time steps and, on each time step, you can choose any button and receive the reward. Your goal is to maximize the total reward over the 100 time steps.}
\end{tcolorbox}
\vspace{-5mm}
\begin{tcolorbox}[colback=blue!15, colframe=blue!15, width=15.9cm, enlarge left by=-5mm, top=0.5mm, bottom=0.5mm, boxsep=0mm, arc=0mm, outer arc=0mm, left=4.5mm, right=4.5mm]
{\color{brown}At each time step, I will show you a summary of your past choices and rewards.} Then you must make the next choice. {\color{red}You may output a distribution over the 5 buttons formatted EXACTLY like "blue:a,green:b,red:c,yellow:d,purple:e".} {\color{yellow}{Let’s think step by step to make sure we make a good choice.}}
\end{tcolorbox}
\vspace{-2mm}
You must provide your final answer within the tags\ {\color{red}\textless Answer\textgreater DIST\textless\textbackslash Answer\textgreater\ where DIST is the distribution in the format specified above.}} \\
\noindent\rule[0.5ex]{\textwidth}{0.5pt} \\
{[USER] \\
So far you have played 5 times {\color{brown}with your past choices and rewards summarized as follows:\\
blue button: pressed 2 times with average reward 0.5 \\
green button: pressed 1 times with average reward 0.0 \\
red button: pressed 1 times with average reward 1.0 \\
yellow button: pressed 0 times \\
purple button: pressed 1 times with average reward 0.0 \\}Which button will you choose next? Remember, YOU MUST provide your final answer within the tags\ {\color{red}\textless Answer\textgreater DIST\textless\textbackslash Answer\textgreater\ where DIST is formatted like "blue:a,green:b,red:c,yellow:d,purple:e".}
}
\end{mycolorbox}
\caption{A complete example of the prompt in MAB. The different components in the prompt are explained in detail in App.~\ref{subsec:exp:app:more:details:bandit}.}
\label{fig:wholeprompt4BSSCD}
\end{figure*}

\begin{itemize}
    {\color{orange}\item B}utton scenario and {\color{orange}S}uggestive framing, providing the foundational task scenario, clarifying the role of the agent, and framing the objective of the task in a suggestive manner to guide decision-making.
    {\color{teal} \item Description of the multi-armed bandit problem}, offering the agent a detailed task description, including comprehensive information about the task's objectives, constraints, and operational details.
    {\color{brown} \item S}ummarized history, presenting a condensed version of historical decisions and reward feedback to the agent, instead of providing step-by-step decision and reward feedback.
    {\color{yellow} \item C}hain-of-thought or {\color{yellow}N}o CoT, indicating whether to encourage the agent to engage in step-by-step reasoning for decision making.
    {\color{red} \item D}istribution over actions, encouraging the agent to generate a probability distribution over the arms of the bandit, instead of making deterministic decisions.
\end{itemize}
When we use our \alg~algorithm to optimize the task description and meta-instruction, 
% Specifically,
the upper section with {\color{violet}{light purple}} background corresponds to the  {\color{violet}{Task Description}}, where as the section below it with {\color{blue}{light blue}} background represents the {\color{blue}{Meta-Instruction}}. In other words, our \alg~algorithm is used to optimize the text in these two sections.

\subsubsection{Task Setting}
The experiments are conducted for both the BSSND and BSSCD prompts under two pre-defined difficulty levels: \textit{hard} and \textit{easy}. For the \textit{hard} setting, the MAB instance consists of $K=5$ arms, where the best arm has a mean reward of $\mu^\star = 0.5 + \Delta / 2$ with $\Delta = 0.2$, and all other arms have a mean reward of $\mu = 0.5 - \Delta / 2$. For the \textit{easy} setting, the MAB instance consists of $K=4$ arms with a larger gap $\Delta = 0.5$ between the best arm and the suboptimal arm.
% We evaluated the performance of LLM-based agents on the Multi-Armed Bandit (MAB) problem using two distinct prompt settings: BSSCD and BSSND. The experiments were conducted under two difficulty levels: \textit{Hard} and \textit{Easy}. For the Hard setting, the task involved 5 arms, where one arm followed a Bernoulli distribution with a parameter of 0.6, while the remaining 4 arms had parameters of 0.4. For the Easy setting, the task used 4 arms, with one arm having a Bernoulli parameter of 0.75 and the remaining 3 arms having parameters of 0.25. 
We set the blue button as the optimal arm in experiments, corresponding to the arm with the highest expected reward. Each configuration is tested using two fixed random seeds, with experiments repeated 3 times for each seed, resulting in a total of \(2 \times 3 = 6\) runs per setting. Each experiment consists of 100 iterations, with the LLM-based agents making decisions and updating prompts iteratively to optimize performance.
% The experiments are conducted for both the BSSND and BSSCD prompts under two predefined difficulty settings: \textit{hard} and \textit{easy}. For the \textit{hard} setting, the MAB instance consists of $K=5$ arms, where the best arm has a mean reward of $\mu^\star = 0.5 + \Delta / 2$ with $\Delta = 0.2$, and all other arms have a mean reward of $\mu = 0.5 - \Delta / 2$. For the \textit{easy} setting, the MAB instance consists of $K=4$ arms with a larger gap $\Delta = 0.5$ between the best arm and the suboptimal arm. 
The work of \cite{krishnamurthy2024can} has reported that GPT-3.5 models encounter exploration failures in MAB tasks, making them unsuitable as agents for solving such problems. In contrast, GPT-4 demonstrates the capability to effectively handle the exploration-exploitation trade-off inherent in MAB settings. Therefore, we adopt GPT-4-turbo as the LLM agent for this experiment. 
% In each iteration, GPT-4-turbo is prompted with a temperature setting of $T=0$ to generate one inference result. Each experimental configuration runs for 100 iterations. For each MAB configuration, the experiment is repeated 6 times.

\subsubsection{Evaluation Metric.}
In the LLM-based Multi-Armed Bandit (MAB) task (Sec.~\ref{subsec:exp:bandits}), the performance of the LLM agent is assessed using the \textit{Cumulative Regret} metric. At each iteration, the LLM agent outputs a probability distribution over the arms, representing the likelihood of sampling each arm.

Formally, let there be \(K\) arms, each associated with an expected reward \(\mu_1, \mu_2, \dots, \mu_K\), where \(\mu^* = \max_{k \in \{1, \dots, K\}} \mu_k\) denotes the expected reward of the optimal arm. At iteration \(t\), we sample an arm \(a_t \in \{1, \dots, K\}\), which is determined by the probability distribution provided by the LLM agent. The instantaneous regret for iteration \(t\) is then defined as:
\[
r_t = \mu^* - \mu_{a_t},
\]
where \(\mu_{a_t}\) represents the expected reward of the selected arm \(a_t\) at iteration \(t\).

The cumulative regret after \(T\) iterations is computed as:
\[
R_T = \sum_{t=1}^{T} r_t = \sum_{t=1}^{T} \left( \mu^* - \mu_{a_t} \right).
\]

\subsubsection{Design of Prompt Score.}
The score of the prompt is designed to quantify the expected reward of the LLM agent's sampling strategy at each iteration. At iteration \( t \), the LLM agent outputs a sampling probability distribution \( \{ p_1, p_2, \dots, p_K \} \), where \( p_i \) represents the probability of selecting arm \( i \) (\( i = 1, 2, \dots, K \), with \( K \) being the total number of arms). Simultaneously, the historical records from the first \( (t-1) \) iterations allow us to compute an unbiased estimate of the Bernoulli reward parameter for each arm, \( \hat{\mu}_i \), based on the observed rewards and sampling counts.

For arm \( i \), the Bernoulli parameter \( \hat{\mu}_i \) is estimated as:
\[
\hat{\mu}_i = 
\begin{cases} 
0, & \text{if } n_i = 0, \\
\frac{\sum_{j=1}^{t-1} R_{i,j}}{n_i}, & \text{if } n_i > 0.
\end{cases}
\]
where \( \sum_{j=1}^{t-1} R_{i,j} \) denotes the cumulative reward obtained from arm \( i \) during the first \( (t-1) \) iterations, and \( n_i \) represents the total number of times arm \( i \) was sampled during the same period.

The LLM agent's expected reward \( \hat{R}_{\text{expected}} \) at iteration \( t \) is then calculated by weighting the estimated Bernoulli parameters \( \{\hat{\mu}_1, \hat{\mu}_2, \dots, \hat{\mu}_K\} \) with the sampling probabilities \( \{ p_1, p_2, \dots, p_K \} \) provided by the LLM:
\[
\hat{R}_{\text{expected}} = \sum_{i=1}^{K} p_i \cdot \hat{\mu}_i.
\]

This expected reward \( \hat{R}_{\text{expected}} \) serves as the score of the prompt.

\paragraph{Motivation for the Score Design.}
The design of the prompt score is driven by the objective of guiding the LLM agent to favor arms with higher expected rewards, represented by \( \mu_i \). Since the true values of \( \mu_i \) are not available, the prompt score is designed to estimate this quantity based on observed data. Specifically, the higher the value of \( \mu_i \), the higher the sampling probability \( p_i \) should be assigned to arm \( i \), reflecting the optimal choice. Conversely, arms with lower values of \( \mu_i \) should be assigned lower probabilities.

The original score with the Bernoulli parameters:
\[
R_{\text{expected}} = \sum_{i=1}^{K} p_i \mu_i.
\]
In the absence of the true \( \mu_i \), we rely on the unbiased estimates \( \hat{\mu}_i \):
\[
\hat{R}_{\text{expected}} = \sum_{i=1}^{K} p_i \hat{\mu}_i.
\]

This design is justified because, for most of iterations, the score \( \sum_{i=1}^K p_i \hat{\mu}_i \) is an unbiased estimate of the true expected reward \( \sum_{i=1}^K p_i \mu_i \), and we proceed to formally establish this unbiasedness.

\paragraph{Proof of Unbiasedness.}
For iteration \( t \), where \( n_i > 0 \) for all \( i \), we aim to show that the score \( \sum_{i=1}^K p_i \hat{\mu}_i \) is an unbiased estimate of the true expected reward \( \sum_{i=1}^K p_i \mu_i \). Since \( \hat{\mu}_i \) is an unbiased estimate of \( \mu_i \), we have:
\[
\expect{\hat{\mu}_i} = \expect{\mu_i},
\]
Thus, by the linearity of expectation, we obtain:
\begin{align*}
    \expect{\sum_{i=1}^K p_i \hat{\mu}_i} &= \sum_{i=1}^K p_i \expect{\hat{\mu}_i} \\
    &= \sum_{i=1}^K p_i \expect{\mu_i} \\
    &= \expect{\sum_{i=1}^K p_i \mu_i}
\end{align*}
This shows that the score $\hat{R}_{\text{expected}}$ is an unbiased estimate of the true expected reward $R_{\text{expected}}$.

\subsubsection{Details about the Models and Parameters in Our Algorithms}
\textbf{LLM Agents and Embedding Model.}
For the MAB tasks, the primary LLM agent is 
% \texttt{gpt-4-turbo} 
GPT-4-Turbo
and the fixed inference temperature is set to \(T=0\). For embedding generation, we employed OpenAI's \texttt{text-embedding-3-large} model, which outputs embeddings with a dimensionality of 3072. These embeddings are utilized to represent the prompts provided to the LLM agent during the experiments. At each iteration, the LLM is prompted once using the designed prompt.

\textbf{Neural Network Parameters.}
For the \alg, the input to the neural network consists of the concatenated embeddings of the task description and meta-instruction, resulting in an input dimensionality of \(3072 + 3072 = 6144\). The neural network employs a single hidden layer with a width of 1536 and produces a scalar output. The model is trained by minimizing the Mean Squared Error (MSE) loss function.

For the \alges, the neural network is designed to process \(K\) exemplars, where \(K\) is determined by the total number of available summaries. To ensure fairness, \(K\) distinct exemplar combinations are generated at each iteration using a cyclic rotation mechanism. This mechanism ensures that each summary occupies every possible position within the exemplar sequence. Formally, given \(K\) summaries indexed as \(\{e_0, e_1, \dots, e_{K-1}\}\), the \(i\)-th exemplar combination is defined as:
\[
(e_i, e_{(i+1) \mod K}, \dots, e_{(i+K-1) \mod K}).
\]
This guarantees that each summary appears in every position across all \(K\) combinations.

Each exemplar is embedded into a 3072-dimensional vector using the embedding model, and these embeddings are processed individually by the neural network. The neural network consists of a single hidden layer with a width of 512 and produces a scalar output. Like \alg, the training objective is to minimize the Mean Squared Error (MSE) loss function.

\textbf{EXP3 Learning Rate.} 
For the \alg, the learning rate parameter \(\eta_{\text{desc}}\) is set to \(10\) for selecting task description and meta-instruction combinations. In the \alges, two learning rate parameters are used: \(\eta_{\text{desc}}\) is set to \(10\) for selecting task description and meta-instruction combinations, and \(\eta_{\text{exemplar}}\) is set to \(10\) for selecting exemplars.

\subsection{Improving Numerical Stability}

To prevent numerical overflow during the computation of exponentials in our algorithms,
% the EXP3 algorithm, 
a translation constant \( C^{(t)} \) is introduced at each iteration \( t \). This constant stabilizes the computation by shifting the cumulative scores, ensuring the algorithm operates reliably until convergence without altering the resulting probability distribution. The translation constant is defined as:
\begin{equation}
C^{(t)} = \max_{j} S_j^{(t)}.
\end{equation}

The translated scores are:
\begin{equation}
\tilde{S}_i^{(t)} = S_i^{(t)} - C^{(t)}.
\end{equation}

The probability distribution after translation is:
\begin{equation}
\tilde{P}_t[i] = \frac{\exp\big(\eta \tilde{S}_i^{(t)}\big)}{\sum_{j=1}^k \exp\big(\eta \tilde{S}_j^{(t)}\big)}.
\end{equation}

Substituting \( \tilde{S}_i^{(t)} = S_i^{(t)} - C^{(t)} \):
\begin{equation}
\tilde{P}_t[i] = \frac{\exp\big(\eta \big(S_i^{(t)} - C^{(t)}\big)\big)}{\sum_{j=1}^k \exp\big(\eta \big(S_j^{(t)} - C^{(t)}\big)\big)}.
\end{equation}

Using \( \exp(a - b) = \frac{\exp(a)}{\exp(b)} \):
\begin{equation}
\tilde{P}_t[i] = \frac{\exp\big(\eta S_i^{(t)}\big) / \exp\big(\eta C^{(t)}\big)}{
\sum_{j=1}^k \big(\exp\big(\eta S_j^{(t)}\big) / \exp\big(\eta C^{(t)}\big)\big)}.
\end{equation}

Simplifying:
\begin{equation}
\tilde{P}_t[i] = \frac{\exp\big(\eta S_i^{(t)}\big)}{\sum_{j=1}^k \exp\big(\eta S_j^{(t)}\big)}.
\end{equation}

Thus, the probabilities remain unchanged:
\begin{equation}
\tilde{P}_t[i] = P_t[i].
\end{equation}

% {\color{red}
\subsection{Compute Resources and Licenses}
\label{app:subsec:compute:license}
All experiments were conducted on a server equipped with an NVIDIA L4 GPU and 53 GB of RAM.
\label{app:subsec:compute:resources}
The GSM8K dataset \cite{cobbe2021gsm8k} is released under the MIT license.
We access GPT-3.5-Turbo, GPT-4-Turbo, and the \texttt{text-embedding-3-large} embedding model through the OpenAI API under the OpenAI Terms of Use, and we use the openly released Qwen-2.5-7B-Instruct model under the Apache~2.0 license.
For the baseline methods, the official OPRO implementation \cite{yang2023large} is released under the Apache~2.0 license, the DSPy library that provides the MIPRO baseline \cite{opsahl2024optimizing} is released under the MIT license, and we additionally use the publicly released implementations of INSTINCT \cite{lin2023instinct} and PromptBreeder \cite{fernando2023promptbreeder} for comparison.

% }

\section{More Experimental Results}

\subsection{Results of Optimal Prompts on Tasks}
\label{appx:results_best_prompts}
\begin{figure*}[h]
\vspace{-3mm}
\centering
     \begin{tabular}{cccc}
     \hspace{-7mm}
         \includegraphics[width=0.25\linewidth]{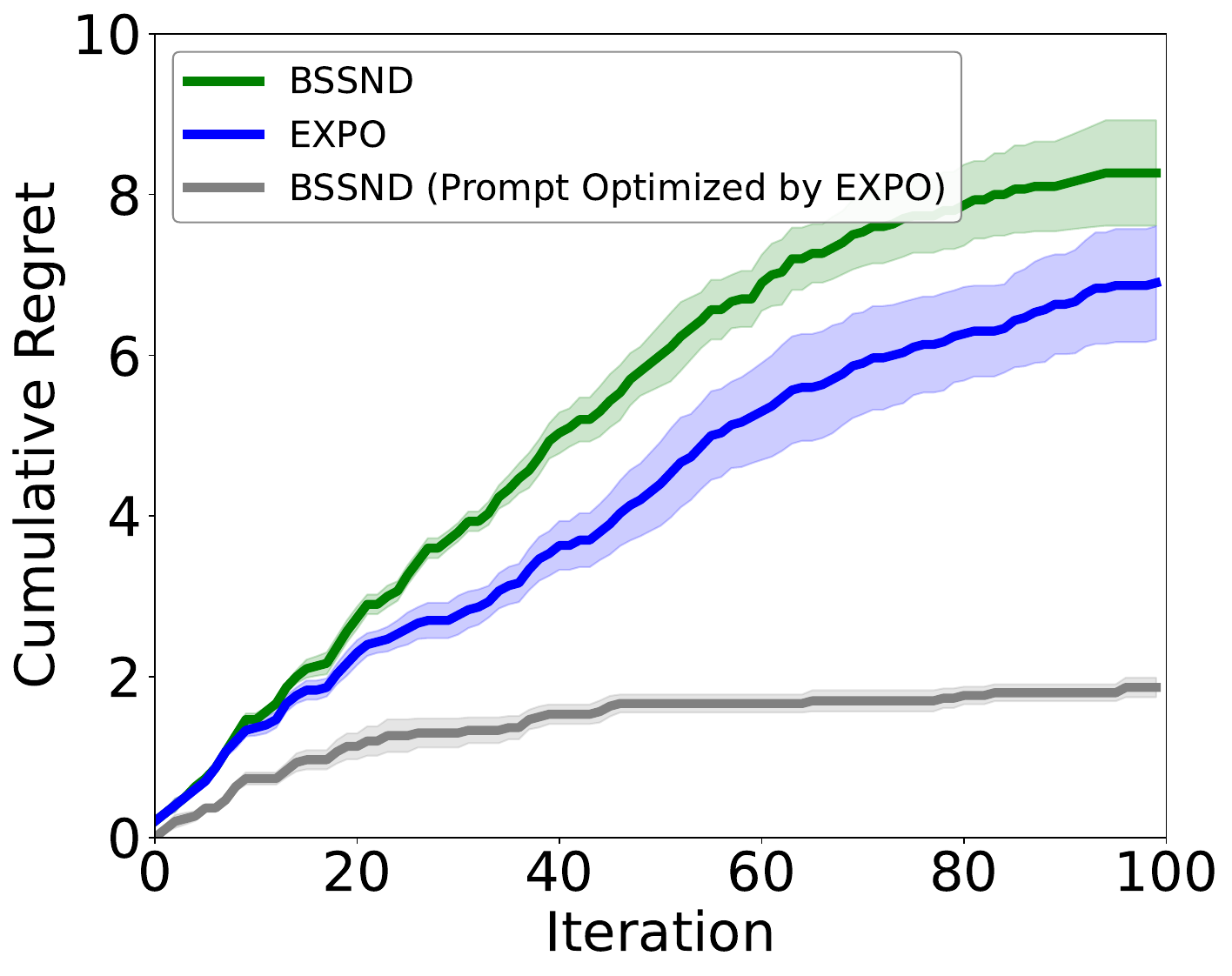} & \hspace{-5mm} 
         \includegraphics[width=0.25\linewidth]{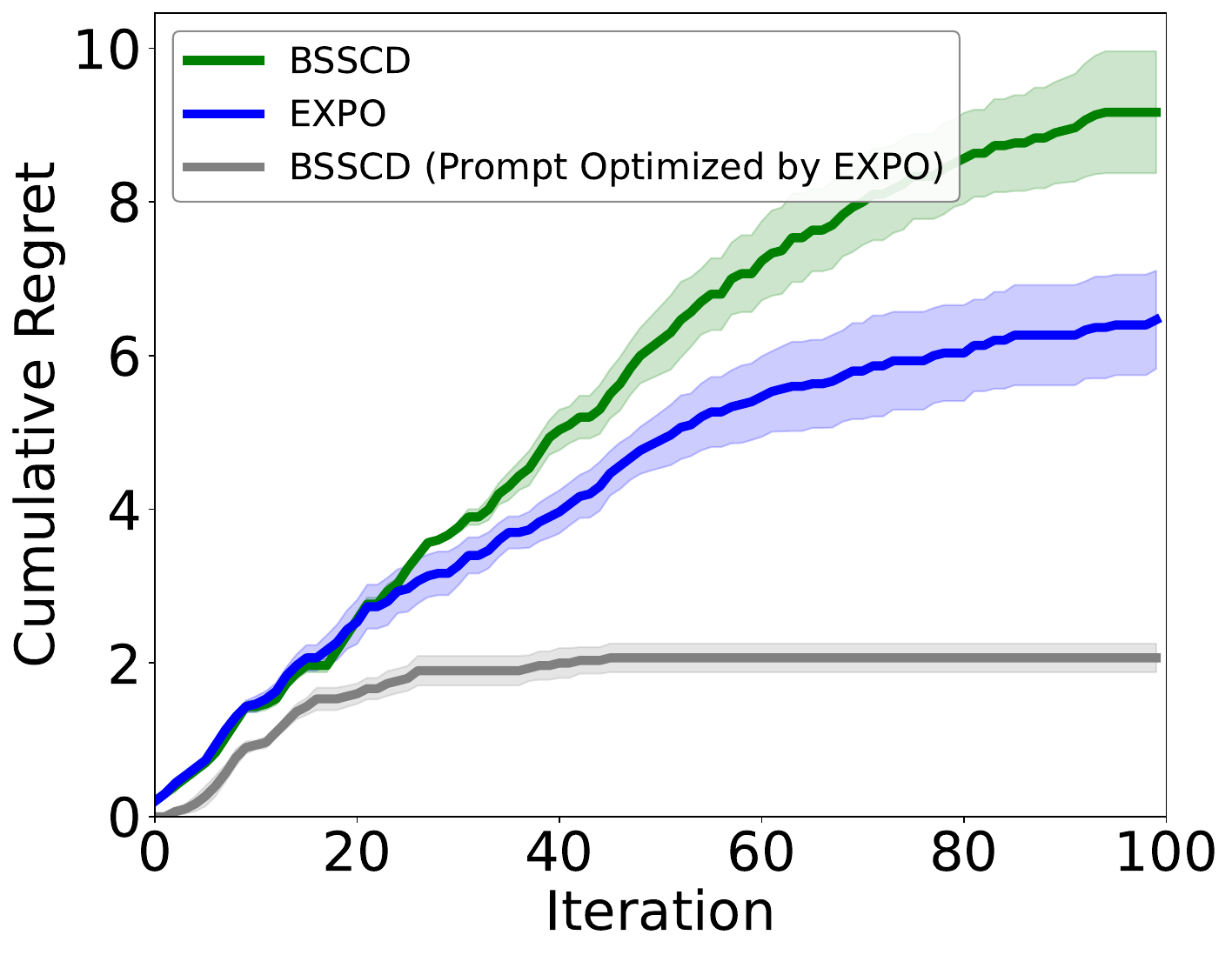}& \hspace{-5mm}
         \includegraphics[width=0.25\linewidth]{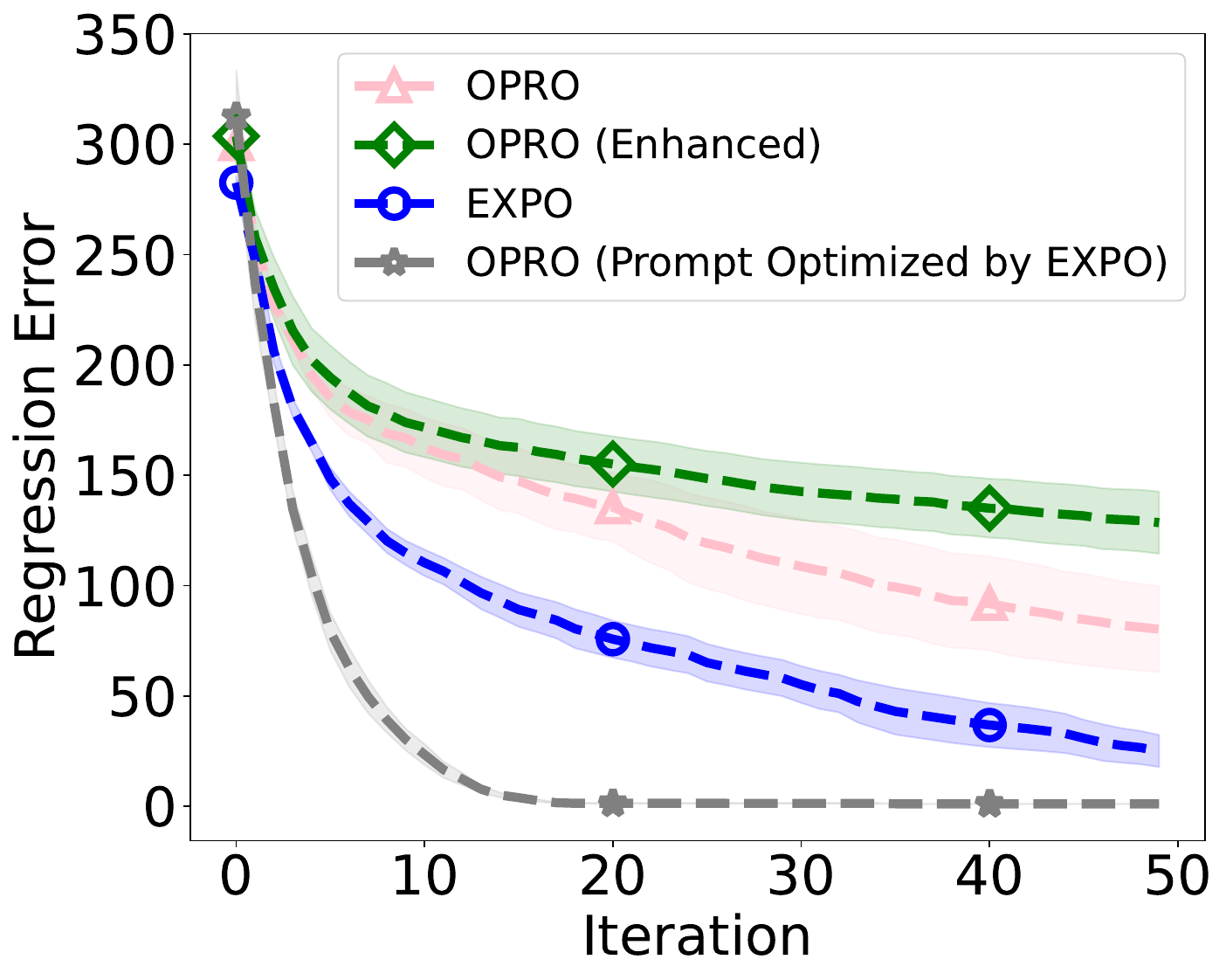}& \hspace{-3mm}
         \includegraphics[width=0.25\linewidth]{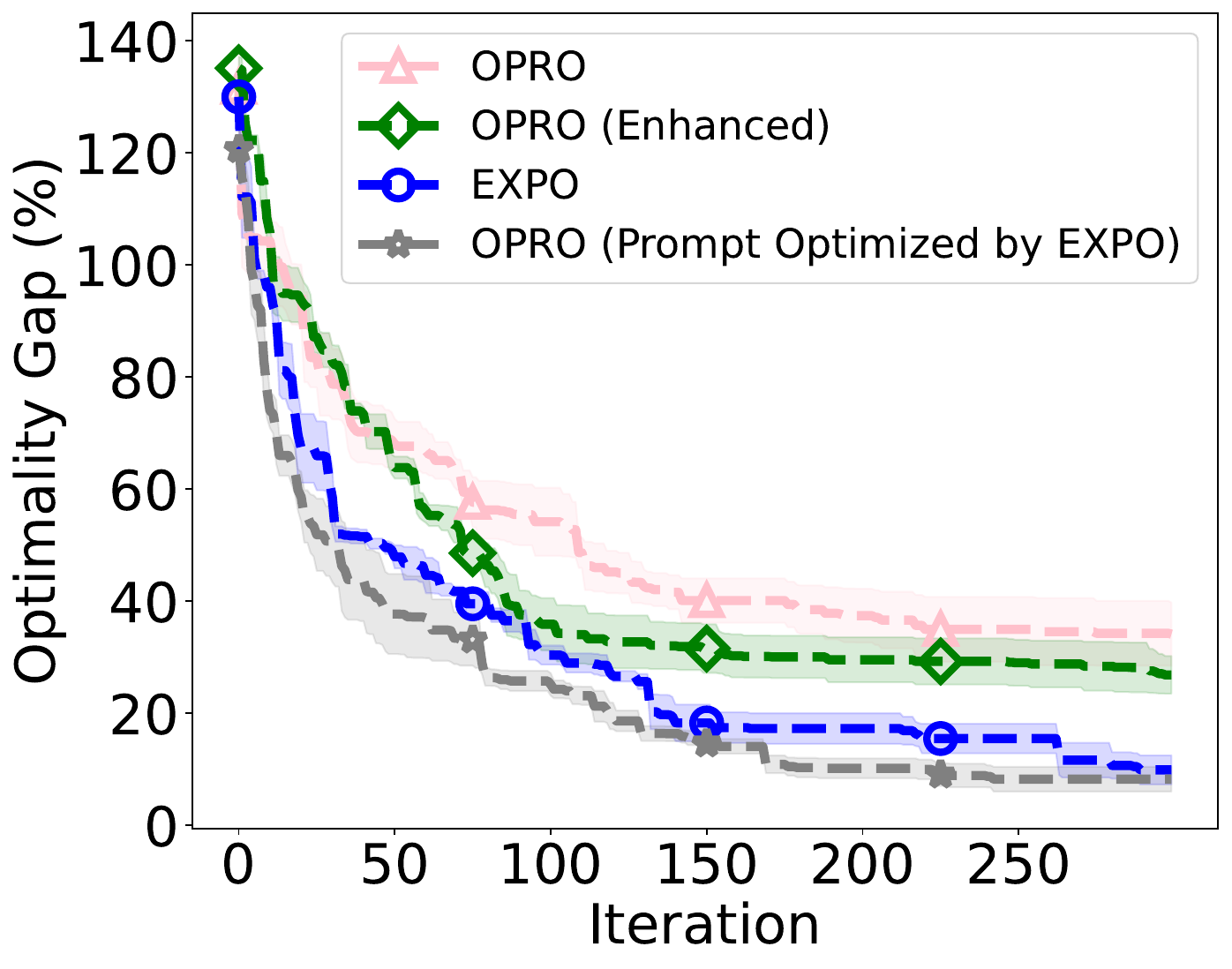}\\
         {\small\hspace{-2mm} BSSND (hard)} & {\small\hspace{-5mm} BSSCD (hard)} & {\small\hspace{-5mm} LR ($w=36$, \ $b=-1$)} & {\small\hspace{-2mm} TSP (20 Nodes, GPT-4-Turbo)}
     \end{tabular}
\vspace{-2mm}
\caption{
    % First three figures: ablation study on impact of exploration parameter $\eta$.
    % Rightmost figure: results using GPT-4-Turbo.
Results achieved by fixing the meta-prompt to be the optimal one discovered by
% using the meta-prompt optimized by 
our \alg~(gray curves).
}
% \label{fig:cumulative_regret_BSSND_BSSCD}
\label{fig:optimal_prompt_on_tasks}
\vspace{-1mm}
\end{figure*}

To further verify the ability of our \alg~to identify effective meta-prompts, we replace the original task description and meta-instruction in an LLM-based sequential decision-making algorithm (e.g., OPRO) by the optimal ones discovered by our \alg: we first run \alg~to completion, and then use the final meta-prompt it selects to execute OPRO again.
The results in Fig.~\ref{fig:optimal_prompt_on_tasks} show that fixing the meta-prompt to the one optimized by our \alg~leads to a dramatic performance boost for LLM-based sequential decision making.

\subsection{Results of GPT-4-turbo for TSP}
\label{app:ablation:subsec:gpt4-turbo}
\begin{figure*}[h]
    \centering
    \includegraphics[width=0.39\linewidth]{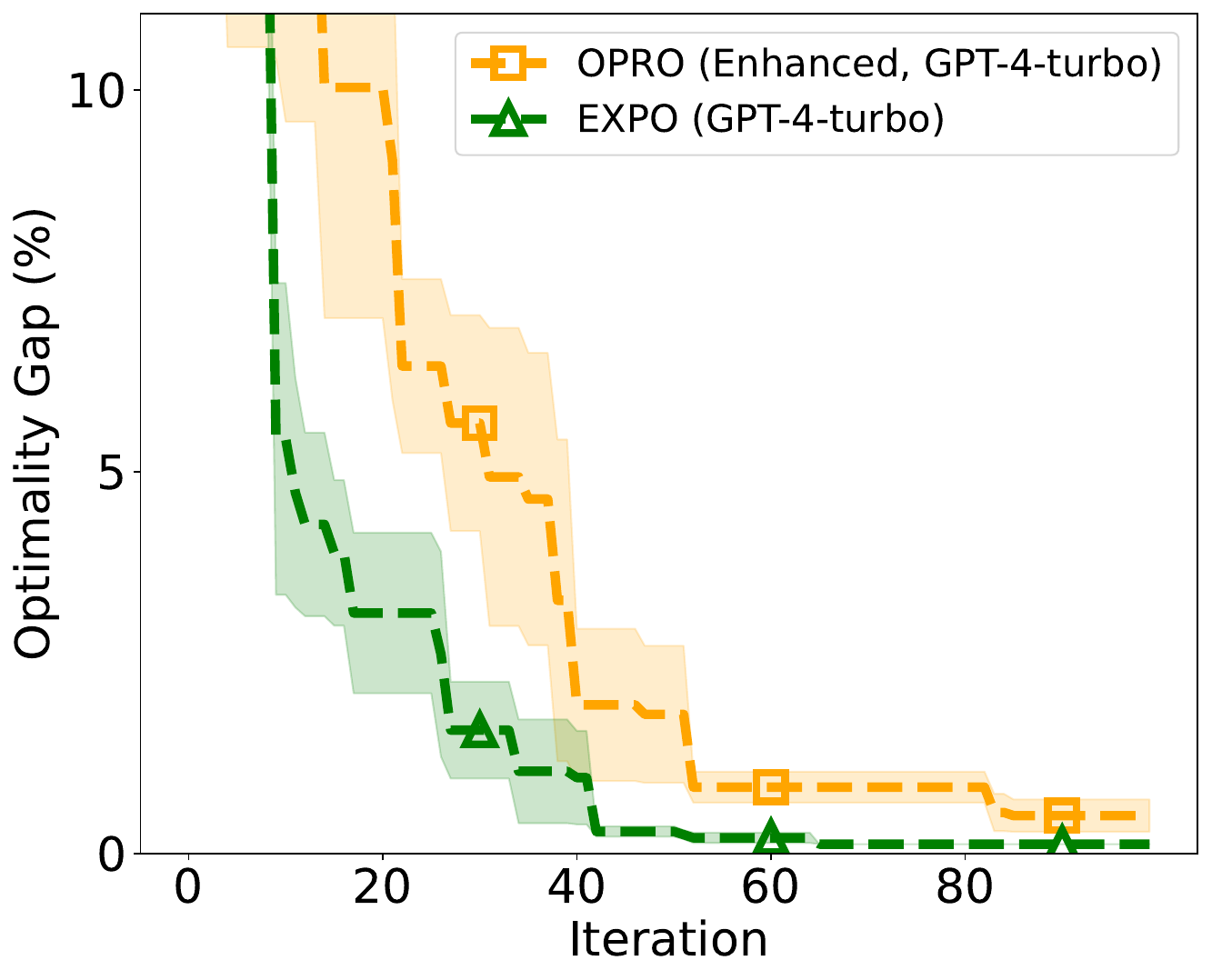}    
    \caption{Ablation study results using GPT-4-turbo in the TSP task with 20 nodes.}
    \label{fig:TSP_ablation}
\end{figure*}

Fig.~\ref{fig:TSP_ablation} shows a zoomed version of Fig.~\ref{fig:TSP_ablation:eta} (bottom right) in the main paper. It shows that when GPT-4-Turbo is used as the LLM, our \alg~is still able to significantly outperform OPRO.

\subsection{Results of Other Variants of OPRO}
\label{exp:app:opro:full:results}
\begin{figure*}[h]
\vspace{3mm}
\centering
\begin{tabular}{cc}
    \includegraphics[width=0.28\linewidth]{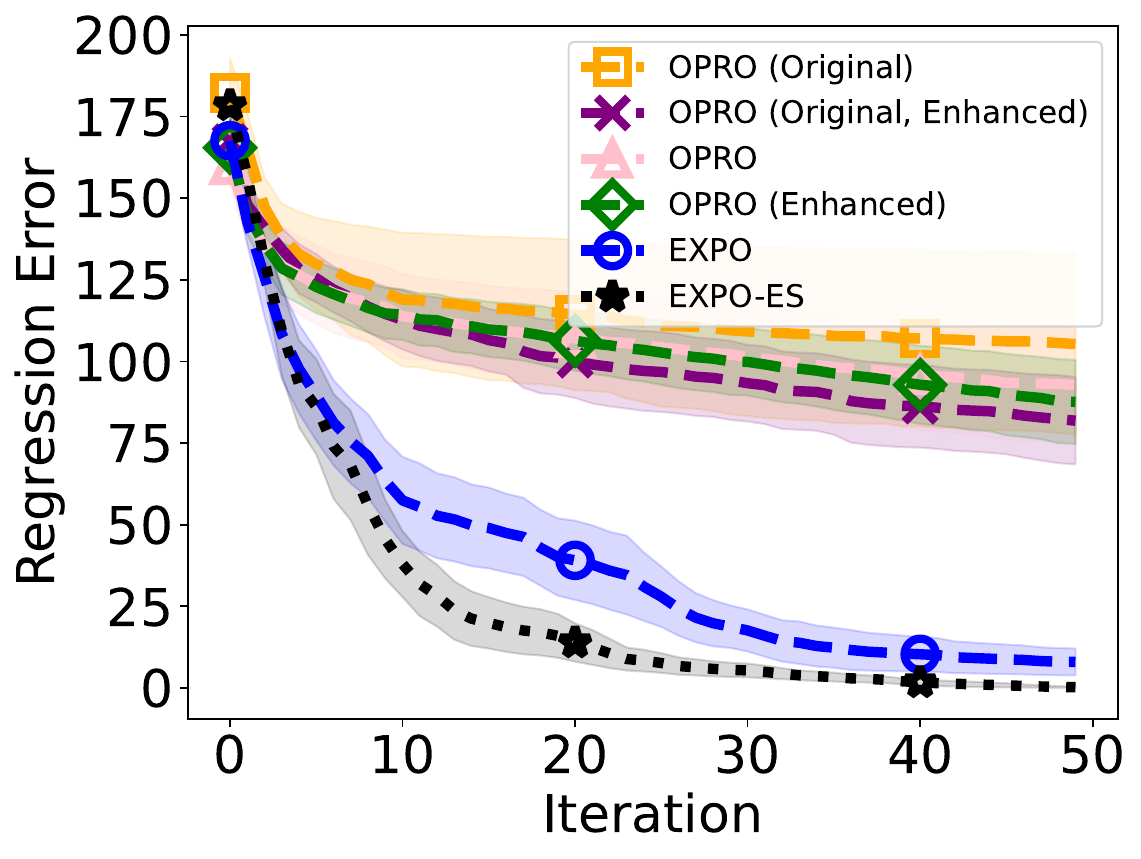} &
    \includegraphics[width=0.28\linewidth]{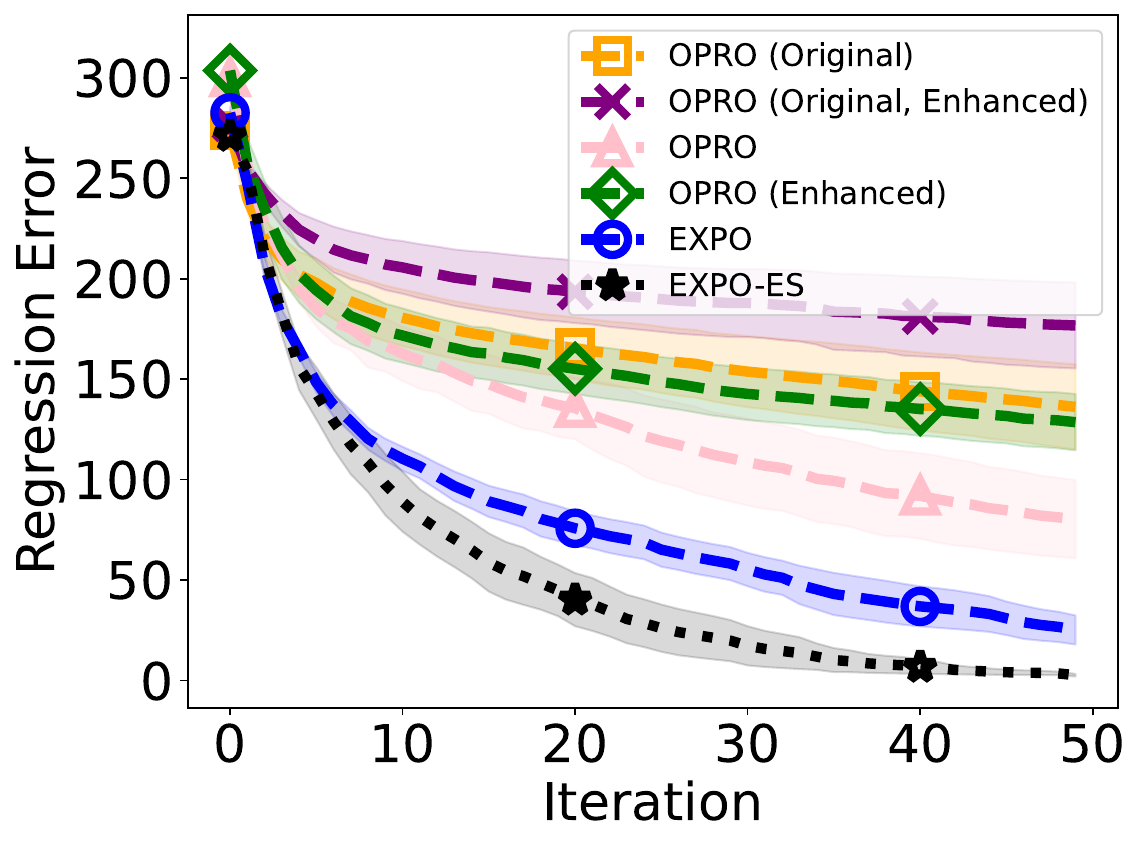} \\
    \makebox[0.28\linewidth]{\makecell{Linear Regression \\ ($w=2, \ b=30$)}} &
    \makebox[0.28\linewidth]{\makecell{Linear Regression \\ ($w=36, \ b=-1$)}} \\
\end{tabular}
\vspace{5mm}
\begin{tabular}{cc}
    \includegraphics[width=0.28\linewidth]{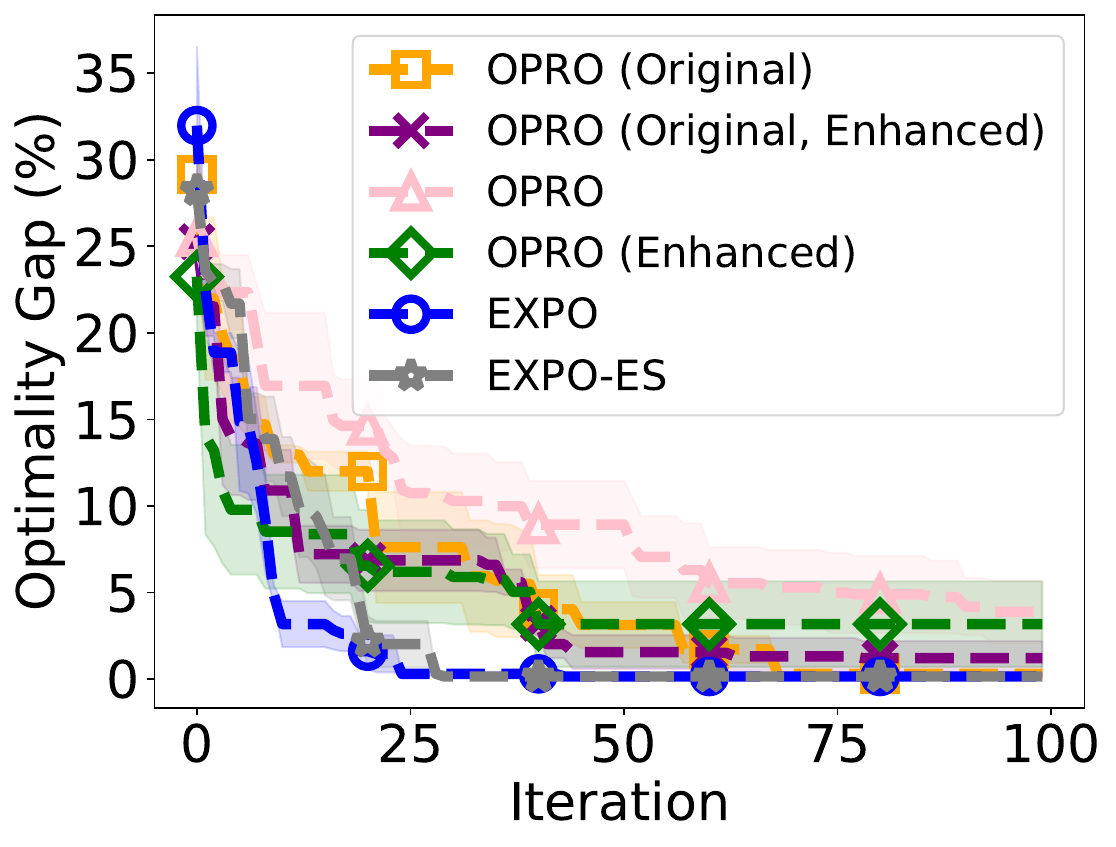} &
    \includegraphics[width=0.28\linewidth]{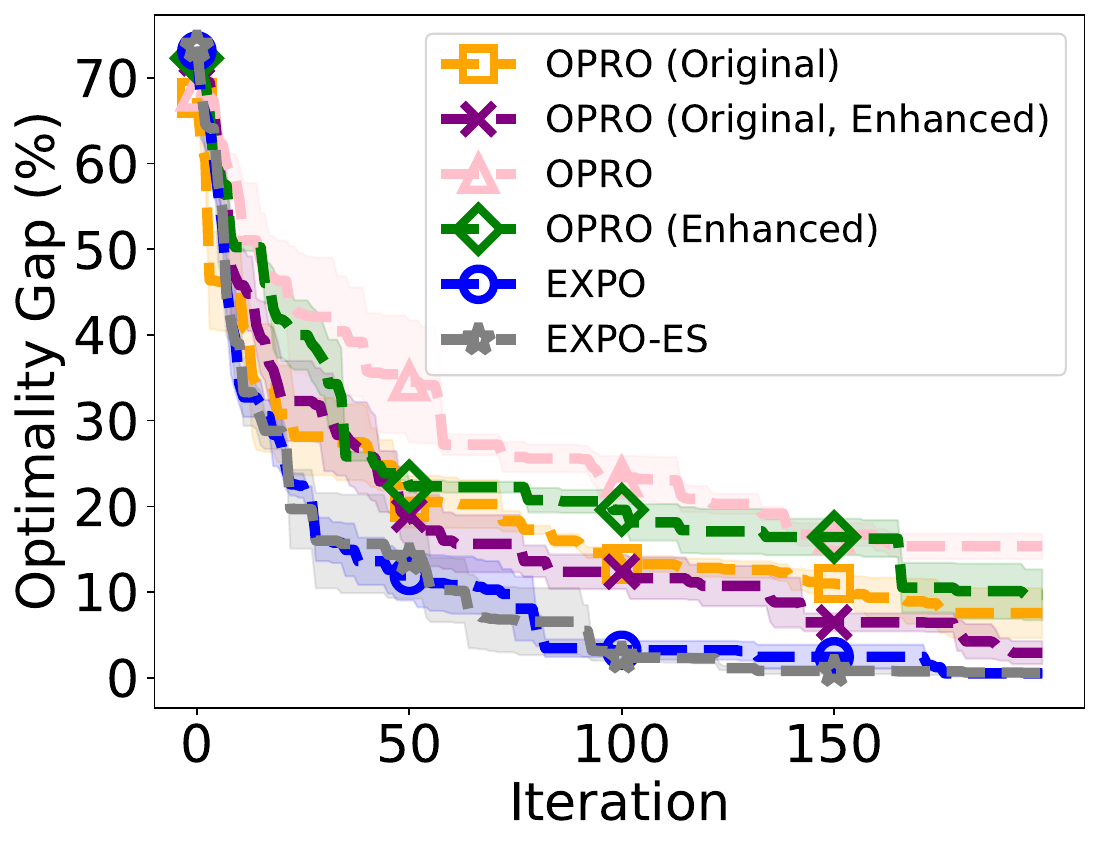} \\
    \makebox[0.28\linewidth]{\makecell{TSP \\ (10 Nodes)}} &
    \makebox[0.28\linewidth]{\makecell{TSP \\ (15 Nodes)}} \\
\end{tabular}
\vspace{5mm}
\begin{tabular}{c}
    \includegraphics[width=0.28\linewidth]{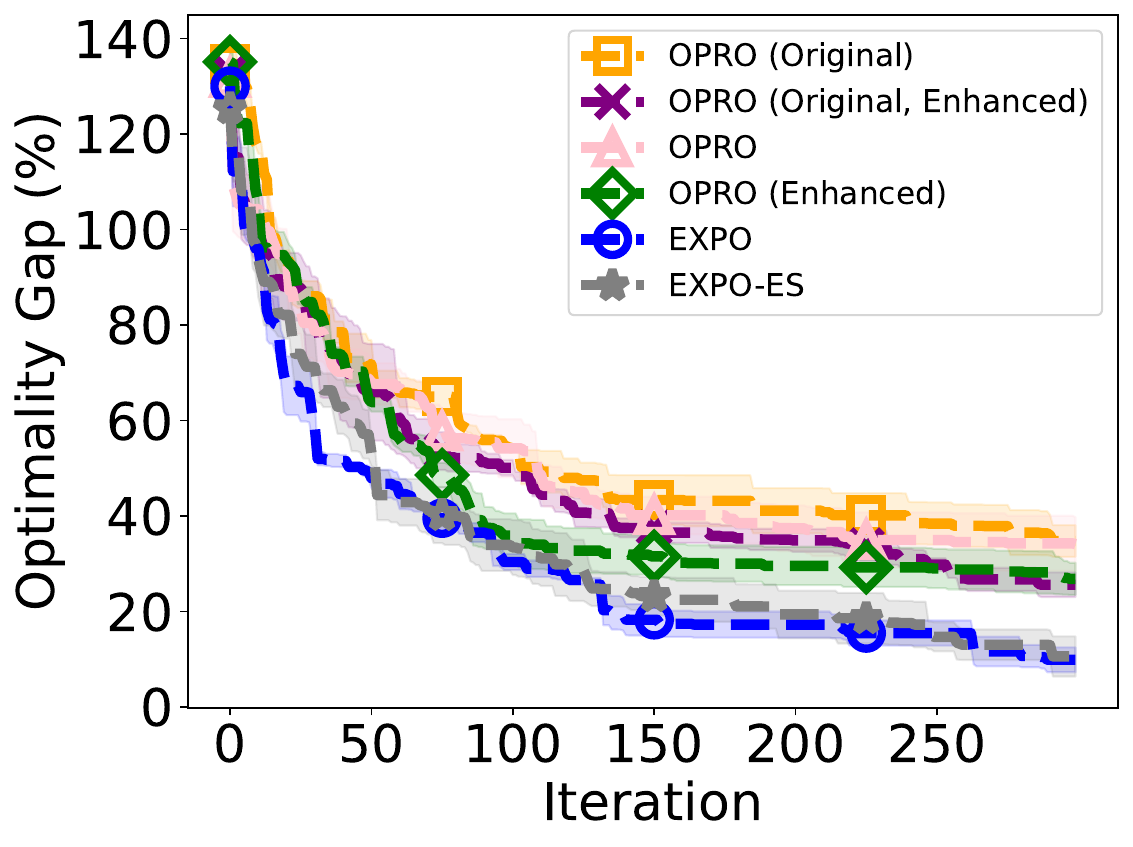} \\
    \makebox[0.28\linewidth]{\makecell{TSP \\ (20 Nodes)}} \\
\end{tabular}
\vspace{-2.5mm}
\caption{
% Full Visualizations for different scenarios: Linear Regression under two settings in the first row and TSP problems for 10, 15, and 20 nodes in the second row.
Results of different algorithms in the Linear Regression task and TSP task (Sec.~\ref{subsec:exp:opro}). We have additionally included the original OPRO (which selects all $8$ actions using a temperature of 1), as well as its enhanced variant.
Lower is better.
}
\label{fig:full_results}
\vspace{-3mm}
\end{figure*}

As we have discussed in Sec.~\ref{subsec:expo} and Sec.~\ref{subsec:exp:opro}, the original OPRO uses a temperature of $1$ to choose all $8$ actions in a batch, while we have made a slight modification such that we choose the last action in the batch with a temperature of $0$. Here we show that this has a minimal impact on the performance of OPRO (Fig.~\ref{fig:full_results}).
Specifically, in Fig.~\ref{fig:full_results}, the orange curves represent the original OPRO (using a temperature of $1$ for all $8$ actions) and the pink curves correspond to our modified version.
We have also compared the performances of the enhanced variants (see Sec.~\ref{subsec:exp:opro} for details) for both the original (purple) and modified OPRO (green).
The results show that setting the temperature to $0$ while selecting the last action has negligible impact on the performance of OPRO.
Importantly, \textbf{our \alg~and \alges~algorithms consistently and dramatically outperform all variants of OPRO}.

\subsection{Impact of Adding Exemplar Embedding to the NN in \alg}
\begin{figure*}[h]
    \centering
    \includegraphics[width=0.24\linewidth]{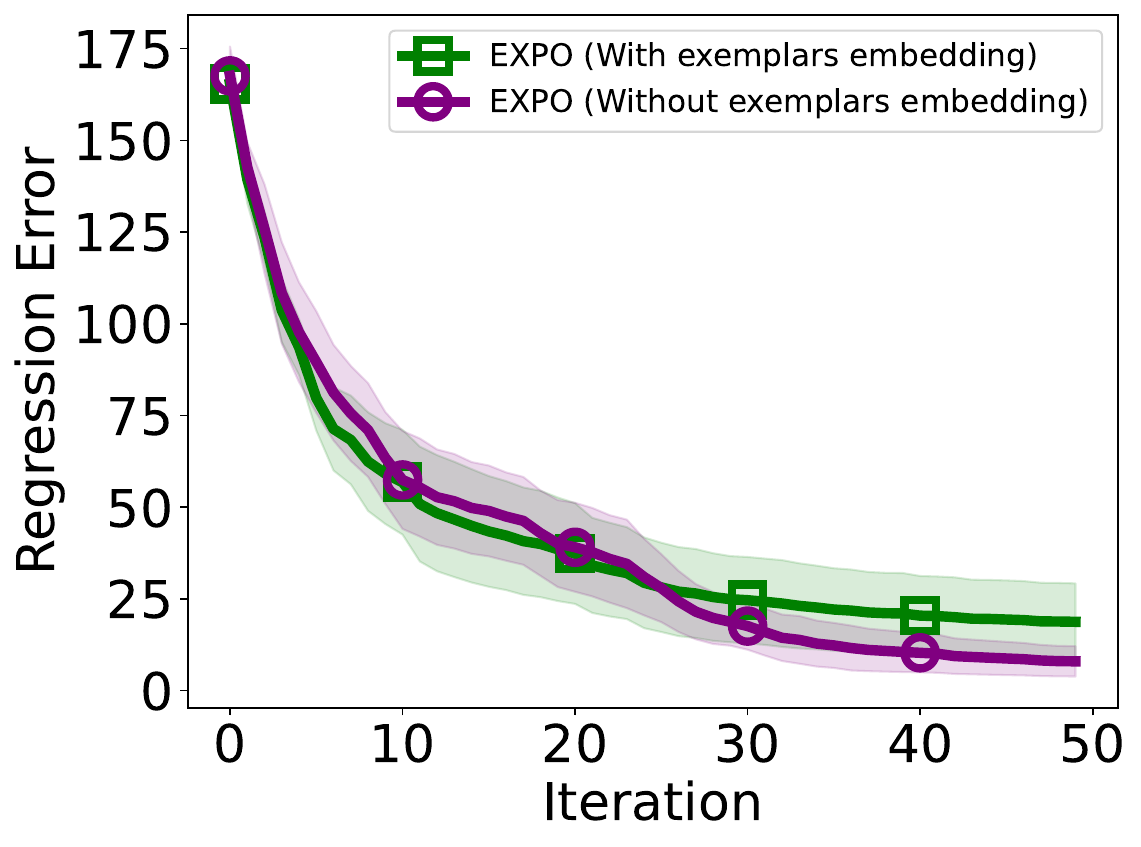}
    \includegraphics[width=0.24\linewidth]{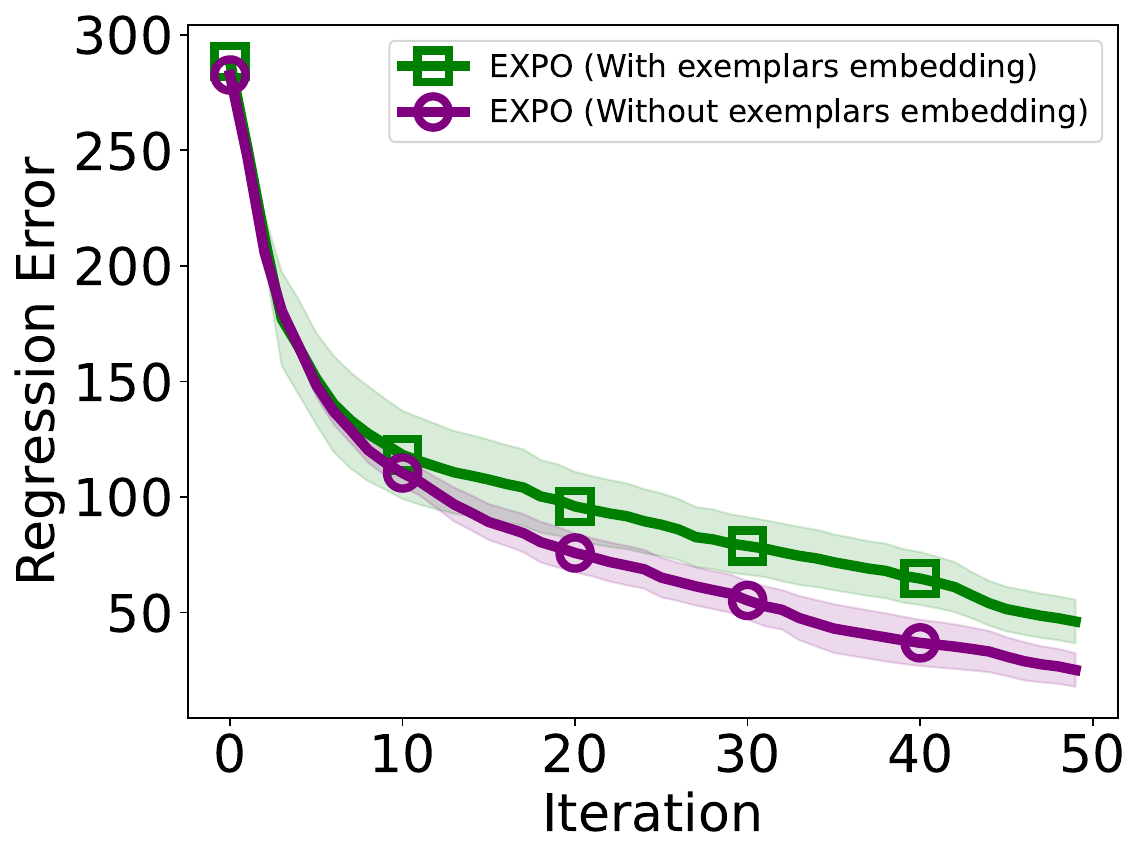}
    \includegraphics[width=0.24\linewidth]{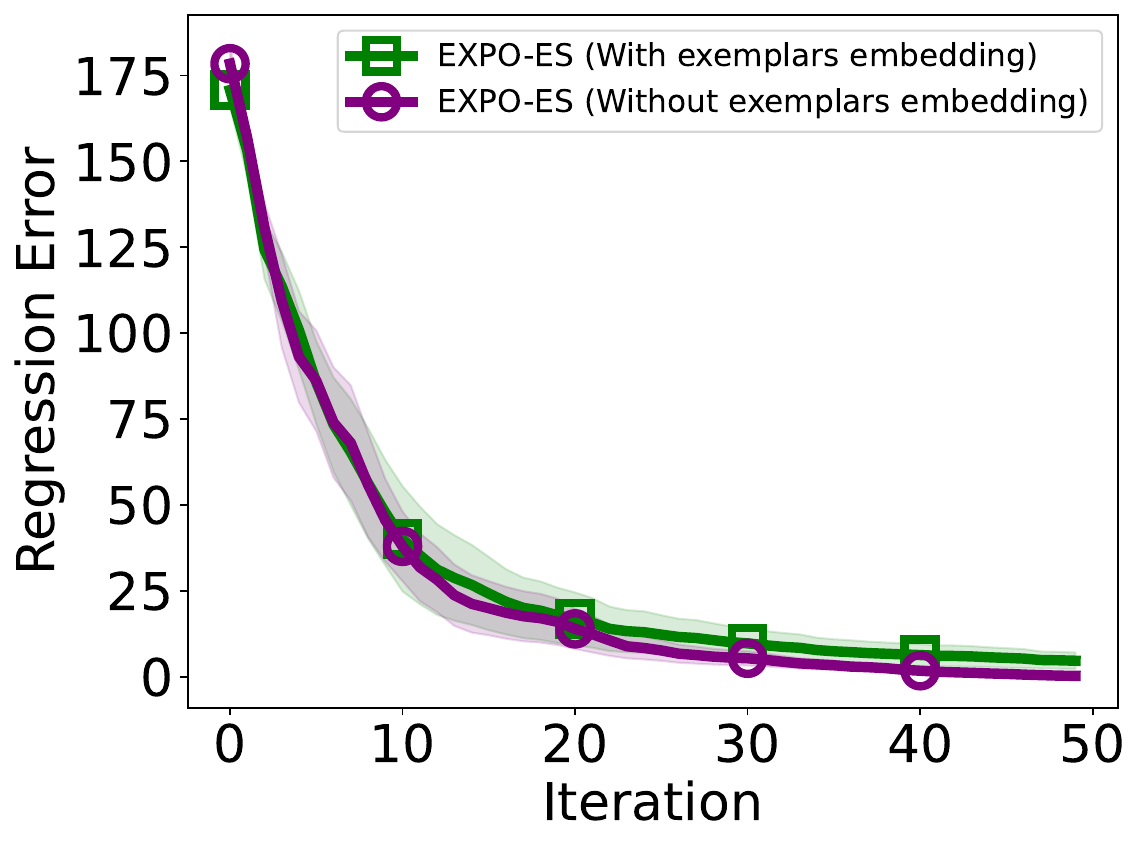}
    \includegraphics[width=0.24\linewidth]{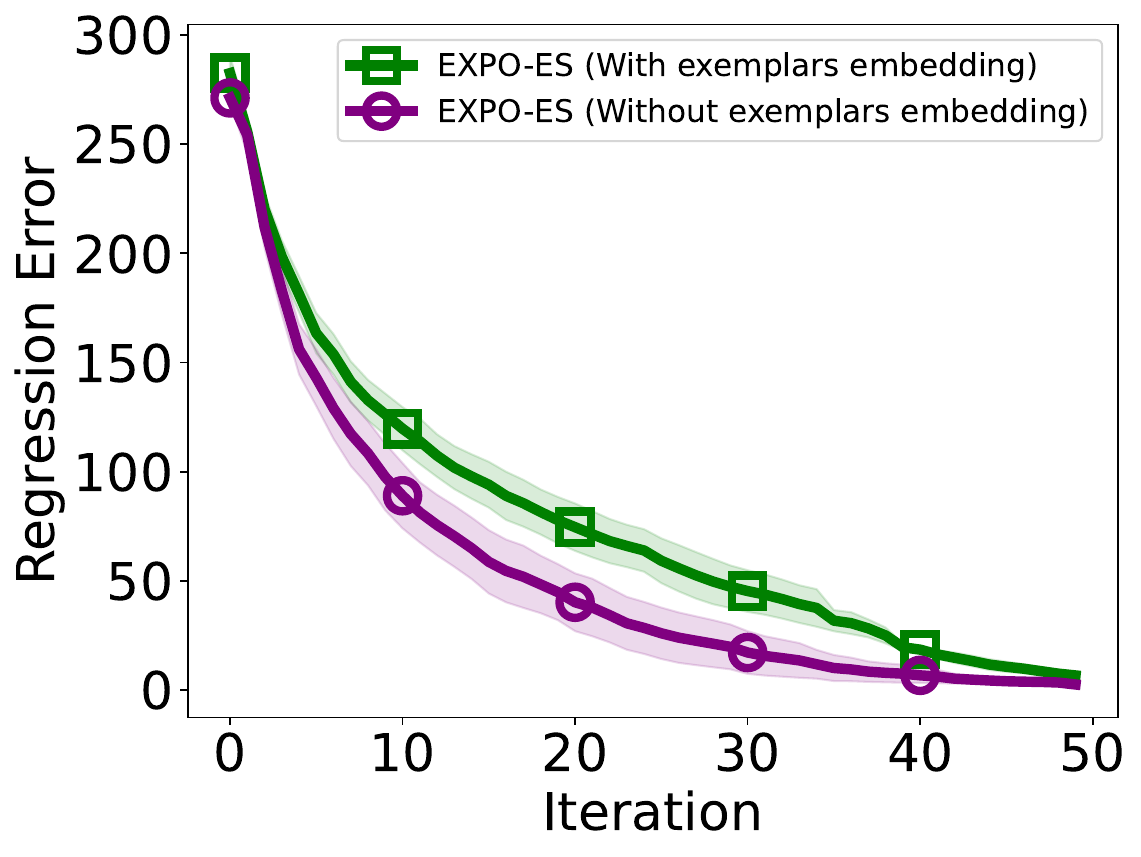}
    \includegraphics[width=0.24\linewidth]{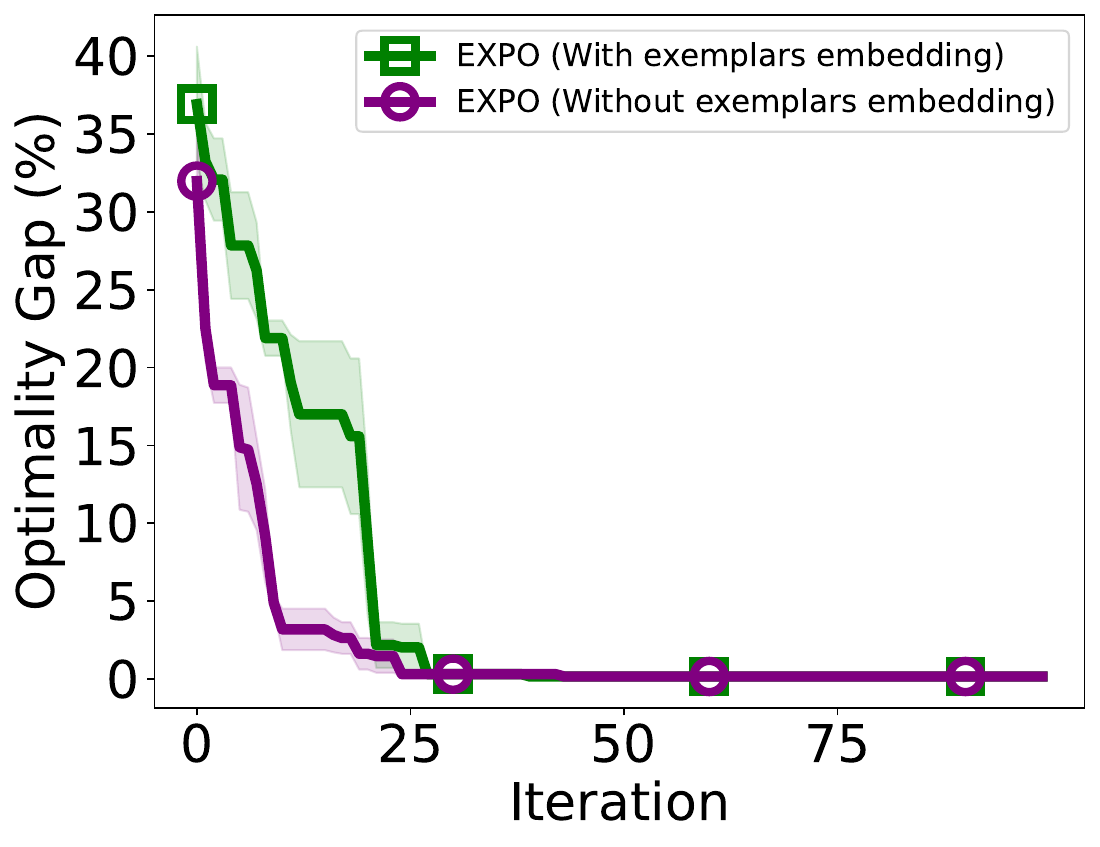}
    \includegraphics[width=0.24\linewidth]{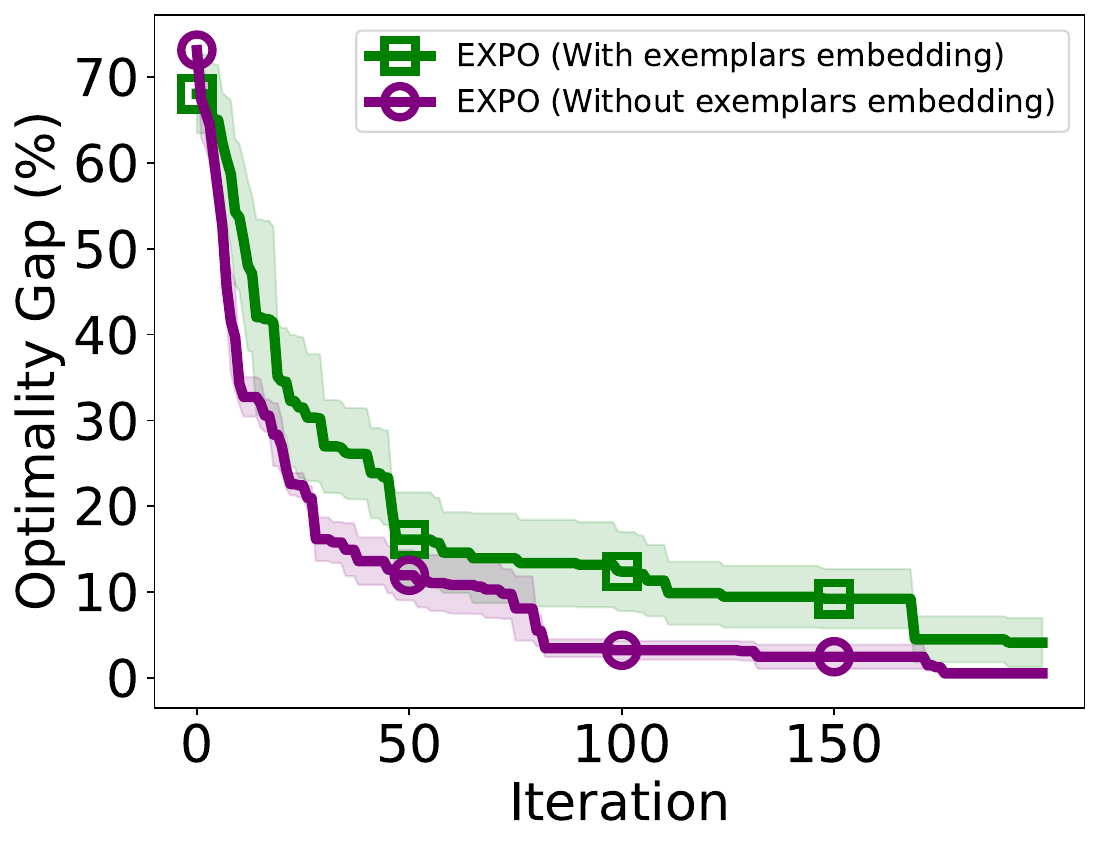}
    \includegraphics[width=0.24\linewidth]{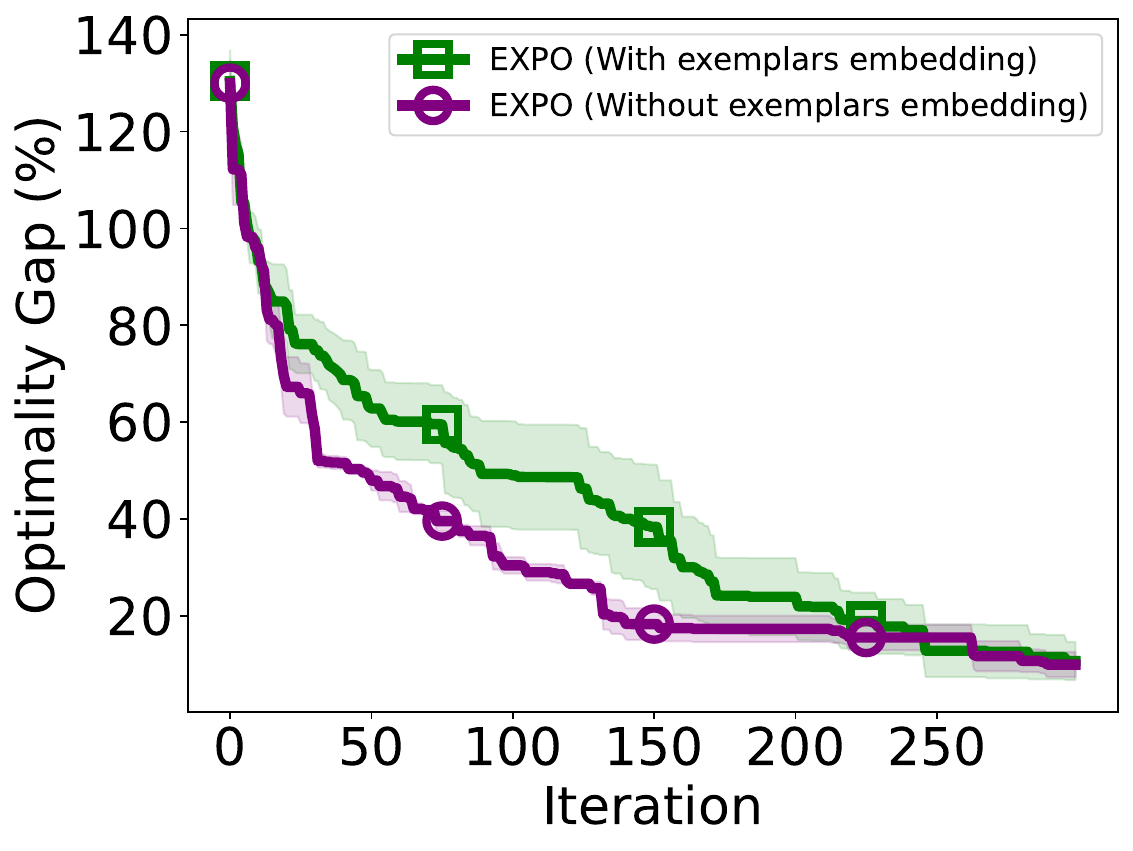}
    \caption{Convergence curves of our \alg~with and without exemplars embedding across different tasks: Linear Regression (top row) and TSP with 10, 15, and 20 nodes (bottom row).}
    \label{fig:with_without_exemplars_embedding}
\end{figure*}

% Recall that in every iteration of our \alg~(Algo.~\ref{algo:EXPO}), we need to train a neural network (NN) $\mathcal{M}(g(\cdot);\theta)$ to estimate the scores of the task descriptions and meta-instructions in the domain (line 8 of Algo.~\ref{algo:EXPO}).
% Note that the training set used to train this NN $\{(\left[g(\mathcal{D}_i) \oplus g(\mathcal{I}_i)\right], s_i)\}_{i=1,\ldots,t+1}$ (line 7 of Algo.~\ref{algo:EXPO}).
% However, also note that in our \alg~algorithm, the set of exemplars included in the meta-prompt $\mathcal{E}'_{t}$ also changes in every iteration.
% Therefore, one may naturally wonder whether additionally including the embedding of $\mathcal{E}'_{t}$ in the input of the NN can further improve the performance of the trained NN and hence the performance of the overall \alg.
% We use ablation study to validate this here, and the results are shown in Fig.~\ref{fig:with_without_exemplars_embedding}.
% The results show that additionally including the embedding of the exemplars in the input of the NN does not lead to better performance than our standard approach of not including it (Algo.~\ref{algo:EXPO}).
% This is likely due to the significantly increased dimensionality of the input to the NN, which makes the training of the NN more difficult.
% Therefore, these results may imply that the benefit of additionally taking into account the changing exemplars is outweighed by the drawback of significantly increased dimensionality of the input to the NN.
Recall that in every iteration of our \alg~(Algo.~\ref{algo:EXPO}), we need to train a neural network (NN) $\mathcal{M}(g(\cdot);\theta)$ to estimate the scores of the task descriptions and meta-instructions in the domain (line 8 of Algo.~\ref{algo:EXPO}).  
Note that the training set used to train this NN is $\{(\left[g(\mathcal{D}_i) \oplus g(\mathcal{I}_i)\right], s_i)\}_{i=1,\ldots,t+1}$ (line 7 of Algo.~\ref{algo:EXPO}).  
However, it is also important to note that in our \alg~algorithm, the set of exemplars included in the meta-prompt $\mathcal{E}'_{t}$ changes in every iteration and hence may also affect the scores $s_i$'s.  
Therefore, one may naturally wonder whether including the embedding of $\mathcal{E}'_{t}$ in the input of the NN can further improve the performance of the trained NN and, consequently, the performance of the overall \alg.  
We conduct an ablation study to validate this hypothesis, and the results are shown in Fig.~\ref{fig:with_without_exemplars_embedding}.  
The results demonstrate that including the embedding of the exemplars in the input of the NN does not lead to better performance than our standard approach of excluding it (Algo.~\ref{algo:EXPO}).  
This is likely due to the significantly increased dimensionality of the input to the NN, which makes training the NN more challenging.  
Therefore, these results suggest that the benefit of additionally accounting for the changing exemplars is outweighed by the drawback of the significantly increased dimensionality of the input to the NN.

% {\color{red}
\subsection{Computational Efficiency of Our Algorithms}
\label{app:subsec:computational:efficiency}
To compare the computational costs of our \alg~and OPRO, using the Linear Regression experiment ($w=2,b=30$), we recorded the time required for each method to achieve a mean squared error (MSE) below various thresholds. The results are shown in Table \ref{tab:mse_comparison} below.
\begin{table*}[t]
\centering
\resizebox{\textwidth}{!}{%
\begin{tabular}{lccccc}
\toprule
\textbf{Methods} & \textbf{Time (s) to MSE $\leq$ 100} & \textbf{Time (s) to MSE $\leq$ 50} & \textbf{Time (s) to MSE $\leq$ 25} & \textbf{Time (s) to MSE $\leq$ 10} & \textbf{Total Runtime (s)} \\
\midrule
\textbf{OPRO} & 64.1 & 170.1 & NA & NA & 731.4 \\
\textbf{EXPO (Ours)} & 94.5 & 207.4 & 263.1 & 350.8 & 1645.6 \\
\bottomrule
\end{tabular}%
}
\caption{Comparison of our \alg~and OPRO in terms of the time (s) to reach different MSE thresholds and total runtime.}
\label{tab:mse_comparison}
\end{table*}
% }

The results show that EXPO requires only $263.1$ seconds to achieve MSE  $\leq 25$, whereas OPRO is unable to achieve this threshold even after $731.4$ seconds. 
Similarly, EXPO achieves MSE $\leq 10$ in just $350.8$ seconds, which is also less than OPRO’s total runtime of $731.4$ seconds (after 300 iterations).
Therefore, to reach the same performance level, our \alg~requires significantly smaller amount of computational time than OPRO.

\subsection{More Illustrations of the Discovered Task Description and Meta-instruction}
\label{app:subsec:more:illustration:meta-prompt}
Here we provide more illustrations regarding the comparison of the original task description and meta-instruction adopted by the original LLM-based sequential decision-making algorithm (i.e., OPRO or the LLM-based MAB algorithm from \cite{krishnamurthy2024can}) and those optimized by our \alg~algorithm.
We include the comparisons for the TSP task (Fig.~\ref{fig:example:descriptions:tsp}), and the two different prompt designs for the LLM-based MAB task in Sec.~\ref{subsec:exp:bandits} (Fig.~\ref{fig:example:descriptions:MABbssnd} and Fig.~\ref{fig:example:descriptions:MAB}).

\begin{figure*}[h]
% \onecolumn
\begin{minipage}[t]{0.4\textwidth}
\begin{mdframed}[linewidth=0.9pt]  % adjust linewidth as you desire
\footnotesize  % adjust text size as required
\centerline{{\normalsize  OPRO}}
    {\color{purple}You are given a list of points with coordinates below: \{POINTS\}.\\
    Below are some previous traces and their lengths. 
The traces are arranged in descending order based on their lengths, 
where lower values are better.} \\ \\ \\
    \{EXEMPLARS\}
    \\\\
    {\color{blue}Give me a new trace that is different from all traces above, 
and has a length lower than any of the above. 
The trace should traverse all points exactly once. 
The trace should start with \texttt{<trace>} and end with \texttt{</trace>}.}
    % Output:
\end{mdframed}
\end{minipage}
\hfill
\begin{minipage}[t]{0.59\textwidth}
\begin{mdframed}[linewidth=0.9pt]  % adjust linewidth as you desire
    \footnotesize  % adjust text size as required
\centerline{{\normalsize  \alg}}
    {\color{purple}You are provided with a dataset containing a list of coordinates labeled as \{POINTS\}. \par
The dataset also includes a series of previously calculated routes, with associated lengths that are ordered from longest to shortest. 
However, it’s key to note that shorter routes are more desirable. 
Despite the presentation order, understand that the optimal route is identified by the smallest total length.}  \\ \\
    \{EXEMPLARS\}
    \\\\
    {\color{blue}Provide a unique trace that is distinct from any previous traces and shorter in length. 
Ensure that this trace visits each point exactly once and adhere to the specified format 
by starting with \texttt{<trace>} and concluding with \texttt{</trace>}.} \\ \\
\end{mdframed}
\end{minipage}
\caption{
The {\color{purple}task description} (top) and {\color{blue}meta-instruction} (bottom) used by OPRO (left) and optimized by our \alg~(right) in a TSP task.
}
\label{fig:example:descriptions:tsp}
\end{figure*}

\begin{figure*}[h]
% \onecolumn
\begin{minipage}[t]{0.4\textwidth}
\begin{mdframed}[linewidth=0.9pt]  % adjust linewidth as you desire
% \scriptsize  % adjust text size as required
\footnotesize
\centerline{{\normalsize  BSSND}}
    {\color{purple}You are a bandit algorithm in a room with 5 buttons labeled blue, green, red, yellow, purple. Each button is associated with a Bernoulli distribution with a fixed but unknown mean; the means for the buttons could be different. For each button, when you press it, you will get a reward that is sampled from the button's associated distribution. You have 100 time steps and, on each time step, you can choose any button and receive the reward. Your goal is to maximize the total reward over the 100 time steps.} \\ \\ \\ \\
    {\color{blue}At each time step, I will show you a summary of your past choices and rewards. Then you must make the next choice. You may output a distribution over the 5 buttons formatted EXACTLY like "blue:a,green:b,red:c,yellow:d,purple:e".}
    \\ \\
    % Output:
\end{mdframed}
\end{minipage}
\hfill
\begin{minipage}[t]{0.59\textwidth}
\begin{mdframed}[linewidth=0.9pt]  % adjust linewidth as you desire
    % \scriptsize  % adjust text size as required
    \footnotesize
\centerline{{\normalsize  \alg}}
    {\color{purple}You are presented as a bandit algorithm, located in an environment offering five distinct buttons, each emblazoned with colors such as blue, green, red, yellow, and purple. Each button acts a vessel tied to a non-variable yet undisclosed Bernoulli distribution mean which isn't subjected to be uniformly distributed across buttons. In this mechanism, every button acts as a yielder of a capricious reward, constructed from the associated distribution of the respective button. With access to total life -encompassing around 100 temporal stages - your voluntary element grants you control towards opting the button insertion at each such progressive phase. Precisely summoning your approach could perpetually provide you with a regulatory provision\_, the aptitude - is flexibly dwelling within its underlining motive- aiming at optimizing total accumulated cashbacks during several phases of these 100 spatial temporalities.} \\ \\
    {\color{blue}During every step of the process, a recap highlighting your previous selections and the prizes received will be presented to you. Then, it'll now be incumbent upon you to proceed with the new decision-making process. For your ease, a well-structured distribution comprising five buttons in assorted colours such as "blue", "green,", "red", "yellow", and "purple" will be exhibited before you. Make sure to structure your output accordingly; this might look something akin to "blue:a,green:b,red:c,yellow:d,purple:e".}
\end{mdframed}
\end{minipage}
\caption{
The {\color{purple}suggestive framing} (corresponding to the task description) and {\color{blue}MAB problem description} (corresponding to the meta-instruction) used by BSSND \(hard\) (left) and optimized by our \alg~(right) in an LLM-based MAB task.
}
\label{fig:example:descriptions:MABbssnd}
\end{figure*}

\begin{figure*}[t]
% \onecolumn
\begin{minipage}[t]{0.42\textwidth}
\begin{mdframed}[linewidth=0.9pt]  % adjust linewidth as you desire
% \scriptsize  % adjust text size as required
\footnotesize
\centerline{{\normalsize  BSSCD}}
    {\color{purple}You are a bandit algorithm in a room with 5 buttons labeled blue, green, red, yellow, purple. Each button is associated with a Bernoulli distribution with a fixed but unknown mean; the means for the buttons could be different. For each button, when you press it, you will get a reward that is sampled from the button's associated distribution. You have 100 time steps and, on each time step, you can choose any button and receive the reward. Your goal is to maximize the total reward over the 100 time steps.} \\ \\
    {\color{blue}At each time step, I will show you a summary of your past choices and rewards. Then you must make the next choice. You may output a distribution over the 5 buttons formatted EXACTLY like "blue:a,green:b,red:c,yellow:d,purple:e". Let’s think step by step to make sure we make a good choice.}
    \\
    % Output:
\end{mdframed}
\end{minipage}
\hfill
\begin{minipage}[t]{0.57\textwidth}
\begin{mdframed}[linewidth=0.9pt]  % adjust linewidth as you desire
    % \scriptsize  % adjust text size as required
    \footnotesize
\centerline{{\normalsize  \alg}}
    {\color{purple}You are an algorithm designed to function as a bandit, positioned within an environment that features five distinct buttons, each colored blue, green, red, yellow, and purple. These buttons are intricately connected to individual Bernoulli distributions which possess unique and undisclosed mean probabilities. When a button is pressed, it delivers a reward based on its specific distribution. Granted with 100 opportunities to act, your objective is to strategically press these buttons in a manner that optimizes the accrued total reward throughout these attempts. Make your selections wisely to maximize the gains from this stochastic setup.} \\ \\
    \\
    {\color{blue}In each phase, a concise recap of your previous decisions and received rewards will be presented. Your task is to make a subsequent choice based on this data. It is essential to output your selection in an exact format defined as "blue:a, green:b, red:c, yellow:d, purple:e", where 'a', 'b', 'c', 'd', 'e' represent specific z-score values for each color accompanied by the decision choice letter(s). The process is designed to refine our strategy progressively with each move, ensuring an informed and impactful outcome.}
\end{mdframed}
\end{minipage}
\caption{
The {\color{purple}suggestive framing} (corresponding to the task description) and {\color{blue}MAB problem description} (corresponding to the meta-instruction) used by BSSCD \(hard\) (left) and optimized by our \alg~(right) in an LLM-based MAB task.
}
\label{fig:example:descriptions:MAB}
\end{figure*}

\subsection{More Results on the Ablation Study Regarding Comparison with General Prompt Optimization Methods}
\label{app:subsec:more:ablation:ucb}
\begin{figure}[h]
\vspace{-3mm}
\centering
\begin{tabular}{ccc}
    \includegraphics[width=0.33\linewidth]{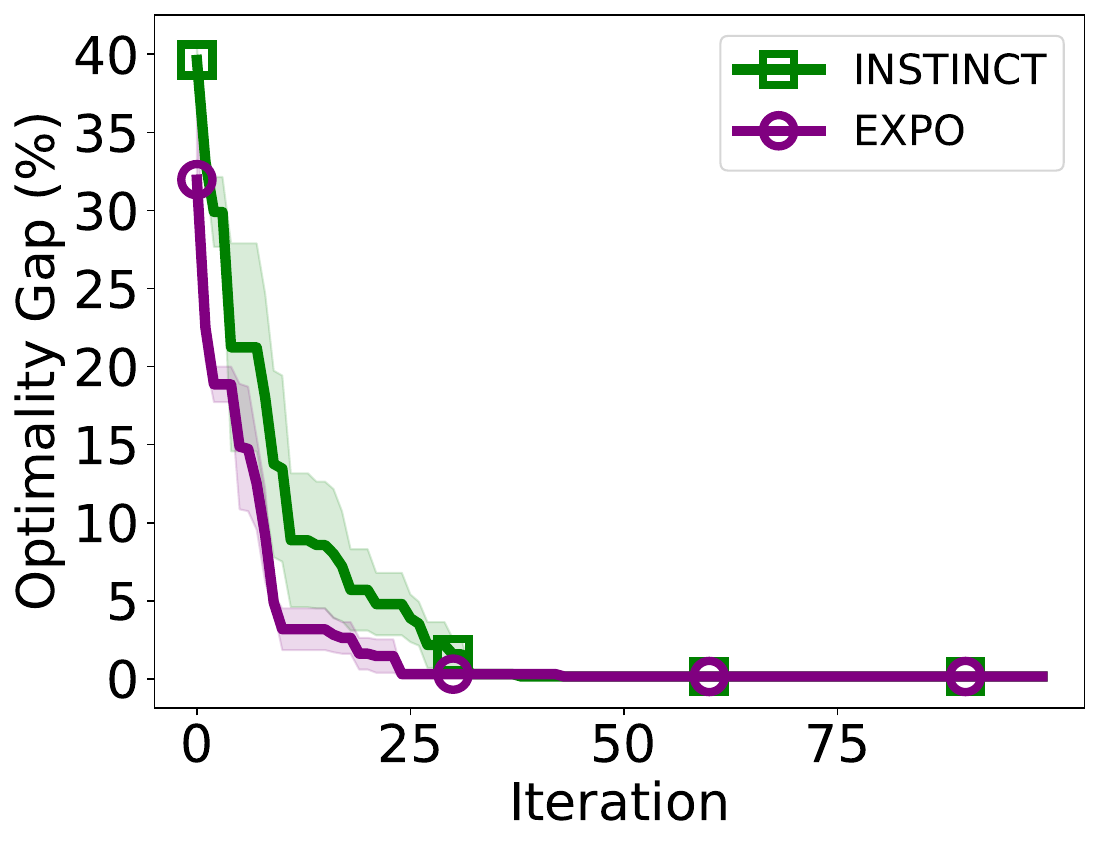} &
    \includegraphics[width=0.33\linewidth]{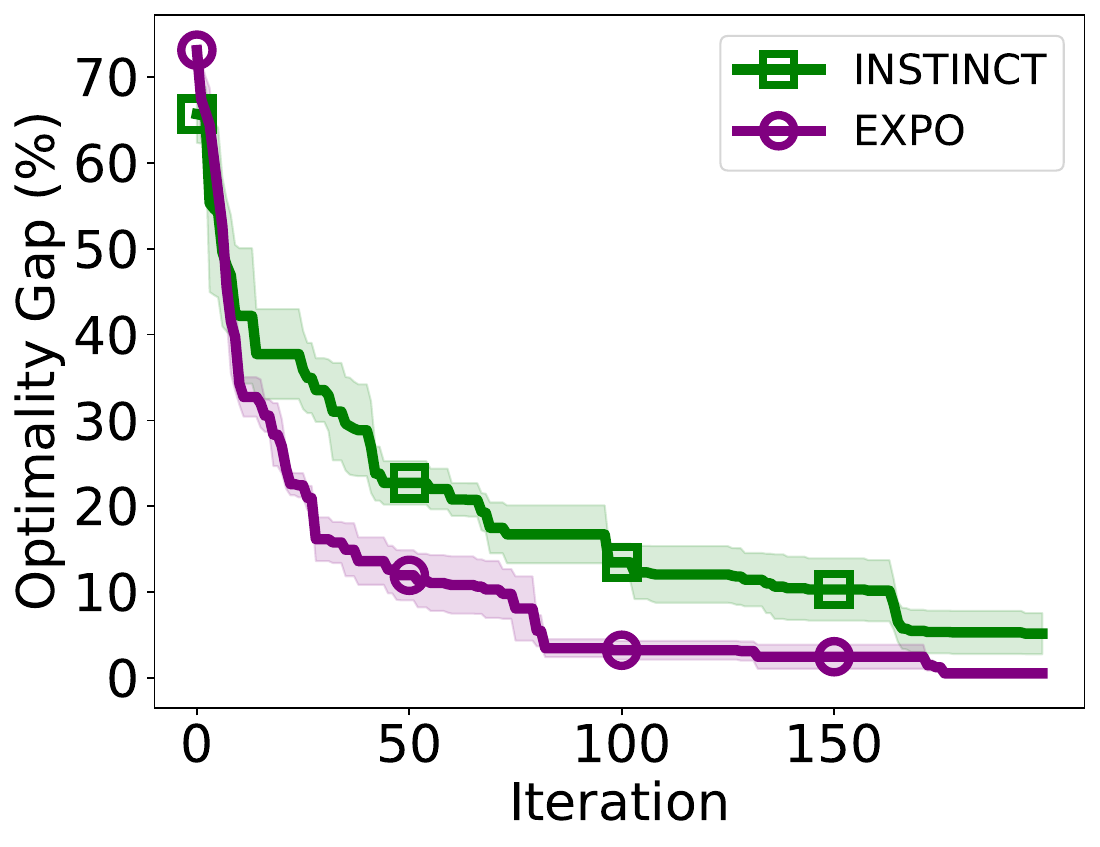} &
    \includegraphics[width=0.33\linewidth]{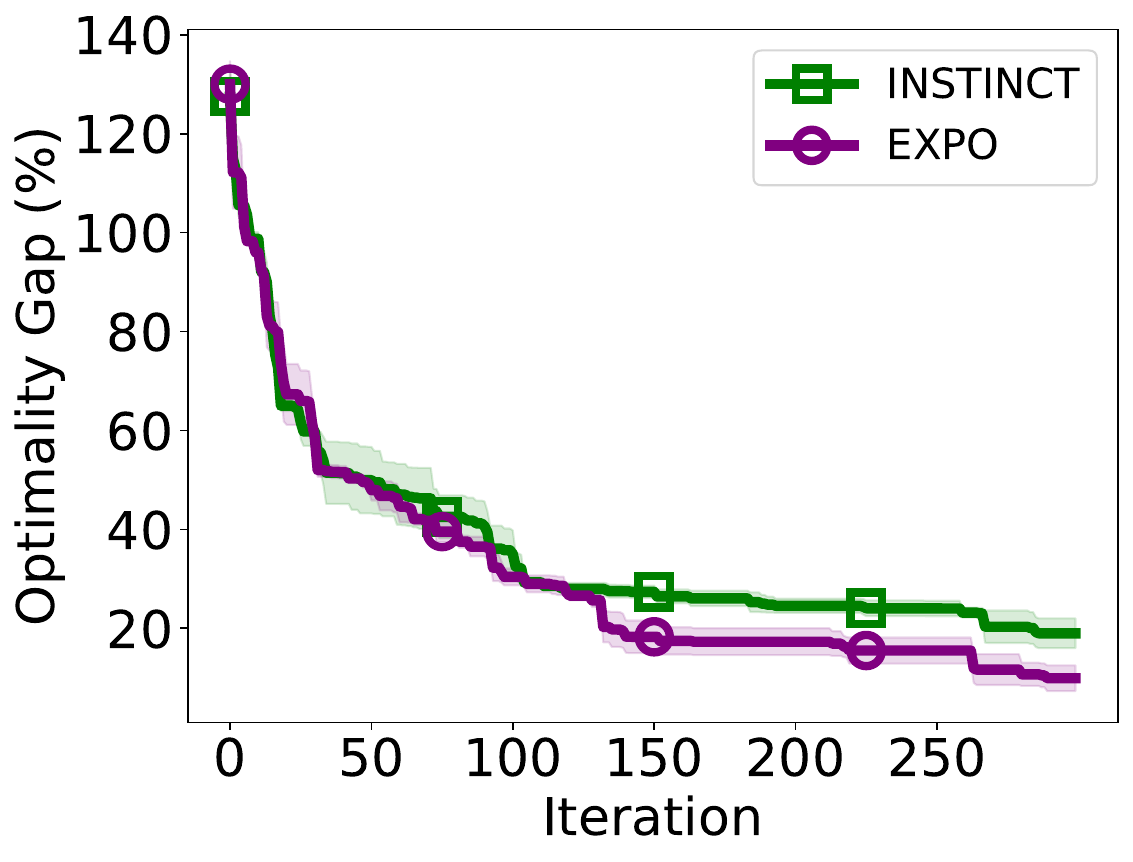} \\
    {\small \makecell{TSP \\ (10 Nodes)}} & {\small \makecell{TSP \\ (15 Nodes)}} & {\small \makecell{TSP \\ (20 Nodes)}} \\
\end{tabular}
\vspace{-2.5mm}
\caption{
Comparison of our \alg~with INSTINCT in the TSP tasks.
% Ablation study results using NeuralUCB. The first row shows results for Linear Regression under two settings. The second row presents the TSP tasks for 10 and 15 nodes, respectively, and the third row provides results for the TSP with 20 nodes.
}
\label{fig:ablation_neuralucb:tsp}
\vspace{-3mm}
\end{figure}

% \textcolor{red}{
Here we use the TSP task to perform an additional ablation study to compare the performance of our \alg~algorithm with the INSTINCT algorithm \cite{lin2023instinct}, which is the best-performing baseline method from Fig.~\ref{fig:ablation_neuralucb:lr}.
The results are shown in Fig.~\ref{fig:ablation_neuralucb:tsp}, which, together with Fig.~\ref{fig:ablation_neuralucb:lr}, demonstrate that our \alg~algorithm based on adversarial bandits significantly and consistently outperforms general prompt optimization methods, including the INSTINCT algorithm which is based on stochastic MAB.
% }

\section{Societal Impacts}
\label{app:sec:societal:impact}
Our paper has the potential to create a positive societal impact by significantly improving the performance of LLM-based agents, thereby enhancing productivity at a societal level. We do not anticipate any negative societal impacts.

\FloatBarrier

\end{document}